%% file: cogcl.tex
\documentclass[10pt,journal,compsoc]{IEEEtran}
\usepackage{diagbox}
\usepackage{graphicx}
\usepackage{verbatim}
\usepackage{multirow}
\usepackage{algorithmic}
\usepackage{longtable}
\usepackage{array}
\usepackage{subfigure}
\usepackage{stfloats}
\usepackage{balance}
\usepackage{multicol}
\usepackage{color}
\usepackage{epstopdf}
\usepackage{bm}
\usepackage{amsmath}
\usepackage{amssymb}
\usepackage{booktabs} % For formal tables
\usepackage{epsfig}
\usepackage{enumitem}
\usepackage{cleveref}
\usepackage{arydshln}
\usepackage{balance}
\usepackage{xspace}
\usepackage{enumitem}

\makeatletter
\newif\if@restonecol
\makeatother

\usepackage{algorithmic}
\usepackage[linesnumbered, ruled, vlined]{algorithm2e}

\newcommand{\tabincell}[2]{\begin{tabular}{@{}#1@{}}#2\end{tabular}}
\newcommand{\hide}[1]{} %hide
\newcommand{\vpara}[1]{\vspace{0.05in}\noindent \textbf{#1 }}
\newcommand{\ipara}[1]{\vspace{0.03in}\noindent \textit{#1 }}

\newcommand{\secref}[1]{Section~\ref{#1}} %section reference
\newcommand{\figref}[1]{Figure~\ref{#1}} %section reference
\newcommand{\beq}[1]{\vspace{-0.03in}\begin{equation}#1\end{equation}\vspace{-0.03in}}
\newcommand{\beqn}[1]{\vspace{-0.04in}\begin{eqnarray}#1\end{eqnarray}\vspace{-0.04in}}

\newtheorem{problem}{Problem}
\newtheorem{definition}{Definition}

\newtheorem{theorem}{Theorem}

\newcommand{\model}{GraphCAD}
\newcommand{\modelgcn}{\text{GraphCAD}_{\text{GCN}}}

\newcommand{\modelgat}{\text{GraphCAD}_{\text{GAT}}}

\newcommand{\modelgin}{\text{GraphCAD}_{\text{GIN}}}

\newcommand{\modelsage}{\text{GraphCAD}_{\text{GSAGE}}}

\newcommand{\smodel}{GraphCAD }

\newcommand{\pmodel}{GraphCAD-pre}
\newcommand{\spmodel}{GraphCAD-pre }

\newcommand{\yx}[1]{\textbf{\color{red}[(YX: #1 )]}}  % to fix

\AtBeginDocument{%
	\providecommand\BibTeX{{%
			\normalfont B\kern-0.5em{\scshape i\kern-0.25em b}\kern-0.8em\TeX}}}

\ifCLASSOPTIONcompsoc
  \usepackage[nocompress]{cite}
\else
  \usepackage{cite}
\fi
\ifCLASSINFOpdf
\else
\fi
\begin{document}
%
% paper title
% Titles are generally capitalized except for words such as a, an, and, as,
% at, but, by, for, in, nor, of, on, or, the, to and up, which are usually
% not capitalized unless they are the first or last word of the title.
% Linebreaks \\ can be used within to get better formatting as desired.
% Do not put math or special symbols in the title.
\title{Graph Contrastive Learning for Anomaly Detection}

\author{Bo~Chen,
    Jing~Zhang*,
	Xiaokang~Zhang,
	Yuxiao Dong,
	Jian Song,
	Peng~Zhang, 
	Kaibo Xu,\\
	Evgeny Kharlamov, 
	and
	Jie~Tang*, \IEEEmembership{Fellow,~IEEE}	
	\IEEEcompsocitemizethanks{
        \IEEEcompsocthanksitem Bo Chen and Peng Zhang are with Department of Computer Science and Technology, Tsinghua University, Beijing, China, 100084. E-mail: \{cb21, zhangp18\}@mails.tsinghua.edu.cn,
		\IEEEcompsocthanksitem Jing Zhang and Xiaokang Zhang are with Information School, Renmin University of China, Beijing, China. E-mail:\{zhang-jing, zhang2718\}@ruc.edu.cn 
		\IEEEcompsocthanksitem Yuxiao Dong is with Department of Computer Science and Technology, Tsinghua University, Beijing, China, 100084. E-mail: yuxiaod@tsinghua.edu.cn
        \IEEEcompsocthanksitem Jian Song is with Zhipu.AI, Beijing, China. Email: sxusjj@gmail.com.
        \IEEEcompsocthanksitem Kaibo Xu is with Mininglamp Technology, Beijing, China. Email: xukaibo@mininglamp.com.
        \IEEEcompsocthanksitem Evgeny Kharlamov is with Bosch Center for Artificial Intelligence, Renningen, Germany. Email: Evgeny.Kharlamov@de.bosch.com and University of Oslo, Norway. Email: Evgeny.Kharlamov@ifi.uio.no. 
		\IEEEcompsocthanksitem Jie Tang is with Department of Computer Science and Technology, Tsinghua University, and Tsinghua National Laboratory for Information Science and Technology (TNList), Beijing, China, 100084. E-mail: jietang@tsinghua.edu.cn,\\ *Corresponding author
	}% <-this % stops a space
	\thanks{}}

\IEEEtitleabstractindextext{%
\begin{abstract}
	Graph-based anomaly detection has been widely used for detecting malicious activities in real-world applications. 
	Existing attempts to address this problem have thus far focused on structural feature engineering or learning in the binary classification regime. 
	In this work, we propose to leverage graph contrastive learning
	and present the supervised \smodel model for contrasting abnormal nodes with normal ones in terms of their distances to the global context (e.g., the average of all nodes). 
    To handle scenarios with scarce labels, we further enable \smodel as a self-supervised framework by designing a graph corrupting strategy for generating synthetic node labels.
	To achieve the contrastive objective, we design a graph neural network encoder that can infer and further remove suspicious links during message passing, as well as learn the global context of the input graph. 
	We conduct extensive experiments on four public datasets, demonstrating that 
	1) \smodel significantly and consistently outperforms various advanced baselines 
	and 
	2) its self-supervised version without fine-tuning can achieve comparable performance with its fully supervised version. 
\end{abstract}

% Note that keywords are not normally used for peerreview papers.
}

% make the title area
\maketitle

% To allow for easy dual compilation without having to reenter the
% abstract/keywords data, the \IEEEtitleabstractindextext text will
% not be used in maketitle, but will appear (i.e., to be "transported")
% here as \IEEEdisplaynontitleabstractindextext when the compsoc 
% or transmag modes are not selected <OR> if conference mode is selected 
% - because all conference papers position the abstract like regular
% papers do.
\IEEEdisplaynontitleabstractindextext
% \IEEEdisplaynontitleabstractindextext has no effect when using
% compsoc or transmag under a non-conference mode.

% For peer review papers, you can put extra information on the cover
% page as needed:
% \ifCLASSOPTIONpeerreview
% \begin{center} \bfseries EDICS Category: 3-BBND \end{center}
% \fi
%
% For peerreview papers, this IEEEtran command inserts a page break and
% creates the second title. It will be ignored for other modes.
\IEEEpeerreviewmaketitle

%\IEEEraisesectionheading{\input{introduction.tex}}
% Computer Society journal (but not conference!) papers do something unusual
% with the very first section heading (almost always called "Introduction").
% They place it ABOVE the main text! IEEEtran.cls does not automatically do
% this for you, but you can achieve this effect with the provided
% \IEEEraisesectionheading{} command. Note the need to keep any \label that
% is to refer to the section immediately after \section in the above as
% \IEEEraisesectionheading puts \section within a raised box.

% The very first letter is a 2 line initial drop letter followed
% by the rest of the first word in caps (small caps for compsoc).
% 
% form to use if the first word consists of a single letter:
% \IEEEPARstart{A}{demo} file is ....
% 
% form to use if you need the single drop letter followed by
% normal text (unknown if ever used by the IEEE):
% \IEEEPARstart{A}{}demo file is ....
% 
% Some journals put the first two words in caps:
% \IEEEPARstart{T}{his demo} file is ....
% 
% Here we have the typical use of a "T" for an initial drop letter
% and "HIS" in caps to complete the first word.

%\IEEEPARstart{T}{his} demo file is intended to serve as a ``starter file''
%for IEEE Computer Society journal papers produced under \LaTeX\ using
%IEEEtran.cls version 1.8b and later.
% You must have at least 2 lines in the paragraph with the drop letter
% (should never be an issue)
%I wish you the best of success.

%\hfill mds
 
%\hfill August 26, 2015
\input{introduction.tex}

%\input{problem.tex}

\input{approach.tex}

\input{exp.tex}

\input{related.tex}

\input{conclusion.tex}

\input{ack}

\bibliographystyle{abbrv}
\bibliography{reference} 

\ifCLASSOPTIONcompsoc
\vspace{-0.6in}
\begin{IEEEbiography}[{\includegraphics[width=1in,height=1.25in,clip,keepaspectratio]{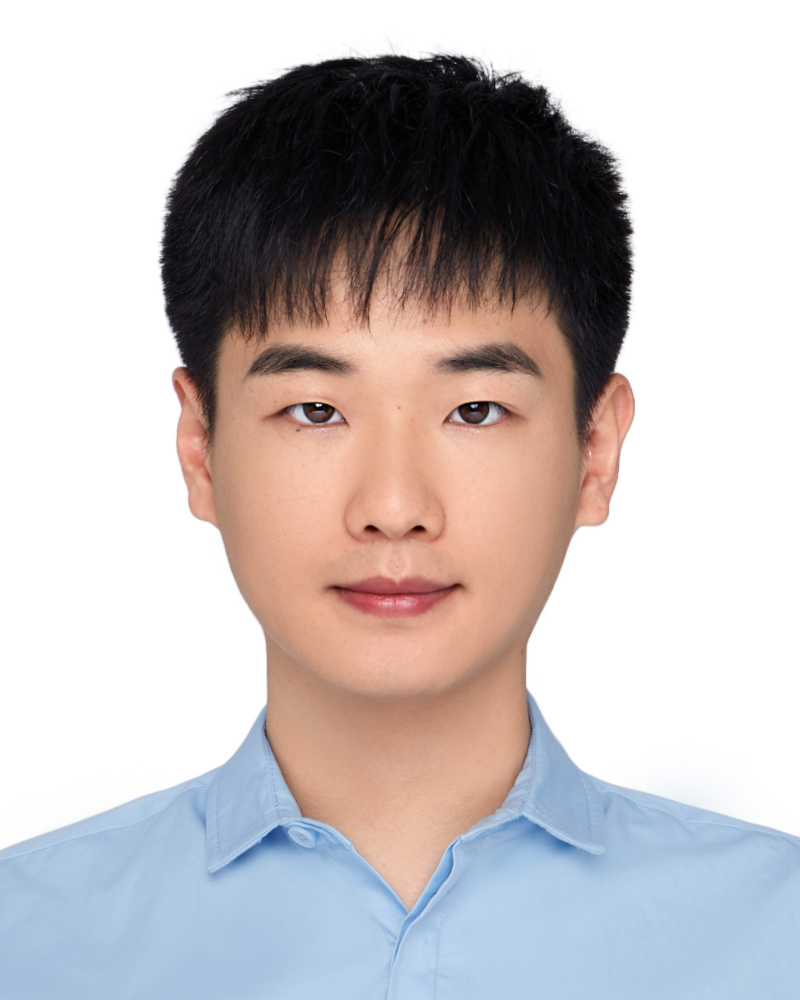}}]{Bo Chen} is a PhD candidate in the Department of Computer Science and Technology, Tsinghua University. He got his master’s degree
from the information school, Renmin University of China. His research interests include data integration and knowledge graph mining. He has published some related papers at the top conferences and journals such as TKDE, AAAI, IJCAI, and so on. 
\end{IEEEbiography}

\vspace{-0.6in}
\begin{IEEEbiography}[{\includegraphics[width=1in,height=1.25in,clip,keepaspectratio]{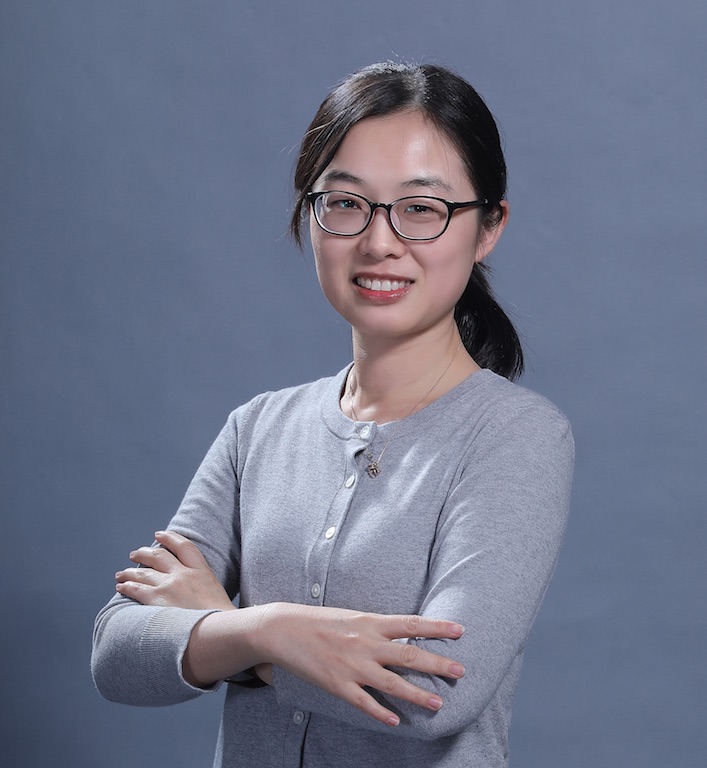}}]{Jing Zhang} is an associate professor at School of Information, Renmin University of China. She received her Ph.D. degree from the Department of Computer Science and Technology in Tsinghua University. She is now interested in knowledge reasoning. She has published more than 50 papers at the top conferences/journals in the area of data mining and artificial intelligence such as KDD, SIGIR, WWW, ACL, TKDE, etc.
\end{IEEEbiography}

\vspace{-0.6in}
\begin{IEEEbiography}[{\includegraphics[width=1in,height=1.25in,clip,keepaspectratio]{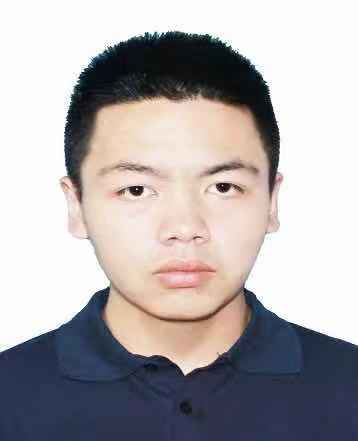}}]{Xiaokang Zhang}is an undergraduate student in Information School, Renmin University of China. His research interests includes knowledge graph mining.
	
\end{IEEEbiography}

\vspace{-0.4in}
\begin{IEEEbiography}[{\includegraphics[width=1in,height=1.25in,clip,keepaspectratio]{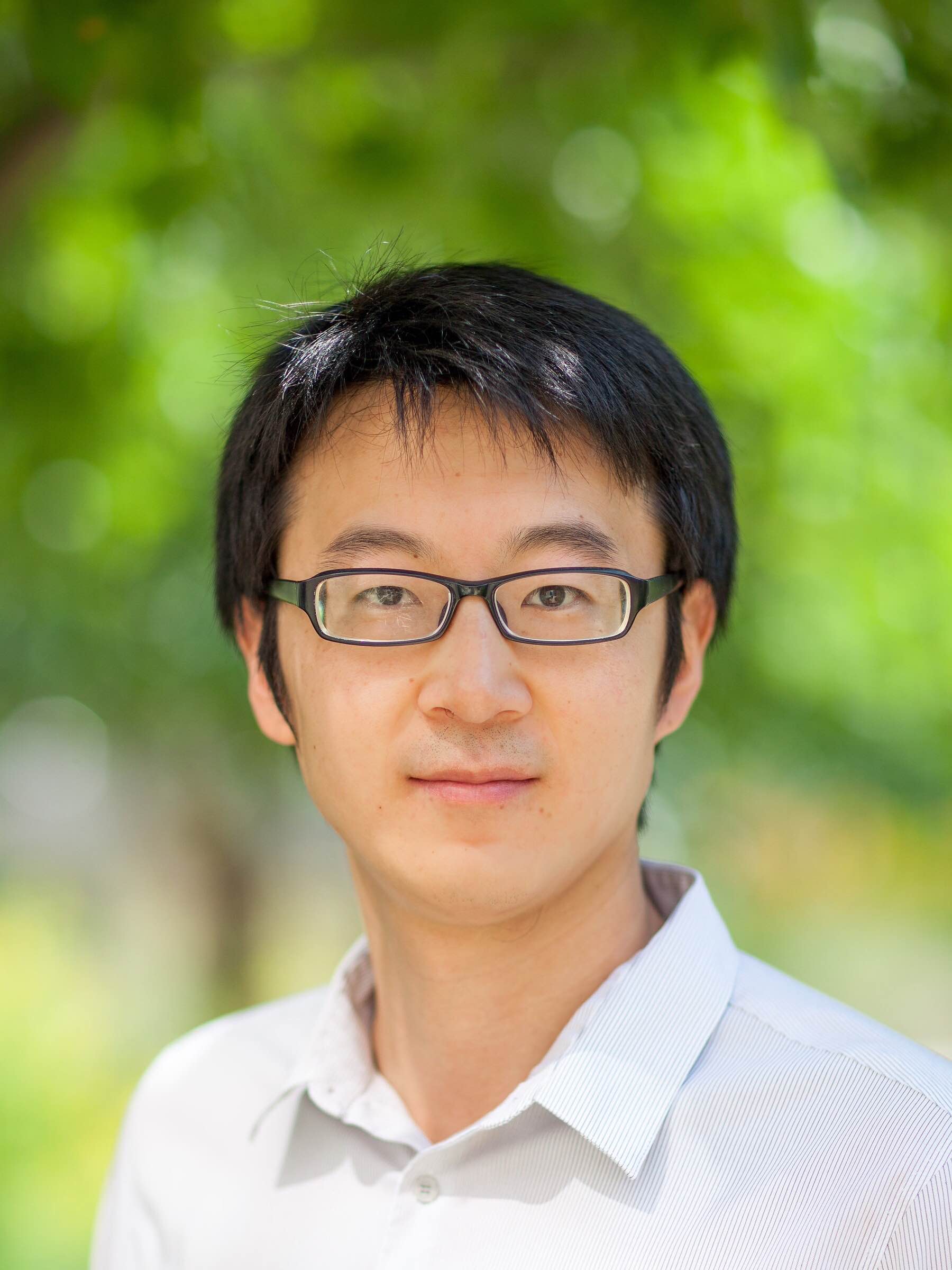}}]{Yuxiao Dong}  
is an assistant professor of computer science at Tsinghua University. His research focuses on data mining, graph representation learning, pre-training models, and social networks, with an emphasis on developing machine learning models to addressing problems in Web-scale systems. 
He received the 2017 SIGKDD Dissertation Award Honorable Mention and 2022 SIGKDD Rising Star Award.
\end{IEEEbiography}

\vspace{-0.4in}
\begin{IEEEbiography}[{\includegraphics[width=1in,height=1.25in,clip,keepaspectratio]{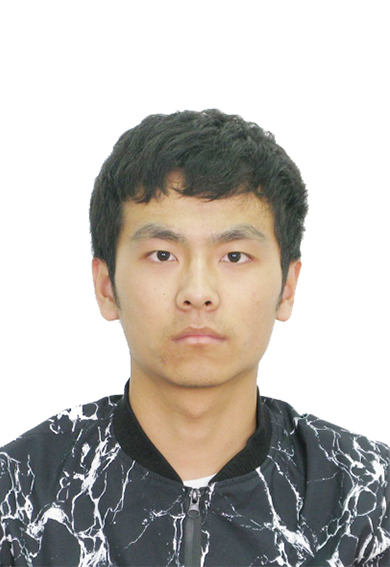}}]{Jian Song} is a research engineer at Zhupu.AI. His mainly work includes author name disambiguation algorithm and paper data processing pipeline.  
\end{IEEEbiography}

\vspace{-0.4in}
\begin{IEEEbiography}[{\includegraphics[width=1in,height=1.25in,clip,keepaspectratio]{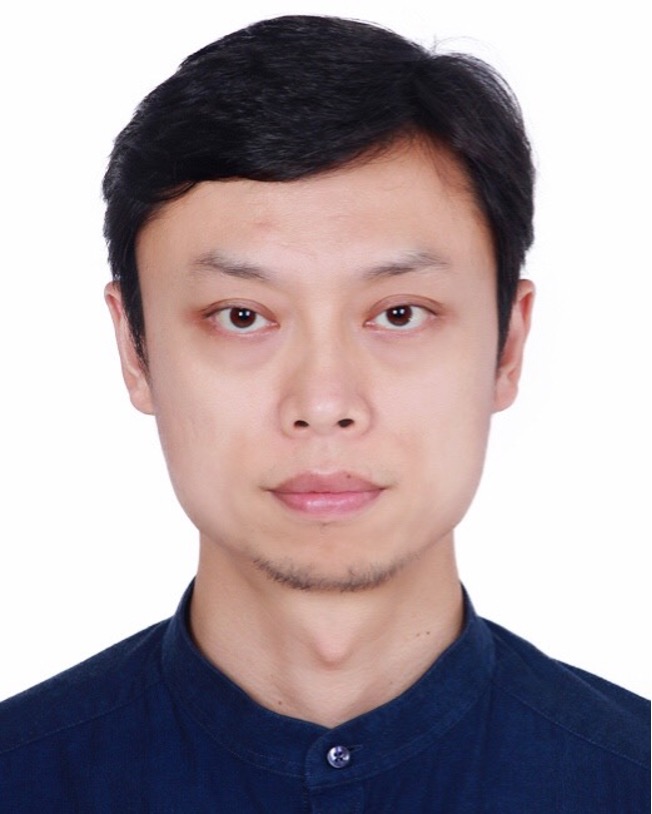}}]{Peng Zhang} is a senior engineer in the Department of Computer Science and Technology, Tsinghua University, and also a Ph.D. of Tsinghua University Innovation Leadership Project. He focused in text mining, knowledge graph construction and application.
	
\end{IEEEbiography}

\vspace{-0.4in}
\begin{IEEEbiography}[{\includegraphics[width=1in,height=1.25in,clip,keepaspectratio]{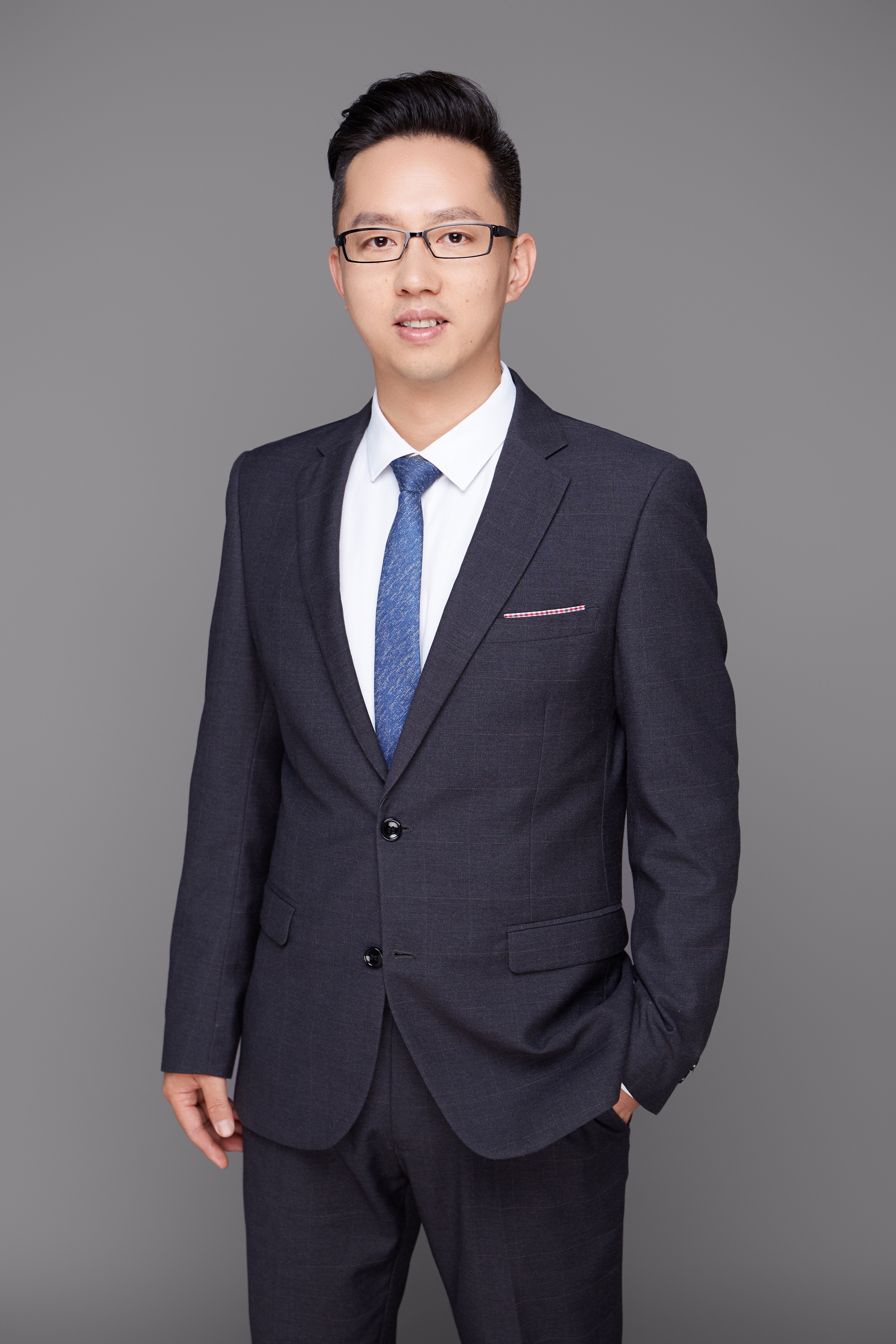}}]{Kaibo Xu} 
%received his Bachelor degree (1998) in Computer Science from Beijing University of Chemical Technology and his Master (2005) and 
received PhD (2010) in Computer Science from the University of the West of Scotland. 
%He worked as a Teaching Assistant (1998-2004), Lecturer (2004-2009), Associate Professor (2009-2017) at Beijing Union University. 
%He has supervised more than 20 master and doctoral students who are successful in their academic and industrial careers. 
As the principal investigator, 
He has received 7 governmental funds and 5 industrial funds with the total amount of 5M in the Chinese dollar. 
% He also consulted extensively and been involved in many industrial projects. 
He worked as the Chief-Information-Officer (CIO) of Yunbai Clothing Retail Group, China (2016-2019). Currently, he is serving as the vice president and principal scientist of MiningLamp Tech. His research interests include graph mining, knowledge graph and knowledge reasoning.
\end{IEEEbiography}

\vspace{-0.4in}
\begin{IEEEbiography}[{\includegraphics[width=1in,height=1.25in,clip,keepaspectratio]{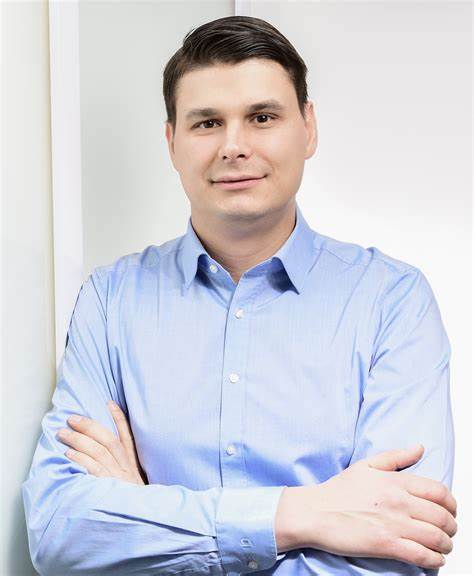}}]{Evgeny Kharlamov} is a Senior Expert at the Bosch Centre for Artificial Intelligence and an Associate Professor at the University of Oslo.
He received his PhD degree in 2011 from the Free University of Bozen-Bolzano in cooperation with INRIA Saclay and Teleocom ParisTech. 
% He worked as a Senior Researcher (from 2013 till 2018) at the University of Oxford and was a visiting researcher at the University of Edinburgh in 2011. 
His research interests are centered around Neuro-Symbolic AI methods that combine Knowledge Graphs and Machine Learning with applications in smart manufacturing. He has published more than 130  papers in major international journals and conferences. 
\end{IEEEbiography}

\vspace{-0.4in}
\begin{IEEEbiography}[{\includegraphics[width=1in,height=1.25in,clip,keepaspectratio]{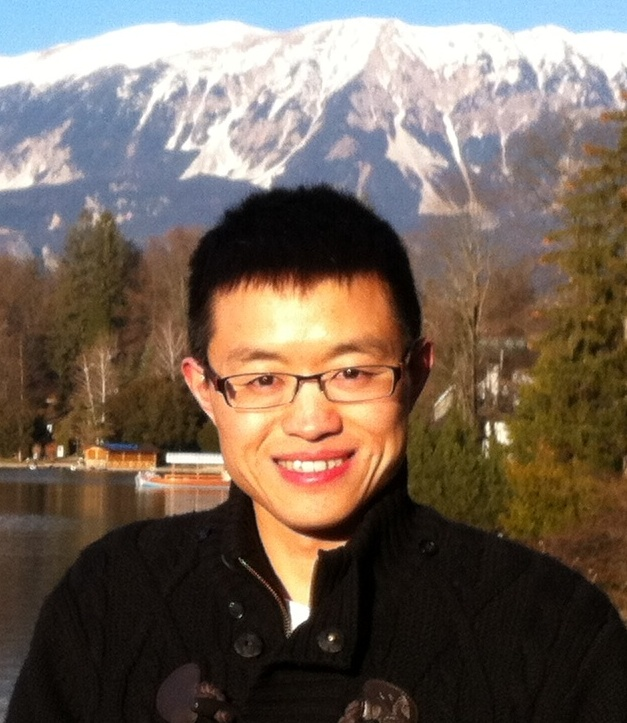}}]{Jie Tang} is a Professor and the Associate Chair of the Department of Computer Science at Tsinghua University. He is a Fellow of ACM and a Fellow of IEEE. His research interests include artificial intelligence, data mining, social networks, and machine learning. He was honored with the SIGKDD Test-of-Time Award, the UK Royal Society-Newton Advanced Fellowship Award, NSFC for Distinguished Young Scholar, and SIGKDD Service Award.
\end{IEEEbiography}

% 
% If you have an EPS/PDF photo (graphicx package needed) extra braces are
% needed around the contents of the optional argument to biography to prevent
% the LaTeX parser from getting confused when it sees the complicated
% \includegraphics command within an optional argument. (You could create
% your own custom macro containing the \includegraphics command to make things
% simpler here.)
%\begin{IEEEbiography}[{\includegraphics[width=1in,height=1.25in,clip,keepaspectratio]{mshell}}]{Michael Shell}
% or if you just want to reserve a space for a photo:

%\begin{IEEEbiography}{Michael Shell}
%Biography text here.
%\end{IEEEbiography}

% if you will not have a photo at all:
%\begin{IEEEbiographynophoto}{John Doe}
%Biography text here.
%\end{IEEEbiographynophoto}

% insert where needed to balance the two columns on the last page with
% biographies
%\newpage

%\begin{IEEEbiographynophoto}{Jane Doe}
%Biography text here.
%\end{IEEEbiographynophoto}

% You can push biographies down or up by placing
% a \vfill before or after them. The appropriate
% use of \vfill depends on what kind of text is
% on the last page and whether or not the columns
% are being equalized.

%\vfill

% Can be used to pull up biographies so that the bottom of the last one
% is flush with the other column.
%\enlargethispage{-5in}

% that's all folks
\end{document}

%% file: introduction.tex
\section{Introduction}
\label{sec:intro}	
\IEEEPARstart{A}{nomaly} detection has profound impacts on preventing malicious activities in various applications, such as detecting online review spams~\cite{nabeel2021cadue}, financial frauds~\cite{SemiGNN,liang2019uncovering}, fake users~\cite{fawcett1996combining}, and misinformation~\cite{cui2020deterrent, silva2021propagation2vec}. 
The most promising development is to utilizing graph structures via machine learning models for distinguishing the anomalies from normal nodes, as graphs can be used for modeling the structural dependencies underlying the data~\cite{kumar2018rev2,liu2020alleviating,dou2020enhancing}. 

%For example, studies have shown that in online review platforms, the structure of the user-item graph plays a critical role in identifying fraudulent users~\cite{kumar2018rev2,liu2020alleviating,dou2020enhancing}. 

%A significant line of works has been focused on discovering and incorporating the structural and attributive patterns of anomalies into learning the classification-based models~\cite{akoglu2015graph, rayana2015collective, kumar2018rev2}.
Recently, the advances of graph neural networks (GNNs)~\cite{kipf2016semi,velivckovic2017graph,Graphsage} have inspired and empowered various attempts to adopt GNNs for detecting anomalies~\cite{liu2020alleviating,dou2020enhancing,shi2022h2,SemiGNN, huang2022auc}. 
%The recent advances of graph neural networks (GNNs), such as GCN~\cite{kipf2016semi}, GAT~\cite{velivckovic2017graph}, GraphSAGE~\cite{hamilton2017inductive}, and GIN~\cite{xu2018powerful}, have promoted the graph-based anomaly detection models. 
%For example, GraphConsis~\cite{liu2020alleviating} and CARE-GNN~\cite{dou2020enhancing} are proposed for review fraud detection, GeniePath~\cite{liu2019geniepath} and SemiGNN~\cite{SemiGNN} are for financial fraud detection, and ASA~\cite{AsAWWW} is for mobile fraud detection. 
The main idea of GNN-based anomaly detection is to leverage the power of GNNs to learn expressive node representations with the goal of distinguishing abnormal nodes from normal ones in the embedding space. Most of the GNN models are based on the inductive bias that two neighboring nodes tend to have the same labels. However, suspicious links between abnormal and normal nodes violate the above assumption, making GNNs produce confusing node embeddings.   

\begin{figure}[t]
	\centering
	%\hspace{-0.2in}
	%\subfigure[AMiner]{\label{subfig:paper_network}
	%	\includegraphics[width=0.33\textwidth]{figures/motivation}
	%}
	\subfigure[t-SNE]{\label{subfig:t-sne}
		\includegraphics[width=0.205\textwidth]{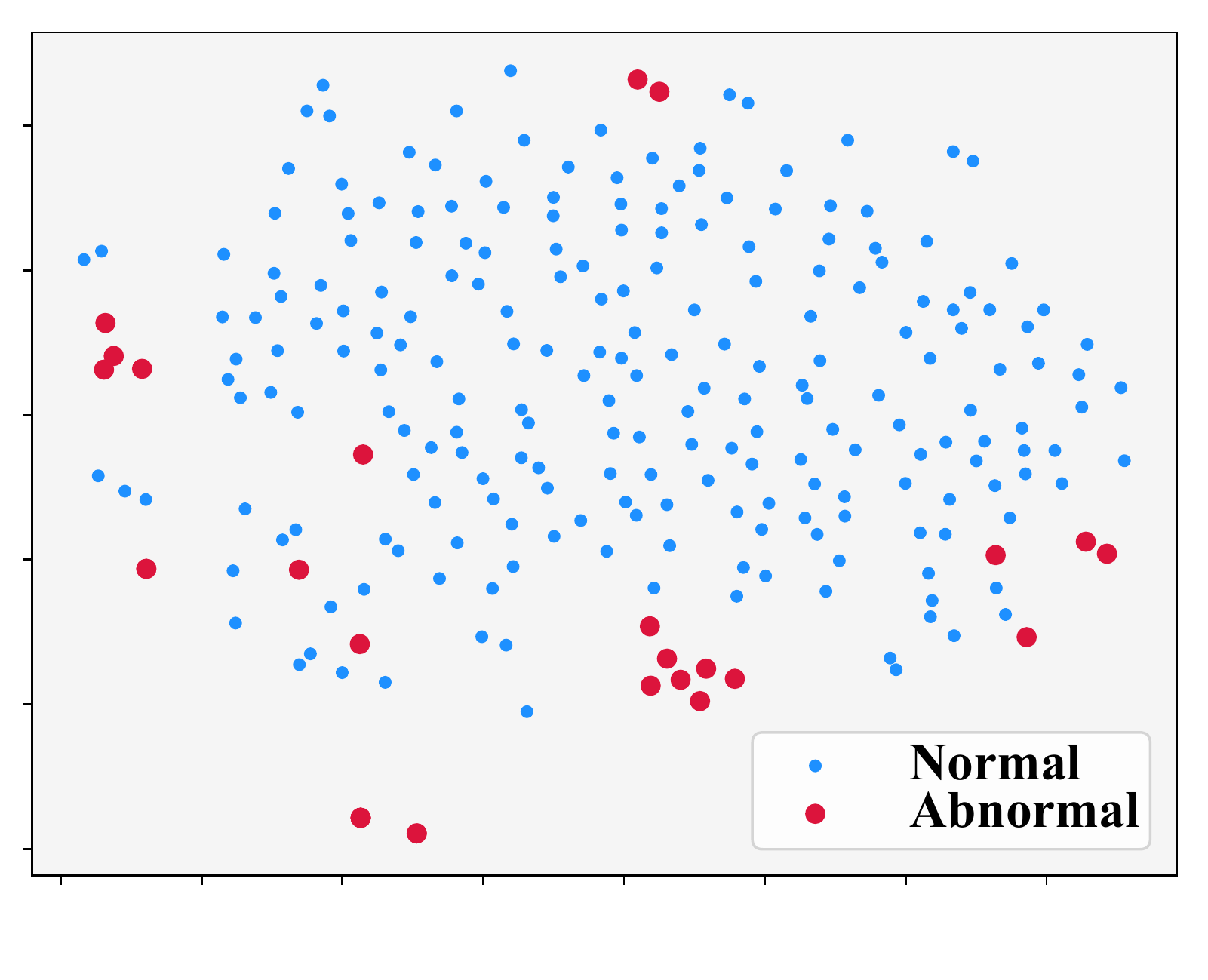}
	}
	\subfigure[Input Features]{\label{subfig:similarity}
		\includegraphics[width=0.21\textwidth,  height=0.121\textheight]{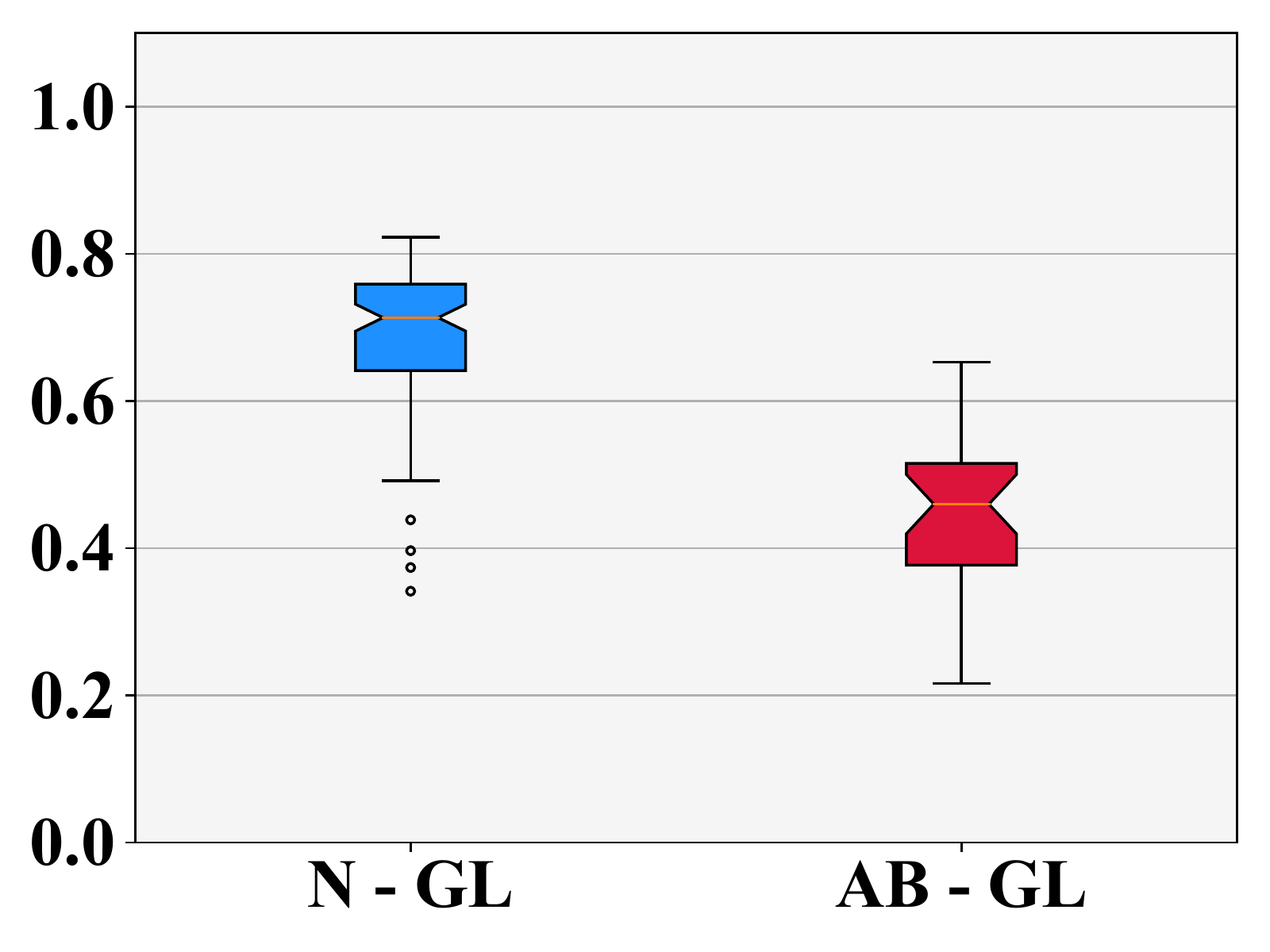}
	}
	%\hspace{-0.9in}
	\subfigure[Embeddings by GCN]{\label{subfig:gcn_sim}
		\includegraphics[width=0.22\textwidth]{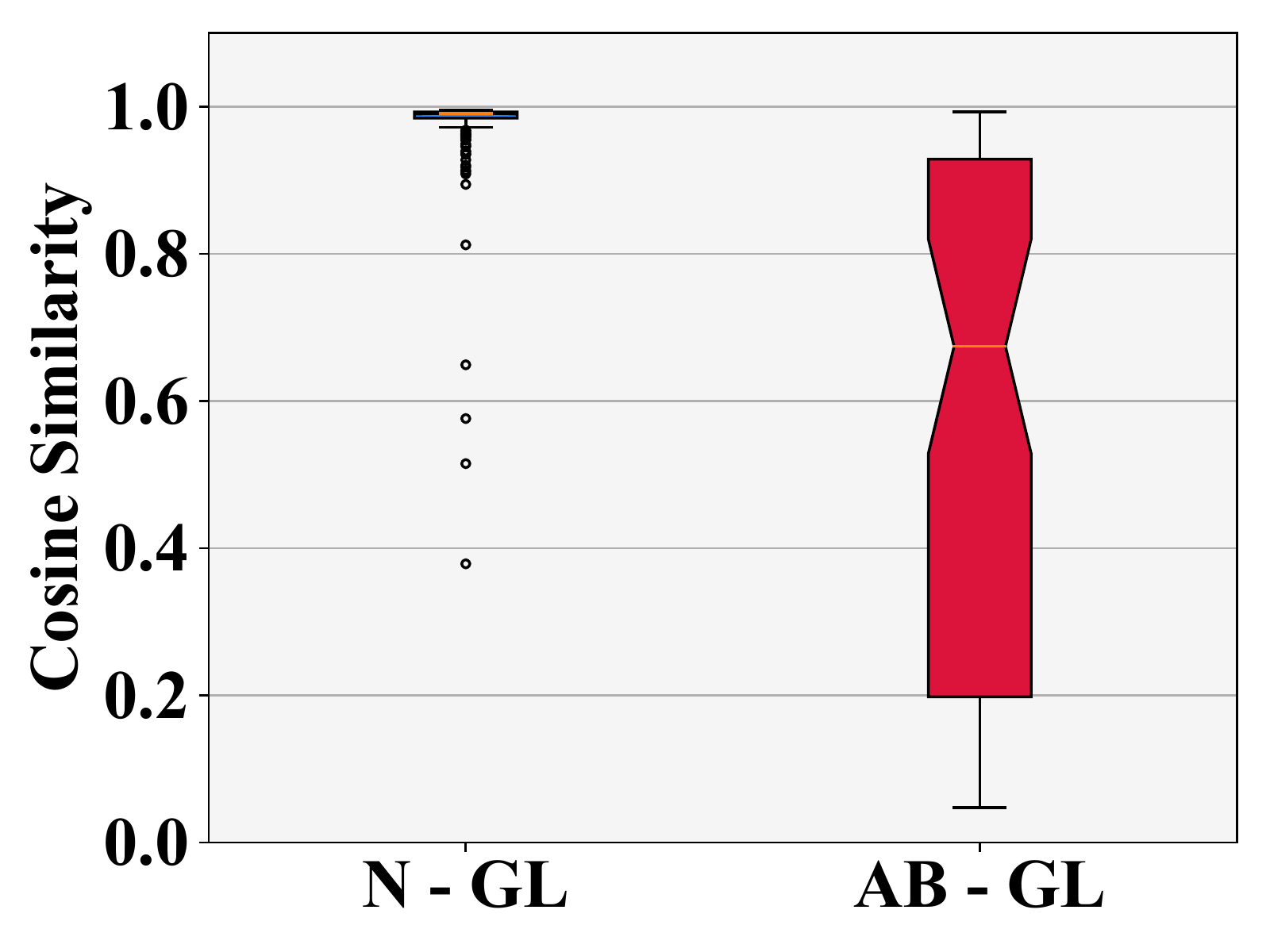}
	}
	\subfigure[Embeddings by \model]{\label{subfig:cogcl_sim}
		\includegraphics[width=0.21\textwidth,  height=0.121\textheight]{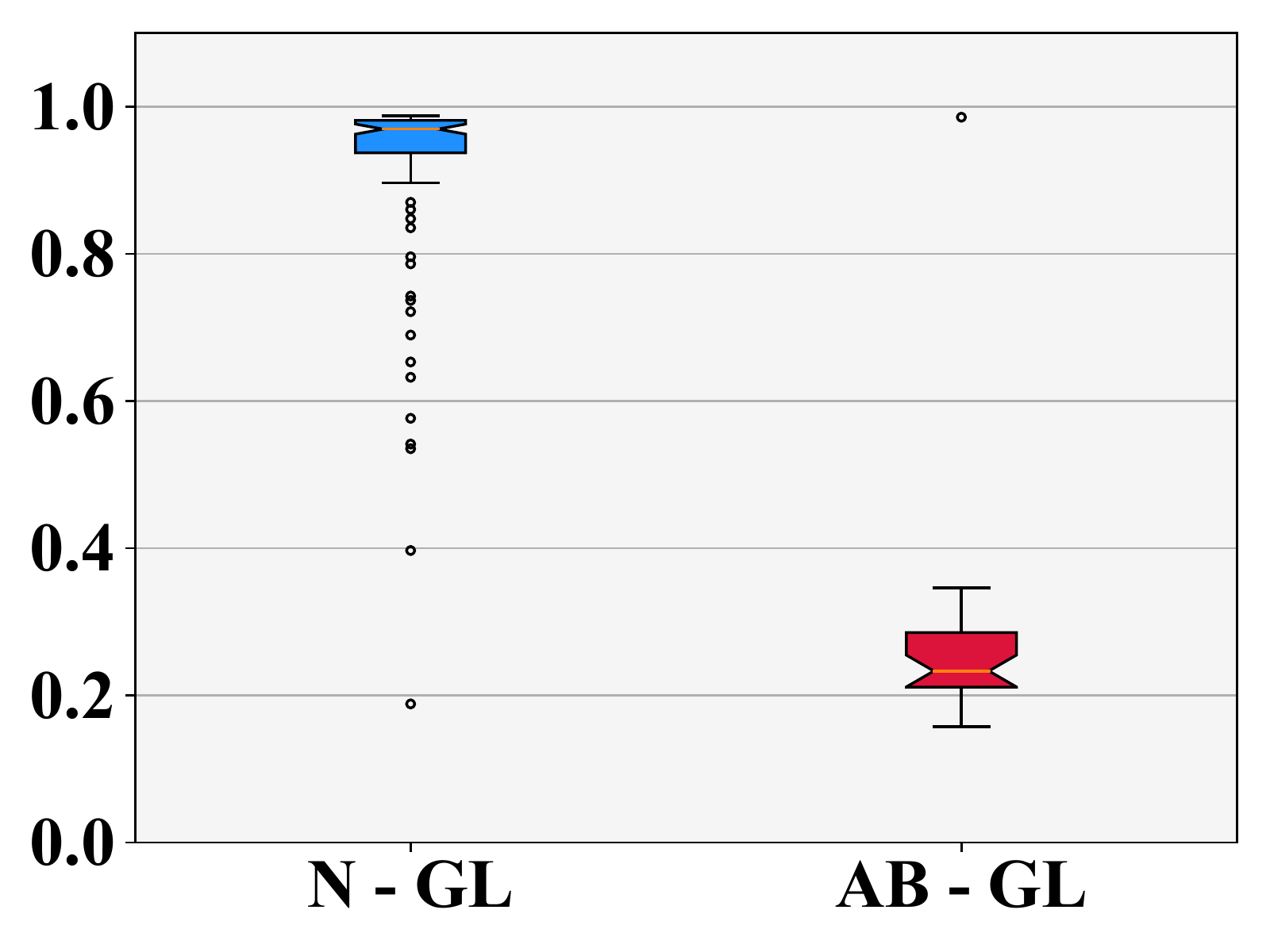}
	}
	\caption{\label{fig:para} A real example of detecting the papers (red) that don't belong to ``Jun Lu". %, a professor from University of Munich. 
		(a) The t-SNE projection of the graph of all his papers, wherein nodes are papers and if two papers share the same coauthors, affiliations, or venues, an edge links them; 
		(b) The similarities between normal nodes and the global context (blue) vs. that between abnormal nodes and the global context (red) by using input features (N, AB, GL are abbreviated for Normal, ABnormal nodes, and GLobal context, respectively);
		(c) The similarities between embeddings generated by GCN. 
		(d) The similarities between embeddings generated by \model.}  
\end{figure}

To further understand the behavior of anomalies, we take the author ``Jun Lu", a professor from Yale University, from the published author profile of Google Scholar as a case study by building a graph with papers assigned to him as nodes, and connect two papers with edges if they share the same coauthors, affiliations, or venues. 
The goal here is to detect wrongly assigned papers (anomalies), i.e., those are not authored by him. 
This anomaly detection problem can be motivated by a news\footnote{https://www.universityworldnews.com/post.php?story=\\20120807160325397} reported in 2012 that another researcher named ``Jun Lu" from Beijing University of Chemical Technology cheated the awards using the papers of ``Jun Lu" from Yale. This event caused by the wrongly assigned papers of the same author name implies the importance of anomaly detection. Thus, we illustrate how the wrongly assigned papers can be distinguished from the right ones in Fig.~\ref{fig:para}.
Fig.~\ref{subfig:t-sne} shows the t-SNE\footnote{https://lvdmaaten.github.io/tsne/} projection of each node's input features---the BERT embedding~\cite{devlin2019bert} of its title---with blue as normal nodes and red as abnormal ones. 
We observe both abnormal and normal nodes are distributed diversely with abnormal ones being relatively more diverse. 
Intuitively, we quantify this observation by computing the similarity between each node and the global context---the average of all node features%\footnote{The global context is later estimated by the memory-based context updater.}
, which is shown in Fig.~\ref{subfig:similarity}. 
It suggests that though having slight overlaps, the two similarity distributions can be clearly distinguished. 
Inspired by these observations, we explore whether there is a straightforward way to capture them for distinguishing abnormal nodes from normal ones.

%The study implies that the distance to the global context is  an important but  often neglected factor for detecting the anomalies. 

%Such diverse feature distributions of the normal and abnormal nodes in a graph makes it challenging to find a clear boundary by binary classification, which drives us to explore a more effective solution for anomaly detection.

\vpara{Present Work.} 
In light of the recent progress in contrastive learning~\cite{wu2018unsupervised,he2020momentum}, we propose to contrast each node with the global context of the input graph. 
The underlying assumption is that abnormal nodes tend to be more distant from the majority of nodes, namely the graph context, than normal ones. We name the model as \model, a %\underline{Co}ntext-aware \underline{G}raph \underline{C}ontrastive \underline{L}earning model. 
Graph Contrastive Learning for Anomaly Detection.
Specifically, we design the context-aware graph contrastive loss in a supervised manner, i.e., labeled normal and abnormal nodes are treated as positive and negative keys, respectively (Cf. Section \ref{subsec:overview}), differing it from most existing studies that use graph contrastive learning in a self-supervised pre-training setup~\cite{you2020graph,velickovic2019deep,qiu2020gcc}. 
Fig.~\ref{subfig:cogcl_sim} plots the similarity distributions of embeddings generated by \smodel compared with that by GCN~\cite{kipf2016semi} shown in Fig.~\ref{subfig:gcn_sim}, demonstrating \model's striking capacity of separating abnormal nodes from normal ones when considering their distances to the global context of the graph.

In addition to the supervised \smodel model, we also extend it in an unsupervised pre-training manner \spmodel for handling cases with scarce labels. 
Straightforwardly, we need to synthesize node labels that can be directly used to replace the ground-truth labels in the supervised contrastive loss. 
% To generate synthetic abnormal nodes, 
To achieve this, we design a strategy to corrupt each part of the original graph by injecting the nodes outside this part.

To achieve the contrastive goal, we propose a context-aware GNN encoder with three modules: \textit{edge update}, \textit{node update}, and \textit{graph update}. 
First, \textit{edge update} is used to estimate the suspicious likelihood of each link and then update the adjacency matrix by removing the most suspicious links. 
Then, \textit{node update} is to update node embeddings by message passing on the updated adjacency matrix. 
Finally, \textit{graph update} is designed to update the global context iteratively.

We verify the proposed model by two genres of anomaly detection tasks, i.e., detecting wrongly assigned papers in researchers' profiles on two academic datasets --- AMiner and MAS, and detecting users who give fraudulent ratings on two business websites --- Alpha and Yelp. The academic datasets have multiple author profiles where each profile can be viewed as a graph, dubbed as the multi-graph setting. While the business datasets, which have one large graph, are viewed as the single-graph setting.
Experiments show that: 1) \smodel yields substantial improvements on multi-graph datasets, while presenting subtle but consistent performance gain on single-graph datasets compared with state-of-the-art baselines;
2) the unsupervised  \spmodel is comparable with the fully-supervised \model. With further fine-tuning, \spmodel can outperform \smodel on most of the datasets. 
In general, \spmodel yields consistently better performance than other compared graph pre-training methods. The main contributions are summarized as follows:

\begin{itemize}[leftmargin=*]
	
	\item We propose the idea of using graph contrastive learning for anomaly detection and present \smodel by designing context-aware graph contrastive objective. 

	\item We design an effective strategy to yield synthetic labels for extending \smodel to unsupervised \pmodel. 

	%We extend \smodel to \pmodel, a pre-training version in an unsupervised manner, by injecting into a graph the anomalies outside the graph. 
	%Surprisingly, \spmodel can achieve comparable performance with the fully-supervised \smodel without any labeled data in the AMiner dataset.
	\item We devise a context-aware GNN encoder via injecting context information to obtain both node and context representations.
	
	\item We conduct experiments, showing the substantial improvements brought by \smodel and \pmodel. 
	%\item We conduct experiments, demonstrating the significant performance advantages brought by \smodel and \pmodel. 

\end{itemize}

%% file: approach.tex
%end of hide%%%%%%%%%%%%%=================================================================

%\subsection{The \smodel Model}

\section{\model}
\label{sec:approach}
% We present \model, the graph contrastive coding model, 
In this section, we first introduce the problem definition of anomaly detection (\secref{sec:problem}), and then conduct the preliminary observations to verify the motivation of \smodel(\secref{sec:problem}). After that, we propose the  learning objective with theoretical guarantees (\secref{subsec:overview}), and further extend the supervised objective to the unsupervised setting (\secref{subsec:self}). 
Finally, we introduce the context-aware GNN encoder of \smodel and \spmodel(\secref{subsec:gnn}).

\subsection{The Studied Problem}
\label{sec:problem}
% We formalize the problem of graph-based anomaly detection. 
%\subsection{Graph-based Anomaly Detection Problem}
We define a graph as $G = \left(V, X, A, Y\right)$, where 
$V$ is the set of $N$ nodes, 
$A \in \mathbb{R}^{N \times N}$ denotes the adjacency matrix,
and ${X} \in \mathbb{R}^{N \times d}$ is the corresponding feature vectors with $\boldsymbol{x}_i \in \mathbb{R}^d$ representing the $d$-dimensional feature vector of node $v_i$.
%Generally, $A$ can carry arbitrary edge features to represent various graph properties such as the unweighted or weighted, the undirected or directed, and the single-relational or multi-relational graphs. 
Without loss of generality, 
%To generalize to various scenarios of anomaly detection, 
we consider $G$ as an undirected and single-relational graph, i.e., $A_{ij}\, \textgreater\, 0$ if there exists an edge between $v_i$ and $v_j$ and $A_{ij} = 0$ otherwise.

\begin{problem} \textbf{Graph-based Anomaly Detection.} Given a graph $G = \left(V, X, A, Y\right)$, $Y$ is the set of node labels with $y_i \in Y$ equals to 1 if $v_i$ is abnormal and 0 otherwise. 
%	The goal is to learn a predictive function $f: V \rightarrow \mathbb{R}^d$ that maps each node into a $d$-dimensional space ($d \ll |V|$). 
	%The function $f$ needs to preserve both the structural information and the input features of the nodes. 
	The goal is to learn a function $g: \mathbb{R}^d \rightarrow \{0,1\}$ to determine whether a given node is abnormal (1) or normal (0). 
\end{problem}

To resolve this problem, most existing GNN-based models directly instantiate $g$ as a binary classifier~\cite{dou2020enhancing,liu2020alleviating}. 
We conduct the following preliminary observations to verify the motivation of the proposed model.
%Inspired by the observations about the distinguished distances to the global context (Figure~\ref{fig:para}), we instead leverage contrastive learning for detecting the anomalies. 

\begin{figure}[t]
	\centering
	\hspace{-0.1in}
	%\subfigure[AMiner]{\label{subfig:paper_network}
	%	\includegraphics[width=0.33\textwidth]{figures/motivation}
	%}
	%\subfigure[t-SNE]{\label{subfig:t-sne}
	%	\includegraphics[width=0.19\textwidth,  height=0.11\textheight]{figures/diverse_tsne}
	%}
%\hspace{-0.15in}
	\subfigure[Input Feature (Distribution)]{\label{subfig:class_pro}
		\includegraphics[width=0.22\textwidth]{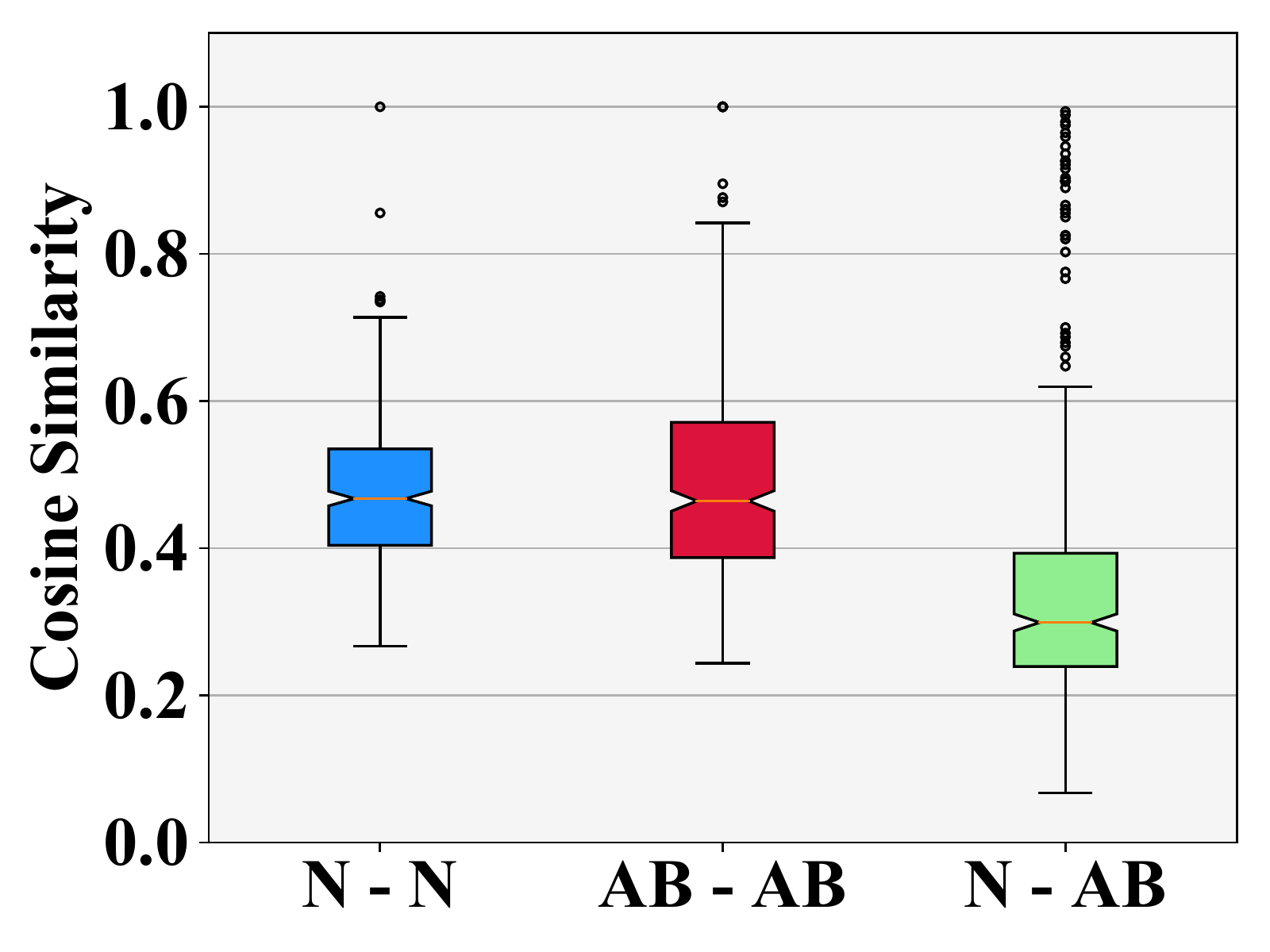}
	}
%\hspace{-0.15in}
	\subfigure[Embeddings by GCN]{\label{subfig:class_after}
	\includegraphics[width=0.21\textwidth,  height=0.121\textheight]{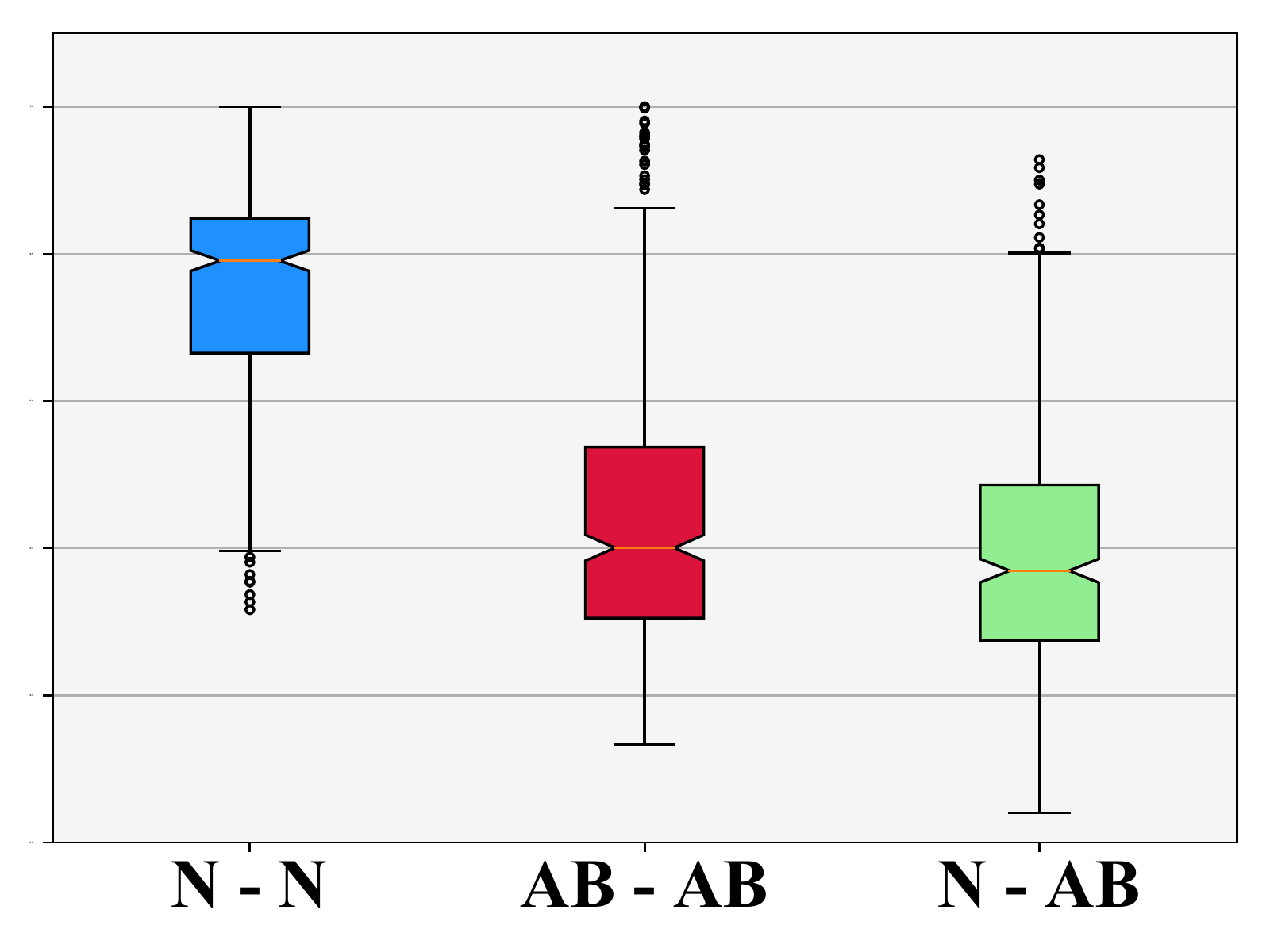}
}
%\hspace{-0.15in}
	\subfigure[Input Feature (Global Cons.)]{\label{subfig:con_pro}
		\includegraphics[width=0.22\textwidth]{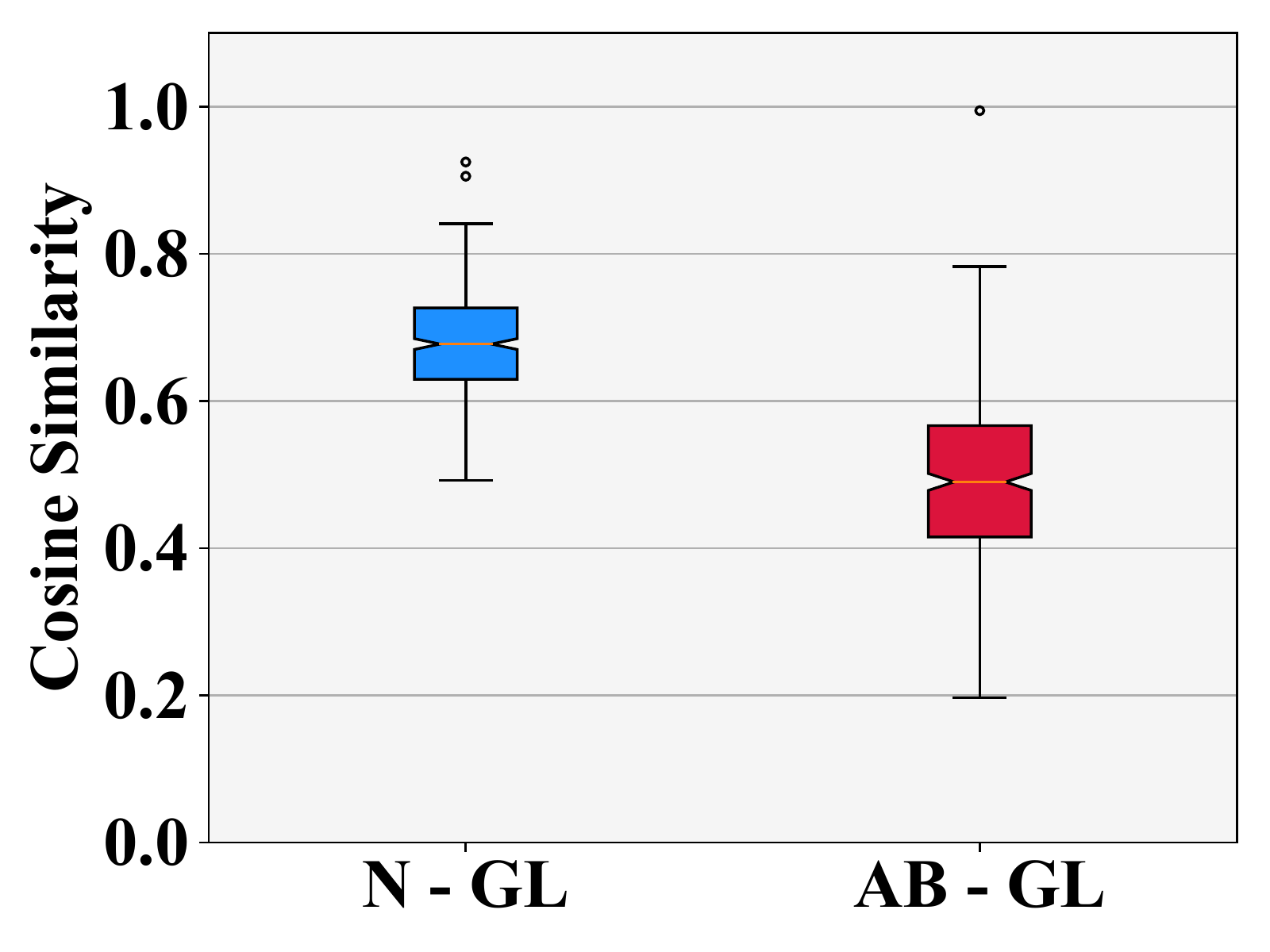}
	}
%\hspace{-0.15in}
	\subfigure[Embeddings by \model]{\label{subfig:con_after}
	\includegraphics[width=0.21\textwidth,  height=0.121\textheight]{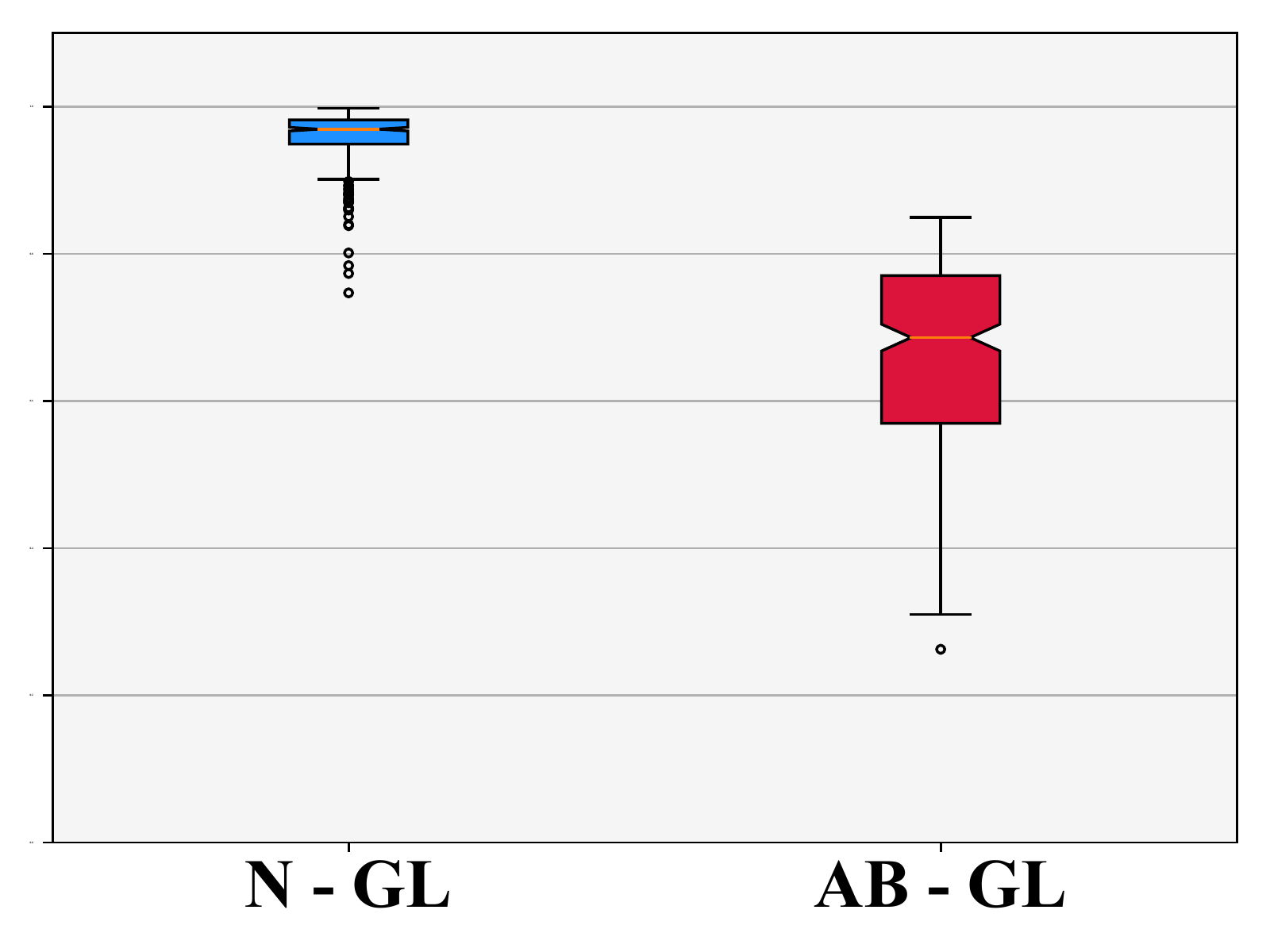}
	}

	\caption{\label{fig:obervation}  %, a professor from University of Munich. 
		%(a) The t-SNE projection of the graph of all his papers, wherein nodes are papers and if two papers share the same coauthors or venues, an edge links them; 
		(a) The similarities between normal and normal nodes (blue), abnormal and abnormal nodes (red), and normal and abnormal nodes (green) by the BERT-initialized input features; 
		%(N and AB are abbreviated for \underline{N}ormal nodes and \underline{AB}normal nodes respectively);
		(b) The similarities between embeddings generated by GCN.
		(c) The similarities between normal nodes and the global context (blue) vs. those between abnormal nodes and the global context (red) by the input features;
		%(GL is abbreviated for \underline{GL}obal context); 
		(d) The similarities between embeddings generated by \model.}  
\end{figure}

\vpara{Preliminary Observations.} To further verify the motivation of contrasting a node with the global context in Fig.~\ref{fig:para}, we additionally extract 10,000 authors owning more than 500,000 papers from AMiner\footnote{https://www.aminer.org/}. Each author profile is treated as a graph with its papers as nodes and connections among papers as edges.
% a free online academic search and mining system. 
Notably, we mainly focus on detecting anomalies in the multi-graph setting in this paper, which is a more challenging and under-explored scenario (Cf. Section 3.2). 
% Thus, here we mainly discuss related cases based on the multi-graph datasets, e.g., AMiner. 
However, our proposed method can also be instantiated as the single-graph setting.
Likewise, the goal here is to detect the wrongly assigned papers (anomalies). For each author, the wrongly assigned papers (anomalies) are labeled by professional annotators or the author themselves. For each paper, we obtain its BERT embedding~\cite{devlin2019bert} by its title. Then we calculate the cosine similarity between a pair of papers and average the pairwise cosine similarities of three groups, i.e.,  \underline{N}ormal and \underline{N}ormal (N-N), \underline{AB}normal and \underline{AB}normal (AB-AB), and \underline{N}ormal and \underline{AB}normal (N-AB). 
Fig.~\ref{subfig:class_pro} shows the following phenomenon.
\begin{itemize}
	\item \textbf{Intra-Diversity:} The similarities in both N-N and AB-AB are extremely diverse (being scattered within [0.2,0.8]), which is in concert with the case in Fig.~\ref{subfig:t-sne}; 
	\item \textbf{Inter-Diversity:} The similarities of N-AB are also diverse, and more than 20\% abnormal nodes are similar to the normal nodes (y $>$ 0.5).
\end{itemize}

%We can see that the similarities distributions in both \textit{N-N} and \textit{AB-AB} are extremely %diverse (Scatter evenly between 0.2 and 1.0). Moreover, from the similarity distribution of %\textit{N-AB}, we find that more than 20\% abnormal nodes are similar with normal nodes (y $>$ 0.5), 

We conjecture this intra-/inter- diversity will degrade the performance of distinguishing anomalies from normal ones by the traditional classifier. To verify this, we investigate the same similarities based on the embeddings generated by GCNs~\cite{kipf2016semi} with the binary classification loss, as shown in 
Fig.~\ref{subfig:class_after}. It demonstrates that although the similarities in N-N increase from [0.2, 0.8] to [0.4, 1.0], the intra- and inter- diversity issues are still severe. 
% which can be attributed into two perspectives: 1) The violation of class-homophily assumption catastrophically degrades the representation capability of GNNs,
% and
% 2) The intrinsic diversity shown in Figure~\ref{subfig:class_pro} incapacitate the classifiers to directly separate the abnormal nodes from normal ones. 
%which empirically implies that resolve the anomaly detection problem in a binary classification fashion may not always be the silver bullets for all the scenarios.

Previous efforts~\cite{liu2020alleviating, kumar2018rev2} disclose the behavior patterns of anomalies are different from those of normal nodes, based on which they characterize the inductive bias to detect anomalies. 
However, most of them only focus on modifying the message passing process to reduce propagated noises, while ignoring the limited capability of the binary classification objective 
when the data distribution is diverse. 
% (Fig.~\ref{subfig:class_pro},~\ref{subfig:class_after}). 
% As the default setting of anomaly detection is the normal nodes are the majority of a whole graph, the behavior pattern deviations between abnormal and normal nodes can further be viewed as that between abnormal nodes and the global context of a graph. 

Inspired by the case observed in Fig.~\ref{subfig:similarity}, we compute the similarity between each node and the global context---the average of all the node features\footnote{The global context is later estimated by the memory-based context updater.}, and show the average similarities in N-GL and AB-GL in Fig.~\ref{subfig:con_pro}. 
We can see that, compared with Fig.~\ref{subfig:class_pro}, although still having overlaps, the two similarity distributions can be distinguished much more clearly. 
% Furthermore, the similarities based on the embeddings of the normal and abnormal ones generated by the proposed \smodel, shown in Fig.~\ref{subfig:con_after}, can be further distinguished compared with Fig.~\ref{subfig:con_pro}.
Furthermore, we plot the similarities based on the embeddings generated by \smodel in Fig.~\ref{subfig:con_after}. Compared with Fig.~\ref{subfig:con_pro}, the resultant embeddings of normal and abnormal ones can be further distinguished.  

To this, we empirically verify the motivation of the context-aware contrastive learning method --- instead of directly capturing the absolute difference between normal and abnormal nodes, contrasting the relative distances between the node and global context is expected to be a more promising way to address the diversity of node distributions.

%\subsection{Context-aware Graph Contrastive Learning}
% \subsection{The \smodel Model}
\subsection{Methodology}
\label{subsec:overview}

The basic idea of \smodel is to determine a node's abnormality by contrasting it with the global context of the entire graph. 
This is motivated by the discovery that there exists a significant difference in the distance to the global context between normal and abnormal nodes. In another words, a node is more abnormal if it deviates farther away from the majority of nodes in the feature space. 
%the anomalous exhibit large deviations from the majority of the nodes than its own features. 
%The underlying assumption is a node is more abnormal if it deviates farther away from the whole context. 

In view of this, we distinguish abnormal and normal nodes in the embedding space by the graph contrastive learning (\model).
%\underline{Co}ntext-aware \underline{G}raph \underline{C}ontrastive \underline{L}earning (\model). 
%with a context-aware graph contrastive objective function. % instead of the binary classification objective function. 
Specifically, given a graph ${G}$, we first create a GNN encoder $f_{\text{GNN}}$ that can output an embedding $\boldsymbol{h}_i$ for each node $v_i$ and also an embedding $\boldsymbol{q}$ for the entire graph (global context), i.e., (${H}, \boldsymbol{q}) = f_{\text{GNN}}({X}, {A}, {W})$ with $H = \{\boldsymbol{h}_i\}_{i=1}^{N}$ and $W$ as the trainable parameters of $f_{\text{GNN}}$. 
We take the graph embedding $\boldsymbol{q}$ as the query, a normal node's embedding as the positive key that matches with $\boldsymbol{q}$, and the embeddings of all the abnormal nodes as the negative keys that don't match with $\boldsymbol{q}$. 
For implementation, we use infoNCE~\cite{oord2018representation} in a \textit{supervised} manner 
%---a well-adopted loss function in contrastive learning---
as the concrete loss function such that:

\beq{
	\label{eq:loss}
	\mathcal{L}_{\text{con}} \!=\!\!\!\! \mathop{\mathbb{E}}\limits_{i: y_i = 0 \atop j:y_j=1}\!\!\left[\!-\!\log \frac{\exp \left(\textbf{q}^\top\boldsymbol{h}_i/\tau\right)}{ \sum_{ j} \exp \left(\boldsymbol{q}^\top \boldsymbol{h}_j/\tau\right) + \exp \left(\boldsymbol{q}^\top\boldsymbol{h}_i/\tau\right) }\!\right]
}

\noindent where $\tau$ is the temperature hyperparameter. The objective function is to enforce maximizing the consistency between the positive pairs (normal node, global context) compared with negative pairs (abnormal node, global context). 

Compared with the context-aware contrastive learning, the traditional normal-abnormal contrast is a node-wise contrastive pattern, which usually suffers from noisy labels. For example, \cite{kalantidis2020hard} states that false-negative instances hinder the performance of contrastive learning. On the contrary, the proposed local-global method contrasts each abnormal node with the global context, which can reduce the negative influence of individual noisy labeled nodes.

\vpara{Why does \smodel work?}
%Fundamentally, the bottleneck of anomaly detection is to differentiate the abnormal behavior patterns from normal ones. Inspired by Quantitative Observations aforementioned, we instead to optimize the relative distance between nodes and global context, rather than charactering the absolute distribution gap between abnormal and normal nodes. 
%We now provide two theoretical guarantees to confirm the efficacy of global context contrastive objective,
%We now provide a theoretical guarantee to confirm the efficacy of global context contrastive objective. 
We theoretically prove the effectiveness of the proposed \smodel compared with the original cross-entropy loss for classification.

\begin{theorem} 
	Let $X^+$\footnote{We denote random variables using upper-case letters (e.g. $X^+$, $X^-$ ), and their realizations by the corresponding lower-case letter (e.g. $x^+$, $x^-$).} denote the random variable of normal node and $p^+(\cdot)$ denote its marginal distribution, thus $X^+\!\sim \!p^+(x^+)$. Likewise, we denote the abnormal node by $X^-$ and its marginal distribution by $p^-(\cdot)$, $X^-\!\sim \!p^-(x^-)$. Assume $p^+$ and $p^-$ are mutually independent.
	Then we have: \textit{minimizing the contrastive loss in Eq. \eqref{eq:loss} forms the lower bound of 
		1) the cross-entropy of the two data distributions $p^+(\cdot)$ and $p^-(\cdot)$ plus 
		2) the entropy of $p^-(\cdot)$.} Formally,  

    \beq{
    \label{eq:theory}
    \min \mathcal{L}_{\text{con}} \triangleq \max \left[H(p^-||p^+) + H(p^-)\right].
    }
\end{theorem}

In view of the two parts in Eq.~\eqref{eq:theory}, the contrastive loss $\mathcal{L}_{\text{con}}$ is theoretically and analytically more robust than the cross-entropy loss and node-wise contrastive loss because of the following reasons.

First, the local-global contrast can better deal with the diverse abnormal distribution. Theorem 1 shows that the proposed local-global contrastive loss can be decomposed into the cross-entropy between the distribution of the normal node and the abnormal node plus the entropy of the abnormal node. Since the first term links to the node-wise contrastive loss~\cite{boudiaf2020unifying}, which is also proved to be the cross-entropy between the predicted labels and the ground truth labels~\cite{boudiaf2020unifying}, the proposed local-global contrastive loss has an additional entropy of the abnormal node compared with the cross-entropy loss. Such additional entropy, acting as a normalization constraint, makes the distribution of learned abnormal node features spread uniformly as much as possible in the feature space. 
Thus it can better deal with the diverse feature distributions, especially the abnormal diverse distribution, which can be exactly observed in the anomaly detection datasets (Cf. Fig.~\ref{fig:para}).

Second, the local-global contrast can reduce the label-imbalance issue. Maximizing the additional entropy of abnormal nodes also augments informative features of the abnormal nodes, reducing the label imbalance issue of the abnormal nodes against the normal nodes to some degree. The common assumption in anomaly detection is that the abnormal node is the minority class, because the number of abnormal nodes is much smaller than normal ones. Thus, traditional anomaly detection methods usually suffer from the class imbalance issue. To alleviate it, data augmentation techniques for generating pseudo-labels of the minority classes [3], up-sampling or down-sampling techniques for re-balancing the class distribution [4], etc., are proposed. The core insight is to augment the information of minority classes, which can balance the training bias between majority classes and minority classes. Such the balancing technique somehow agrees with the optimizing goal of the second part of Eq.~\eqref{eq:theory}, which augments informative features of the abnormal nodes to balance the training bias.

% \vpara{Connection to Cross Entropy.}  Although Cross Entropy loss is equivalent to KL divergence when facing fixed target distributions, $p_{\text{abnor}}(\cdot)$ is a unknown deterministic factor condition on different scenarios (especially with multi-graph datasets in \secref{sec:exp}), KL divergence here is more resilient than Cross Entropy when encountering dynamic node distributions.

% \ipara{Connection to Class Imbalance.} Class imbalance is the Achilles' heel of the anomaly detection problem. Many efforts devoted to adopt self-supervised learning to augment and preserve the information of minority classes~\cite{wei2021crest, yang2020rethinking}, which is in concert with maximizing the entropy of abnormal distribution $H(\boldsymbol{h}^-)$.   
%\end{itemize}

\vpara{Proof of Eq.~\eqref{eq:theory}.} 
Let $\boldsymbol{h}^+$ = $f_{\text{GNN}}\left(x^+, \cdot, W\right)$, where $\boldsymbol{h}^+_i \in \mathbb{R}^d$, then we have the normal node embedding matrix $H^+ = \{\boldsymbol{h}^+_i\}_{i=1}^{n}$, where $n$ is the number of normal nodes. Likewise, we define $\boldsymbol{h}^-$ = $f_{\text{GNN}}\left(x^-, \cdot, W\right)$ and the abnormal node embedding matrix $H^- = \{\boldsymbol{h}^-_j\}_{j=1}^{m}$ with the node number $m$. Note that, $n\!\gg\! m$.
Let $\mathcal{R}(\cdot)$ be a deterministic readout function on graphs. Without loss of generality, we assume $\mathcal{R}(\cdot) = \text{MEAN}(\cdot)$. Thus $\boldsymbol{q} = \mathcal{R}(H) = \frac{1}{N}(\sum_{i=1}^{n}\boldsymbol{h}^+_i + \sum_{j=1}^{m}\boldsymbol{h}^-_j)$. 
From Eq.~\eqref{eq:loss}, we can derive

\begin{small}
\beqn{
\label{eq:lossf}
 \mathcal{L}_{\text{con}} &=& 
\underbrace{\mathop{\mathbb{E}}\limits_{x^+\sim p_{\text{n}}}\left[-\boldsymbol{q}^T\boldsymbol{h}^+/\tau\right]}_{\text{alignment}} \\ \nonumber
&+& \!\!\!\!\!\!\underbrace{\mathop{\mathbb{E}}\limits_{x^+\sim p_{\text{n}},\atop \{x^-_j\}_{j=1}^{m}\sim p_{\text{ab}}}\left[\text{log}\left(e^{\boldsymbol{q}^T\boldsymbol{h}^+/\tau} + \sum_{j}e^{\boldsymbol{q}^T\boldsymbol{h}^-_j/\tau}\right)\right]}_{\text{uniformity}},}
\end{small}

\noindent where the ``alignment" term pulls the distances between the normal nodes and the  global context closer, and the ``uniformity" term pushes the distances between the abnormal nodes and the global context away~\cite{wang2020understanding}. 

Since $\boldsymbol{q} = \mathcal{R}(H) = \frac{1}{N}(\sum_{i=1}^{n}\boldsymbol{h}^+_i + \sum_{j=1}^{m}\boldsymbol{h}^-_j)$, we get
\begin{small}
\beq{
\boldsymbol{q}^T\boldsymbol{h}^+ = \frac{1}{N}\left[\sum_{i=1}^{n}(\boldsymbol{h}^+_i)^T\boldsymbol{h}^+ + \sum_{j=1}^{m}(\boldsymbol{h}^-_j)^T\boldsymbol{h}^+\right].
}
\end{small}

Note that $n\gg m$ and learned by Inter-/Intra- Diversity observations, the similarity scores between normal nodes is predominant and usually large, that is, the term $\boldsymbol{q}^T\boldsymbol{h}^+$ is naturally large, thus the main challenge lies in optimizing the ``uniformity''. Asymptotically, suppose the normal node pairs are perfectly aligned, i.e., $\boldsymbol{q}^T\boldsymbol{h}^+=1$, minimizing Eq.~\eqref{eq:lossf} is equivalent to optimizing the second term, i.e.

\begin{small}
\begin{equation}
\label{eq:second_loss}
\begin{split}
 \mathcal{L}_{\text{con}} &=\!\!\!\!\mathop{\mathbb{E}}\limits_{\{x^-_j\}_{j=1}^{m}\sim p_{\text{ab}}}\left[\text{log}\left(e^{1/\tau} + \sum_{j}e^{\boldsymbol{q}^T\boldsymbol{h}^-_j/\tau}\right)\right]\\ 
% \!\!\!\!&\geq\!\!\!\!\mathop{\mathbb{E}}\limits_{\{x^-_j\}_{j=1}^{m}\sim p_{\text{ab}}}\left[\text{log}\left(m(\frac{\sum_{j}e^{\boldsymbol{q}^T\boldsymbol{h}^-_j/\tau}}{m})\right)\right]\\
\!\!\!\!&\geq\!\!\!\!\mathop{\mathbb{E}}\limits_{\{x^-_j\}_{j=1}^{m}\sim p_{\text{ab}}}\left[\text{log}~e^{\boldsymbol{q}^T\boldsymbol{h}^-_j/\tau}\right] \\ 
% \!\!\!\!&\geq\!\!\!\!\mathop{\mathbb{E}}\limits_{\{x^-_j\}_{j=1}^{m}\sim p_{\text{ab}}}\left[\frac{1}{m}\sum_{j}\left(\boldsymbol{q}^T\boldsymbol{h}^-_j/\tau\right)\right], \\ 
\!\!\!\!&=\!\!\!\!\!\!\! \mathop{\mathbb{E}}\limits_{\{x^-_j\}_{j=1}^{m}\sim p_{\text{ab}}}\!\!\frac{1}{N}\!\!\left[\left(\sum_{i=1}^{n}(\boldsymbol{h}^+_i)^T\boldsymbol{h}^-_j \!\!+\!\! \sum_{k=1}^{m}(\boldsymbol{h}^-_k)^T\boldsymbol{h}^-_j\!\!\right)\!/\tau\!\right] \\ 
% \!\!\!\!&=\!\!\!\!\!\!\! \mathop{\mathbb{E}}\limits_{\{x^-_j\}_{j=1}^{m}\sim p_{\text{ab}}}\!\!\!\!\left[\left(\text{log}~e^{\left[\sum_{i=1}^{n}(\boldsymbol{h}^+_i)^T\boldsymbol{h}^-_j + \sum_{k=1}^{m}(\boldsymbol{h}^-_k)^T\boldsymbol{h}^-_j\right] / N\tau}\right)\!\right] \\ 
% \!\!\!\!&=\!\!\!\!\!\!\! \mathop{\mathbb{E}}\limits_{\{x^-_j\}_{j=1}^{m}\sim p_{\text{ab}}}\!\!\!\!\left[\left(\text{log}~e^{\left[\frac{\sum_{i=1}^{n}(\boldsymbol{h}^+_i)^T\boldsymbol{h}^-_j}{N\tau}\right]} + \text{log}~e^ {\left[\frac{\sum_{k=1}^{m}(\boldsymbol{h}^-_k)^T\boldsymbol{h}^-_j}{N\tau}\right]}\right)\!\right] \\ 
% &\geq \mathop{\mathbb{E}}\limits_{\{x^-_j\}_{j=1}^{m}\sim p_{\text{ab}}}\!\!\frac{1}{N}\left[\min\!\left(\text{log}~\frac{1}{n}\sum_{i=1}^{n}e^{(\boldsymbol{h}^+_i)^T\boldsymbol{h}^-_j/\tau}\! +\! \text{log}~n\right) \right.\\ 
% &+\left.\min\!\left(\text{log}~\frac{1}{m}\sum_{k=1}^{m}e^{(\boldsymbol{h}^-_k)^T\boldsymbol{h}^-_j/\tau} \!+\! \text{log}~m\right)\right] ,
% \!\!\!\!&\geq&\!\!\!\! \mathop{\mathbb{E}}\limits_{x^+\sim p_{\text{n}}}\left(\frac{1}{m}\sum_{j} \left[\text{log}(e^{\frac{1}{N}\left[\sum_{i=1}^{n}(\boldsymbol{h}^+_i)^T\boldsymbol{h}^-_j + \sum_{k=1}^{m}(\boldsymbol{h}^-_k)^T\boldsymbol{h}^-_j\right]/\tau})\right]\right) \\ \nonumber
% \!\!\!\!&\geq&\!\!\!\! \frac{1}{N}\left[\mathop{\mathbb{E}}\limits_{x^+\sim p_{\text{n}}}\!\!\left(\frac{1}{m}\sum_{j}\left[\text{log}~e^{\frac{\sum_{i=1}^{n}(\boldsymbol{h}^+_i)^T\boldsymbol{h}^-_j}{\tau}}\!+\!\text{log}~e^{\frac{\sum_{k=1}^{m}(\boldsymbol{h}^-_k)^T\boldsymbol{h}^-_j}{\tau}}\right]\right)\right] \nonumber
\end{split}
\end{equation}
\end{small}

\noindent where the first inequality follows the Jensen Inequality based on the concavity of the log function, namely $\text{log}(\mathbb{E}[x])\geq \mathbb{E}[\text{log}(x)]$. 

As $p^+$ and $p^-$ are mutually independent and the data samples from either $p^+$ or $p^-$ follow i.i.d. assumptions, minimizing the last equation of Eq.~\eqref{eq:second_loss} is equivalent to minimizing both the sum of similarities between normal and abnormal node
embeddings and the sum of the similarities between abnormal and abnormal node embeddings, i.e.

% \begin{small}
% \begin{equation}
% \label{eq:third_loss}
% \begin{split}
% \min \text{Eq.}~\eqref{eq:second_loss}\!&=\!\min\!\! \left[\!\frac{1}{m}\!\sum_{j}\!\left(\sum_{i=1}^{n}(\boldsymbol{h}^+_i)^T\boldsymbol{h}^-_j \!+\! \sum_{k=1}^{m}(\boldsymbol{h}^-_k)^T\boldsymbol{h}^-_j\!\right)\!/\tau\!\right] \\
% \!\!\!\!&\triangleq\!\!\!\!\! \frac{1}{m}\sum_{j}\left[\min\!\left( \sum_{i=1}^{n}(\boldsymbol{h}^+_i)^T\boldsymbol{h}^-_j\right)/\tau + \right. \\
% &\left.\min\!\left( \sum_{k=1}^{m}(\boldsymbol{h}^-_k)^T\boldsymbol{h}^-_j\right)/\tau\right]\\
% \!\!\!\!&\triangleq\!\!\!\!\!\frac{1}{m}\sum_{j}\left[\min\!\left(\text{log}~\frac{1}{n}\sum_{i=1}^{n}e^{(\boldsymbol{h}^+_i)^T\boldsymbol{h}^-_j/\tau}\! +\! \text{log}~n\right) \right.\\ 
% &+\left.\min\!\left(\text{log}~\frac{1}{m}\sum_{k=1}^{m}e^{(\boldsymbol{h}^-_k)^T\boldsymbol{h}^-_j/\tau} \!+\! \text{log}~m\right)\right] ,
% \end{split}
% \end{equation}
% \end{small}

\begin{small}
\begin{equation}
\label{eq:third_loss}
\begin{split}
\min \text{Eq.}~\eqref{eq:second_loss} &= \frac{1}{N}\cdot \frac{1}{m}\sum_{j}\left[\min\!\left( \sum_{i=1}^{n}(\boldsymbol{h}^+_i)^T\boldsymbol{h}^-_j\right)/\tau + \right. \\
&\left.\min\!\left( \sum_{k=1}^{m}(\boldsymbol{h}^-_k)^T\boldsymbol{h}^-_j\right)/\tau\right]\\
&\triangleq \frac{1}{m}\sum_{j}\left[\min\!\left(\text{log}~\frac{1}{n}\sum_{i=1}^{n}e^{(\boldsymbol{h}^+_i)^T\boldsymbol{h}^-_j/\tau}\! +\! \text{log}~n\right) \right.\\ 
&+\left.\min\!\left(\text{log}~\frac{1}{m}\sum_{k=1}^{m}e^{(\boldsymbol{h}^-_k)^T\boldsymbol{h}^-_j/\tau} \!+\! \text{log}~m\right)\right] \\
\!\!\!\!&\triangleq \min\!\left[-H(p^-, p^+) \!-\!H(p^-)+\text{log}~Z_{vMF}\right]
\end{split}
\end{equation}
\end{small}

\noindent 
% where the expectation symbol in the first equality is omit for brevity,  and 
where the second equality is obtained by first applying exponent to the each of similarity score, and then re-scaling the expectation of the similarity scores in the logarithm form. Notably, the second equation holds true under the minimization optimization circumstance. Moreover, inspired by the entropy estimation operation in~\cite{wang2020understanding}, the last equality can be viewed as a resubstitution entropy estimator of $\boldsymbol{h^{+/-}}$~\cite{ahmad1976nonparametric} via a von Mises-Fisher (vMF) kernel density estimation (KDE), where $\text{log}Z_{vMF}$ is the normalization constant and $\hat{H}$ denote the resubstitution entropy estimator.

The results reveal that minimizing Eq. \eqref{eq:loss} is equivalent to maximizing the cross-entropy between the two data distributions, $p^+(\cdot)$ and $p^-(\cdot)$ and the entropy of $p^-(\cdot)$.

\hide{
\begin{enumerate}
	\item $p_{\text{abnor}}(\cdot)$ is the unknown deterministic factor condition on different scenarios,
	\item Cross entropy here is more brittle than KL divergence when encountering various 
\end{enumerate}

and $p_{\text{abnor}}(\cdot)$ is the unknown deterministic factor condition on different scenarios,

Considering cross entropy is more brittle than KL divergence when encountering various

As $p_{\text{abnor}}(\cdot)$ is the unknown deterministic factor condition on different scenarios, thereby optimizing $D_{KL}(\boldsymbol{h}^-||\boldsymbol{h}^+)$ and $H(\boldsymbol{h}^-)$ is more resilient and superior than cross entropy

the entropy $H(\boldsymbol{h}^-)$ can be as a empirical normalize factor. Thus, minimizing the contrastive loss in Eq. \eqref{eq:loss} is equivalent to maximize the KL divergence between the normal and abnormal distribution, which is a relative criteria superior than charactering the absolute distribution gap between the normal and abnormal nodes.
}

\vpara{Connection to Existing Graph CL Frameworks.}
\model, that performs contrastive learning in a supervised manner, differs from existing graph contrastive learning frameworks for GNN pre-training, such as GCC~\cite{qiu2020gcc}, DGI~\cite{velickovic2019deep}, GraphCL~\cite{you2020graph}, because most of them are self-supervised and the contrastive instances should be constructed from the unlabeled data.  
For example, in GCC~\cite{qiu2020gcc}, given the embedding of a randomly sampled ego-network of a node as the query, the positive key is the embedding of another sampled ego-network of the same node, and the negative keys are those sampled from other nodes. 
In DGI~\cite{velickovic2019deep}, given the entire graph embedding as the query, the positive key is the embedding of a node in it, and the negative keys are those of the same node after permutating the graph. GraphCL~\cite{you2020graph} further contrasts between different graph augmentations. Given the embedding of a graph as the query, the positive key is the embedding of the graph augmented from the same graph, while the negative keys are those from different graphs. 
In contrast, in our objective for anomaly detection, given the graph embedding as the query, the positive and the negative keys are constrained to the normal and abnormal nodes respectively. In light of this, \smodel is a kind of \textit{supervised} contrastive learning~\cite{khosla2020supervised}. 

%Furthermore, different from supervised contrastive learning that discriminates between the features of instances, we inject global context and compare the relative distance to this context between instances to overcome the feature diversity issue. \smodel  combines the advantages of the existing frameworks and can be viewed a \textit{supervised} context-aware contrastive learning framework.

\subsection{\spmodel for Unsupervised Settings}
\label{subsec:self}

Sufficient labels of anomalies are often arduously expensive to obtain, which is the bottleneck of most anomaly detection scenarios. 
%For example, in AMiner\footnote{http://aminer.org}, an online academic system, it usually spends up to several hours to correct the collected papers for a top expert by a skilled annotator. 
%And for annotating the fraudsters on business websites, the labeling criteria on abnormal behaviors is hard to be offered. 
%Such scarce labels drives us to extend \smodel in an unsupervised manner. 
Therefore, we further extend \smodel as an unsupervised pre-training model (\pmodel) to handle real-world applications with scarce labels.

We present a pre-training strategy for tackling the label scarce problem in anomaly detection. 
Inspired by the idea of context-aware contrastive objective, we propose to construct pseudo anomalies via corrupting the original graph. 
Considering one small part of the original graph, we inject nodes outside this part into it.  
% by injecting distant nodes into a local structure. 
%The \smodel on the corrupt graphs. 
The underlying assumption is that as nodes in different parts of the graph follow different distributions~\cite{qiu2020gcc, velickovic2019deep}, they can serve as the pseudo anomalies to the context of the small part of the original graph.   
Formally, a corrupt graph is defined as follows:

\begin{definition}
	\label{eq:corrupt}
	\textbf{Corrupt Graph.} 
	Given a graph $G=(V, X, A)$, we break it into $M$ parts $\{G_i\}_{i=1}^{M}$.  
	%we create an additional adjacency matrix $\mathcal{A}$ between the nodes in all the graphs, where the \textbf{intra} links are from individual graphs and the \textbf{inter} links are created by the same way as the intra links. \yx{xx}
	For each part $G_i$, a corrupt graph $\tilde{G}_i$ is created by injecting a set of nodes $\bar{V}_i $ from the parts except $G_i$, i.e., $\bar{V}_i = \{v_j \in G \setminus  G_i \}$, thus the corrupt nodes $\tilde{V}_i$ are the union of $V_i$ and $\bar{V}_i$, i.e., $\tilde{V}_i = V_i \cup \bar{V}_i$. 
	The corrupt adjacency matrix $\tilde{A}_i$ of $\tilde{G}_i$ is obtained by slicing $A$ using the indexes in $\tilde{V}_i$. 
	%the \textbf{intra} links are from individual graphs and the \textbf{inter} links are created by the same way as the intra links. \yx{xx}
	%The corrupt adjacency matrix is obtained by slicing $\mathcal{A}$ using the indexes in $\tilde{V}_i$. 
	%We mark $\mathcal{V}^+$ to denote the original node set $\{v^+_1, v^+_2, ..., v^+_N\}$. We then inject $\mathcal{M}$ ($\mathcal{M} \ll \mathcal{N}$) noisy nodes from other graphs, dented as $\mathcal{V}^-$, such that we obtain a corrupt graph $\mathcal{G}^{'} = \{\mathcal{V}^{'}, \boldsymbol{ \mathcal{X}^{'}}, \boldsymbol{\mathcal{A^{'}}}, \mathcal{Y}^{'}\}$, where $\mathcal{V}^{'} = \{v^+_1, ..., v^+_N, v^-_1, ..., v^-_M\}$, $\boldsymbol{ \mathcal{X}^{'}}, \boldsymbol{\mathcal{A^{'}}}$ are the corresponding feature and adjacency matrix, $\mathcal{Y}_i = 0$ if $i\in \left[1,\mathcal{N}\right]$ and $\mathcal{Y}_i = 1$ if $i\in \left(\mathcal{N}, \mathcal{N}+\mathcal{M}\right]$. 
\end{definition}

%Specifically, suppose we have collected multiple clean graphs, for each graph, we randomly sample several nodes from other graphs into it as the abnormal nodes and link them to the normal nodes in the original graph by the same criteria of creating the original links. 

%For example, suppose we have collected the paper networks of multiple scholars, for each of which, we inject the others' papers into her own network as the abnormal nodes, and link them to the right papers if they share the same coauthor names. \figref{fig:corrupt} illustrates the process of creating a corrupt graph. \yx{to ask}

\begin{figure}[t]
	\centering
	\includegraphics[width=0.4\textwidth]{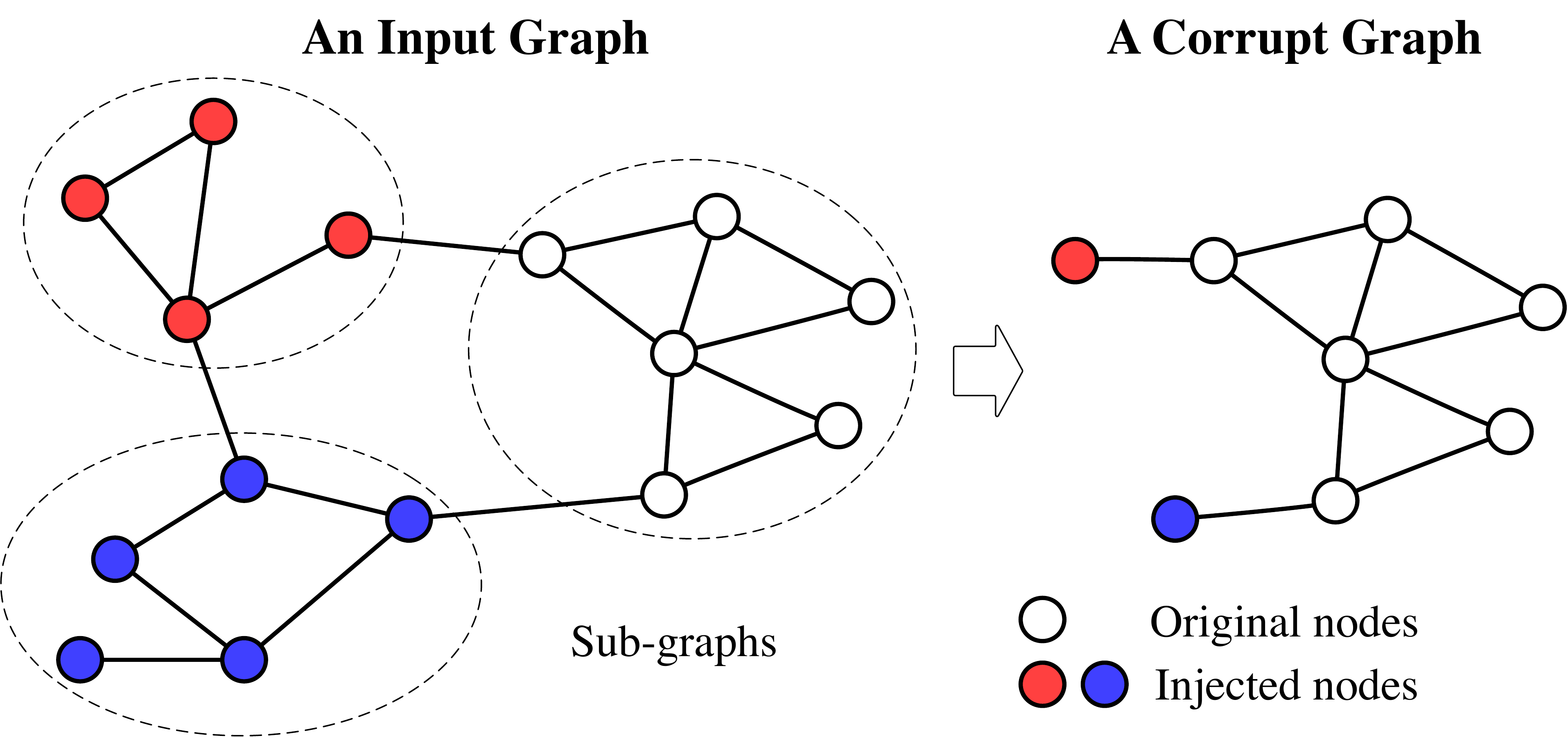}
	\caption{\label{fig:corrupt} Illustration of constructing a corrupt graph. A graph is partitioned into multiple sub-graphs. A corrupt graph is constructed from each sub-graph by injecting the nodes outside it.}
\end{figure}

Fig.~\ref{fig:corrupt} illustrates a toy example of creating corrupt graphs. 
There are different ways for breaking the input graph into multiple parts w.r.t various graph distributions, which can be summarized into two categories:

\begin{itemize}[leftmargin=*]
	\item \textbf{Single-Graph:} If the original graph is a single graph, we partition it into multiple parts using clustering algorithms. 
	%Straightforwardly, we can also take the local structure of each node as one part and build a corresponding corrupt graph. 
	%For the scenario of a single graph, we first cluster the nodes into multiple sub-graphs, and then follow the steps above to construct the pseudo labels. 
	%For example, we cluster a user-rating-product network into multiple sub-graphs. 
	%For each sub-graph, we inject the nodes from other sub-graphs into it as the abnormal nodes, and link them with the normal nodes by the original user-rating-product relationships in the whole graph. 
	%The assumption is users favoring different products can be treated as anomalies with each other. 
	Specifically, we perform the K-means algorithm based on the initial features $X$, and determine the size of the clusters via finding the elbow point of inertia, a well-adopted clustering metric (Cf.~\secref{subsec:pretrain}). 
	\item \textbf{Multi-Graphs:} If the original graph is composed by multiple sub-graphs, without partition, these sub-graphs can be naturally viewed as different parts of the original graph.
	 
\end{itemize}

For the corrupt graphs with injected nodes as pseudo anomalies and original nodes as normal ones, we can directly optimize the context-aware graph contrastive loss defined in Eq. \eqref{eq:loss}, which can be further fine-tuned on the target graph if the ground-truth labels are available.

\vpara{Connection to Graph Pre-Training Frameworks.}
Existing graph pre-training methods offer limited benefits to anomaly detection. 
For example,  GAE~\cite{kipf2016variational} reconstruct the adjacency matrix under the local proximity assumption, which is hurt by suspicious links.
%  connecting nodes with opposite labels.
GCC~\cite{qiu2020gcc} maximizes the consistency between the sampled ego-networks of the same node compared with those sampled from different nodes, such feature consistency assumption cannot explicitly help distinguish the abnormal nodes from the normal ones. 
DGI~\cite{velickovic2019deep}, which contrasts a node with the whole graph, is akin to the contrastive objective of \pmodel, while the negative instances are addressed differently. 
In DGI, given a graph as the context, positive and negative nodes are convolved and represented in independent graphs before contrasting.
But in \pmodel, negative nodes sampled from other graphs are injected and may link to normal nodes in the corrupted graph, which increases the difficulty to distinguish normal and abnormal nodes during training, and thus improves the discrimination ability of the model. 
Essentially, DGI focuses more on representing the normal pattern while \spmodel additionally reinforces the differences of the anomalies from the normal pattern.
%Under the circumstance, benign nodes are those from original graph, yet anomaly nodes are the ones from other different graphs. Hence we can also following the optimization methods motivated by the task of instance discrimination~\cite{wu2018unsupervised} that contrasts the benign nodes and the anomaly nodes from the graph context in an unsupervised contrastive learning manner. The infoNCE loss function~\cite{oord2018representation} is the same as the one defined in section.x.

\begin{figure*}[t]
	\centering
	\includegraphics[width=0.9\textwidth]{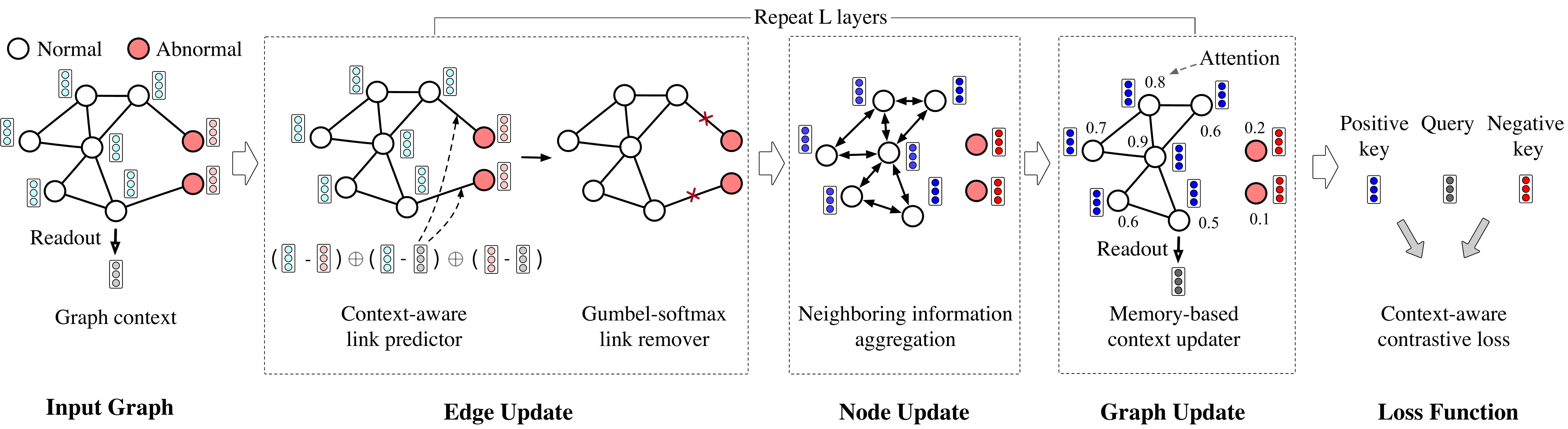}
	\caption{\label{fig:metric_function} The framework of \model. 
		At each layer, \smodel estimates the likelihood for each edge being suspicious and then removes the most suspicious ones with the adjacency matrix updated. 
		Its next step is to update the node embeddings by message passing based on the updated adjacency matrix. 
		Finally it updates the global context. 
		After $L$-layer graph convolutions, the context-aware contrastive loss function is optimized. }
\end{figure*}

\subsection{The GNN Encoder of \model}
\label{subsec:gnn}

\subsubsection{Overview}

\hide{
To optimize the context-aware contrastive objective in Eq. \eqref{eq:loss}, in addition to the message passing and neighbor aggregation process in the general GNN models, we design 
1) an edge update step to remove the suspicious links. and
2) a graph update step to represent the global context 
That is,  \model's GNN encoder $f_{\text{GNN}}$ contains three modules: 
\begin{itemize}[leftmargin=*]
	\item \textbf{Edge Update} is to estimate the suspicious probability of a link, based on which it refines the adjacent matrix at the beginning of each convolution layer, i.e., 
	\beq{
		\label{eq:edge_update}
		{A}^{(l)} =f_{\text{edge}}\left({H}^{(l-1)}, {A}^{(l-1)}, \boldsymbol{q}^{(l-1)}\right).
	}
	\item \textbf{Node Update} is to aggregate the neighboring information based on the refined adjacent matrix ${A}^{(l)}$ and combine the aggregated result with the concerned node, i.e.,
	\beq{
		\label{eq:node_update}
		H^{(l)} = f_{\text{node}}\left(H^{(l-1)}, A^{(l)}\right).
	}
	\item \textbf{Graph Update} is to update the context of the whole graph based on the updated node embeddings $H^{(l)}$ and last-layer's graph embedding $\boldsymbol{q}^{(l-1)}$ via a readout function:  
	
	\beq{
		\label{eq:graph_udpate}
		\boldsymbol{q}^{(l)} = f_{\text{graph}}({H}^{(l)}, \boldsymbol{q}^{(l-1)}). 
	}
\end{itemize}
}

To optimize the context-aware contrastive objective in Eq. \eqref{eq:loss}, we design the edge update, node update, and graph update modules in the GNN encoder of \model. 
First, Edge update is to estimate the likelihood of each link being a suspicious one and then remove the suspicious ones. % via \textit{edge update}; 
Then, node update is to update node embeddings by message passing on the updated adjacency matrix. 
Finally, the global context is updated via graph update. 
The three modules are executed at each graph convolution layer successively. After $L$-layer forwards, the context-aware contrastive objective is optimized. 
The framework is illustrated in Fig. \ref{fig:metric_function}.

%All existing GCN-based methods are aim at decreasing the rotten influences from anomalies with their corresponding implementing inductive bias~\cite{battaglia2018relational}. However they either suffer from limited capabilities or too rigid to generalize in different circumstances.

%In summary, we first create a new adjacency matrix $\boldsymbol{{A}}^{(l)}$ via a link predictor $f_{\text{link}}$, which estimates the suspicious likelihood of links based on the node embeddings $H^{(l-1)}$, the adjacency matrix ${A}^{(l-1)}$, and the graph context embedding $\textbf{q}^{(l-1)}$ of the last layer. Then, the node update function $f_{\text{node}}$ can be adopted to aggregate the neighboring information $H^{l-1}$ based on the adjacency matrix ${A}^{(l)}$. Finally, a readout function $f_{\text{graph}}$ is to update the graph embedding $\boldsymbol{q}^{(l)}$ via comparing the graph context $\boldsymbol{q}^{(l-1)}$ of the last layer with the updated node embeddings $H^{(l)}$. We explain the concrete design of each step as below.

\subsubsection{Edge Update} 
\hide{
We define suspicious links as the ones that connect nodes with opposite labels. 
%For example, although a paper doesn't belong to the a researcher, it may still be linked to the right papers by the same coauthor names. 
%Similarly, almost every fraudster in business websites can be linked to some benign users by the commonly rated products. 
%For example, almost every fraudster in business websites can be linked to some benign users by the commonly rated products. 
Such suspicious links violate the homophily assumption between neighbors---two neighboring nodes tend to have similar features and same labels---and impact the performance of the traditional message-passing along with the links.
}
\label{subsec:edge}
We define suspicious links as the ones that connect nodes with opposite labels. For example, almost every fraudster in business websites can be linked to some benign users by the commonly rated products. Such suspicious links violate the homophily assumption between neighbors—two neighboring nodes tend to have similar features and same labels—and impact the performance of the traditional message-passing along with the suspicious links.

Although many attempts have been made to detect such suspicious links~\cite{zhang2019key, dou2020enhancing}, 
%by attention mechanisms~\cite{wang2019semi, zhang2019key} or reinforcement learning~\cite{dou2020enhancing}, 
they merely predict the likelihood based on the linked node features regardless of the node labels. 
%In another word, they predict a high linkage likelihood for two nodes if their features are similar to each other.
%However, the links between the nodes with same labels but dissimilar features are regular and should be kept. 
Differently, we additionally model the relative distance between nodes and the global context $\boldsymbol{h}-\boldsymbol{q}$. 
The promise here is the distance can serve as implicit labels to guide the model to remove suspicious links --- those connected by the nodes with distinguished distance to the global context.
%For example, two papers of a scholar may have different topics but be linked for a few shared coauthors. 
%Reversely, two users with different preferences may also be linked as they occasionally rated the same product. 
%These links can facilitate the graph convolution of the normal nodes, smoothing their features, and benefiting the encoding of the global context. 
%\yx{to ask jing}

%Such regular links will be removed if predicting links only depending on the node features regardless of the node labels.
%Although the ground-truth labels cannot be used as features in order to avoid information leakage, we can adopt the relative distance to the global context to imply the nodes labels. 

%\yx{what is context aware in one or two sentences?}
\hide{
Thus, we propose 1) a context-aware link predictor to estimate the suspicious linkage likelihood between two nodes 
and 
2) adopt a binary Gumbel-softmax reparameterization trick~\cite{maddison2016concrete, jang2016categorical} to remove the suspicious links. 
3) perform a edge residual to smooth evolution of the edge updating.
}
\vpara{Context-Aware Link Predictor.}
The link predictor accepts the representations of two linked nodes $\boldsymbol{h}_i^{(l-1)}$ and $\boldsymbol{h}_j^{(l-1)}$ as the input, and then estimates the linkage likelihood between node $v_i$ and $v_j$ by:

\beqn{
	\label{eq:linkage_likelihood}
	p_{ij}^{(l)} = \text{MLP}\left((\boldsymbol{h}_i^{(l-1)} - \boldsymbol{h}_j^{(l-1)}) \oplus (\boldsymbol{h}_i^{(l-1)} - \boldsymbol{q}^{(l-1)})\right. \\
	\left.\oplus (\boldsymbol{h}_j^{(l-1)} - \boldsymbol{q}^{(l-1)})\right) , \nonumber
}

\noindent where $\oplus$ is the concatenation operator and MLP is a projection layer with a sigmoid activation function to normalize the likelihood into [0,1]. 
$(\boldsymbol{h}_i^{(l-1)} - \boldsymbol{h}_j^{(l-1)})$ explicitly denotes the similarity of two nodes,
while $(\boldsymbol{h}_{i}^{(l-1)} - \boldsymbol{q}^{(l-1)})$, the distance from $v_{i}$ to the global context, implicitly indicates that two nodes---even if their local features are slightly different---can be highly probably linked  given their relative positions to the global context are similar. The ablation studies in Section 3.2 empirically prove the effectiveness of this context-aware link predictor.

To accelerate the optimization of the link predictor, we optimize the prediction loss in addition to the contrastive objective in Eq.~\eqref{eq:loss}. Specifically, we treat links between normal nodes as positive ones and suspicious links as negative ones. The prediction loss is defined as

\beq{
\label{eq:constraint}
\mathcal{L}_{\text{link}} = \mathop{\mathbb{E}}\left[\sum_{i,j: y_i = y_j = 0} \!\!\!\!\!-\log p_{ij}^{(l)} - \!\!\! \sum_{\substack{i,j: y_i \neq y_j = 0} } \!\!\!\!\!\left(1-\log p_{ij}^{(l)}\right)\right],
}

\noindent which acts as the constraint to directly maximize probabilities of normal links and minimize these of suspicious ones.

\vpara{Gumbel-softmax Link Remover.}
%The linkage likelihood $p_{ij}^{(l)}$ can be injected by attention mechanisms to distinguish the weights of different neighbors in the node update step (Eq.\eqref{eq:node_update})~\cite{wang2019semi, zhang2019key}. However, the suspicious links are not thoroughly removed and the negative influence is still propagated during message passing. Some other researchers attempt to remove the suspicious links before graph convolution~\cite{franceschi2019learning, liu2020alleviating}. Since the link removing and the graph convolution are decoupled, the gradient cannot be passed to the link predictor during backward procedure. 
We remove the suspicious links to reduce their negative influence thoroughly by employing the Bernoulli approximation, the binary case of the Gumbel-softmax reparameterization trick~\cite{maddison2016concrete, jang2016categorical} to resolve the discrete non-differentiable problem.
%deal with the gradient vanishing issue caused by discrete sampling actions. 
Specifically, we define the indicator matrix $I \in \mathbb{R}^{N \times N}$, and for the link likelihood $p_{ij}^{(l)}$, 
we decide whether to keep the link $(I^{(l)}_{ij} = 1)$ or not $(I^{(l)}_{ij}=0)$ by the following equation: 

% \beq{
% 	\label{eq:GUMBEL}
% 	I^{(l)}_{ij} = \left \lfloor \frac{1}{1 + \exp \left(-(\log p_{ij}^{(l)}+ \varepsilon)/\lambda \right)} + \frac{1}{2} \right \rfloor,
% }

\beq{
	\label{eq:GUMBEL}
	I^{(l)}_{ij} = \text{Bernoulli}\left[\frac{1}{1 + \exp \left(-(\log p_{ij}^{(l)}+ \varepsilon)/\lambda \right)}\right],
}

\noindent where $\text{Bernoulli (p)}$ is the Bernoulli approximation with the probability $\text{p}$ to keep the link, $\lambda$ is the temperature hyper-parameter to control the sharpness of the sampling process (the lower $\lambda$ is, the more links will be kept), 
and $\varepsilon \sim \text{Gumbel}~(0,1)$ is the Gumbel noise for reparameterization.
In the forward process, we sample according to Eq.~\eqref{eq:GUMBEL} to obtain $I^{(l)}_{ij}$. 
In the backward process, a straight-through gradient estimator~\cite{bengio2013estimating} is employed to pass the gradients to the relaxed value $p_{ij}^{(l)}$ instead of the discrete value ${I}^{(l)}_{ij}$. %Practically, we adopt the tools provided by pyro, a universal probabilistic programming language\footnote{https://pyro.ai/}. 

 %To enable the discrete sampling of the links, we further 
%Then we adopt  the straight-through gradient estimator~\cite{bengio2013estimating} to round $\tilde{p}^{(l)}_{ij}$ as 1 if it is larger than 0.5 and 0 otherwise and assign the discrete value to ${A}_{ij}^{(l)}$ for the forward computation, 
%According to straight-through gradient estimator~\cite{bengio2013estimating}, we pass the gradients to the relaxed value $p_{ij}^{(l)}$  instead of the discrete value ${A}^{(l)}_{ij}$ during the backward procedure. 

\vpara{Edge Residual.} Recently, RealFormer~\cite{he2020realformer} verifies the effect of the residual connection of attention scores. Inspired by this, we perform the residual connection of the original edge likelihood and the estimated edge likelihood to smooth the edge evolution as follows:

\beq{
A_{ij}^{(l)} = \left(\alpha A_{ij}^{(l-1)} + (1-\alpha) p_{ij}^{(l)}\right) \odot I^{(l)}_{ij},
}

\noindent where $\alpha$ is a learnable parameter to balance the weight from the last layer and the estimated weight $p_{ij}$, and $\odot$ denotes the Hadamard product. 
%which rounds the relaxed samples in Eq.\eqref{eq:GUMBEL} during the forward procedure, 

\hide{
\beqn{
	\label{eq:round}
	{A}^{(l)}_{ij}  &=& \left\{\begin{array}{cl} 
		1 & \tilde{p}^{(l)}_{ij} \geq 0.5;   \\ 
		0 & \mbox{otherwise,}   
	\end{array}\right.  \\  \nonumber
}
}

%\noindent but passes the gradients to the relaxed samples in Eq.\eqref{eq:GUMBEL} rather than the rounded values in  Eq.\eqref{eq:round} during the backward procedure.

\subsubsection{Node Update}
Updating node embeddings can be divided into an AGGREGATION and a COMBINE operation:

\beqn{
	\label{eq:aggregation}
	\boldsymbol{a}_i^{(l)}\!\!\! &=& \!\!\!\text{AGGREGATION}(\{\boldsymbol{h}_{j}^{(l-1)}: A_{ij}^{(l)} > 0\}), \\
	\label{eq:combination}
	\boldsymbol{h}_{i}^{(l)} \!\!\!&=&\!\!\! \text{COMBINE}(\boldsymbol{h}_i^{(l-1)}, \boldsymbol{a}_i^{(l)}),
}

\noindent where Eq. \eqref{eq:aggregation} is to aggregate the neighboring messages of the last layer based on the updated adjacency matrix and Eq.\eqref{eq:combination} is to combine the aggregated information with the concerned node. Generally, the two operations are flexible enough to fit any GNNs. Experiments in \secref{sec:exp} have investigated several concrete functions.

%The AGGREGATION operation is the sum, the same as GIN~\cite{xu2018powerful}, and the COMBINE operation is concatenation. 

\subsubsection{Graph Update}
%The graph embedding $\boldsymbol{q}$, i.e., the global context, needs to be updated at each layer after all nodes' embeddings are updated. 
Straightforwardly, we can perform sum, max, or average pooling of all the nodes to update the global context. However, they do not distinguish the normal and abnormal nodes, which results in an inaccurate global context. To alleviate it, we introduce a memory buffer $\boldsymbol{m}$ to register the global context $\boldsymbol{q}^{(l-1)}$ of the last layer, based on which we calculate the contribution of each node to the global context. The assumption is a node that is deviated from the current global context should be weakened in the next update step. Such the memory-based mechanism has been successfully applied to general computation machines~\cite{bahdanau2015neural, weston2014memory}. Formally,

\beqn{
	\label{eq:readout}
	&&\!\!\!\!\!\!\!\!\!\!\!\!\!\! s_i^{(l)} = \text{cosine}\left( \boldsymbol{h}_i^{(l)}, \boldsymbol{m}\right), \quad \alpha_i^{(l)} = \frac{\exp (s_i^{(l)})}{\sum_{j=1}^{N}\exp(s_j^{(l)})}, \\ \nonumber
	&& \!\!\!\!\!\! \boldsymbol{q}^{(l)} = \sum_{i=1}^{N}\alpha_i^{(l)} \cdot \boldsymbol{h}_i^{(l)}, \quad \boldsymbol{m} = \boldsymbol{q}^{(l)}.
}

First, we compute the attention $\alpha_i^{(l)}$ for each node $v_i$ based on its cosine similarity with the memory vector. Then we update the global context $\boldsymbol{q}^{(l)}$ by weighted aggregating different nodes and update the memory vector $\boldsymbol{m}$. 
Such a memory-based graph update strategy can get rid of the negative influence of the abnormal nodes on the ``positive" global context as much as possible.

%===============================================================
	\begin{algorithm}[t]
		\label{al: algo}
		{\small \caption{\smodel or \pmodel\label{algo:training}}
			\KwIn{A set of labeled/pseudo-labeled graphs $\{G_i\}_{i=1}^M$. Each graph $G_i$ is consist of the adjacency matrix $A \in \mathbb{R}^{N \times N}$, and feature matrix ${X} \in \mathbb{R}^{N \times d}$. Also the learning rate $\eta$, an $L$-layers GNNs encoder: $f_{\text{GNN}}({X}, {A}, {W})$.} 
			\KwOut{Learned parameter $W$ of $f_{\text{GNN}}$, where $W = \{W_{\text{edge}}, W_{\text{node}}, W_{\text{global}}\}$}
			Initialize the global context $\boldsymbol{q}^{(0)} = \frac{1}{N}(\sum_{j=1}^{N}\boldsymbol{x}_j)$; \\
			Initialize the trainable parameter $W$ of $f_{\text{GNN}}$; \\			
			\Repeat{Converges}{
				\ForEach{ $G_i$}{
					\For{ $l$ from 1 to $L$}{
						Edge update to get the adjacency matrix: 
						${A}^{(l)}=f_{\text{edge}}\left({H}^{(l-1)}, {A}^{(l-1)}, \boldsymbol{q}^{(l-1)}, W_{\text{edge}}\right)$.\\
						Node update to get the node embeddings:
						$H^{(l)} = f_{\text{node}}\left(H^{(l-1)}, A^{(l)}, W_{\text{node}}\right)$.\\
						Graph update to get the global context:
						$\boldsymbol{q}^{(l)} = f_{\text{graph}}({H}^{(l)}, \boldsymbol{q}^{(l-1)}, W_{\text{global}})$\\
					}
					Compute the loss function in Eq.~\eqref{eq:final}.\\
					Update $W$ by $W = W - \eta \nabla_{W} \mathcal{L}$;\\
				}
			}
		}
	\end{algorithm}
%===============================================================

\subsection{Training and Inference}
% Instead of directly predicting a binary label for each node, for node $v_i \in G$ to be predicted, we get its embedding $\boldsymbol{h}_i^{(L)}$ and the graph embedding $\boldsymbol{q}^{(L)}$ of the final layer, and compute the cosine similarity between them as the abnormal extent of $v_i$ (the lower, the more anomalous).

In each epoch, we optimize both the context-aware contrastive loss in Eq.~\eqref{eq:loss} and the link predictor constraint loss in Eq.~\eqref{eq:constraint} as follows:

\beq{
\label{eq:final}
\mathcal{L} = \mathcal{L}_{\text{con}} + \lambda \mathcal{L}_{\text{link}},
}

\noindent where $\lambda$ is a balanced hyper-parameter. We empirically set $\lambda$ as 0.2 in our model. Algorithm~\ref{al: algo} outlines the training process. 
During inference, for node $v_i$ to be predicted, instead of directly predicting a binary label, we estimate its abnormality score by computing the cosine similarity between its embedding $\boldsymbol{h}_i^{(L)}$ and the graph embedding $\boldsymbol{q}^{(L)}$ of the final layer (the lower, the more anomalous).

\subsection{Complexity}
The time complexity of \smodel is the same order of magnitude as the vanilla GNN.
Because on top of the node update step of GNN, \smodel adds two additional efficient steps, edge update, and graph update. The time complexity of edge update in layer $k$ is $\mathcal{O}\left(D_kE\right)$, where $E$ is the number of edges and $D_k$ is the embedding size in layer $k$. 
The time complexity of graph update in layer $k$ is $\mathcal{O}\left(D_kN\right)$, where $N$ is the number of nodes. Since the embedding size is far smaller than the number of nodes or edges, the time complexity of \smodel grows linearly with the graph scale, which is the same as the vanilla GNN.

%% file: exp.tex
\section{Experiment}
\label{sec:exp}
In this section, we perform two major experiments to verify the performance of the supervised \smodel and the unsupervised \spmodel respectively on two genres of anomaly detection applications. All the codes and the datasets are online now\footnote{https://github.com/THUDM/GraphCAD}.

\begin{table}
	\newcolumntype{?}{!{\vrule width 1pt}}
	\newcolumntype{C}{>{\centering\arraybackslash}p{4em}}
	\caption{
		\label{tb:dataset} Data statistics. We present the total and the average number of the nodes, and edges in all the graphs. The number of relations contained in the links. Concentration of a graph is calculated as the average cosine similarity of all pairs of nodes, wherein a node is instantiated by its eigenvector. Concentration of a dataset is the average concentration of all the graphs.
	}
	\scriptsize
	%\footnotesize
	%\small
	\centering 
	\renewcommand\arraystretch{1.0}
	\begin{tabular}{@{~}l@{~}?*{1}{CC?CC}@{~}}
		\toprule
		\multirow{2}{*}{\textbf{Datasets}}
		&\multicolumn{2}{c?}{\textbf{Multi-graph}}
		&\multicolumn{2}{c}{\textbf{Single-graph}} 
		\\
		\cmidrule{2-3} \cmidrule{4-5}
		
		& \textbf{AMiner} & \textbf{MAS} & \textbf{Alpha} & \textbf{Yelp} 
		\\
		\midrule
		\tabincell{l}{\#Nodes \\ (Normal, Abnormal\%)	}       
		&  \tabincell{c}{192,352\\ (90.5, 9.5)}   
		& \tabincell{c}{117,974\\(84.4, 15.6)}  &  \tabincell{c}{3,783\\(3.6, 2.7)}  
		&\tabincell{c}{45954\\(85.4, 14.5)}        \\
		\#Total Links 	& 12,669,793 &  2,543,833  & 24,186 & 3,846,979\\
		%\#Multiple Links 	& \tabincell{c}{P-A-P\\(9,418,884)\\P-O-P\\(2,510,200)} &  2,543,833  & 24,186 & 3,846,979\\
		\#Graphs   &  1,104   & 2,098   & 1  &   1 \\
		Average \#nodes & 174   &   56    &   -  &  -\\
		Average \#links    & 11,476   &  1,212     &  -   & -\\
		\#Relation    & 3   &  2     &  2   & 3\\
		Concentration    & 0.0063 & 0.0075 & 0.0088 & 0.0003 \\
		Normal links \% &   97.23           &   97.00        & -       & 77.30 \\
		\bottomrule
	\end{tabular}
	
\end{table}

\subsection{Experimental Setup}
\vpara{Datasets.}
We evaluate the proposed model for detecting the wrongly assigned papers in researchers' profiles on two academic datasets and detecting users who give fraudulent ratings on two business websites. The academic datasets are both multi-graph datasets and the business datasets are both single-graph datasets. Table~\ref{tb:dataset} shows the statistics.
%We conduct experiments on four datasets derived from both academic and rating graphs to verify the effectiveness of \model. For academic graphs, we employ the authors' profiles, i.e., papers in their homepage, from AMiner~\cite{tang2008arnetminer} and Microsoft Academic Service (MAS)~\cite{sinha2015overview, roy2013microsoft} respectively. For rating graphs, we utilize two benchmark datasets, the YelpChi spam review dataset~\cite{rayana2015collective} and Bitcoin Alpha~\cite{kumar2016edge}. Table.\ref{tb:dataset} shows their properties. All datasets are publicly available.

\begin{itemize}[leftmargin=1em]
	\item \textbf{AMiner:} is a free online academic system\footnote{www.aminer.org} collecting over 100 million researchers and 260 million publications~\cite{tang2008arnetminer, zhang2021name}. 
	We extract 1,104 online researchers' profiles, then for each profile, we build a graph by adding papers as nodes and creating a link between two papers if they share the same coauthors, venues  or organizations
% 	\footnote{The profiles of the system are created automatically and cannot avoid the mistakes.}. 
	The wrongly assigned papers are provided by human annotators.
	
	\item \textbf{MAS:} is the Microsoft academic search system\footnote{cn.bing.com/academic} containing over 19 million researchers and 50 million publications~\cite{sinha2015overview}. 2,098 graphs are extracted and constructed similarly as AMiner, and the ground-truth labels are provided by KDD cup 2013~\cite{roy2013microsoft}. 
	
	\item \textbf{Alpha~\cite{kumar2016edge}:} is a Bitcoin trading platform. A single graph is created by adding users as nodes and the rating relationships between users as links. Benign users are the platform's founders or users who are rated positively. Fraudsters are the users who are rated negatively by the benign users.
	
	\item \textbf{Yelp~\cite{rayana2015collective}:} is a platform for users to rate hotels and restaurants. A single graph is created by adding spam~(abnormal) and legitimate~(normal) reviews filtered by Yelp as nodes. The links are created between two reviews if they are posted by the same user, on the same product with the same rating, or in the same month. 
	
	%spam review dataset includes hotel and restaurant reviews filtered (spam) and recommended (legitimate) by Yelp. A single graph is created by adding the users and products as nodes and the rating relationships between users as links. Benign users are those with more than 80\% helpful votes and fraudsters are those with less than 20\% helpful votes.
\end{itemize}

For the two multi-graph academic datasets, we use the paper title embedded by BERT as the initial feature for each node. For Yelp, we leverage a sparse matrix of 100-dimension Bag-of-words initial features for each node~\cite{dou2020enhancing}. For Alpha, since we do not have the side information of nodes, we conduct eigen-decomposition on the graph Laplacian s.t. $I-D^{-1/2}AD^{-1/2} = U\Lambda U^\top$ with $I$ as the identity matrix, $D$ as the degree matrix, and use the top eigenvectors in U~\cite{von2007tutorial} as the initial 256-dimensional features for each node.

\begin{table}
	\newcolumntype{?}{!{\vrule width 1pt}}
	\newcolumntype{C}{>{\centering\arraybackslash}p{2.5em}}
	\renewcommand\tabcolsep{3.5pt} 
	\caption{
		\label{tb:overall_res} Performance of \smodel compared with the baseline models  (\%) .
	}
	%\footnotesize
	\scriptsize
	\centering 
	\renewcommand\arraystretch{1.0}
	\begin{tabular}{@{~}l?@{~}*{1}{CC?}*{1}{CC?}*{1}{CC?}*{1}{CC}@{~}}
		\toprule
		
		\multirow{3}{*}{\vspace{-0.3cm} Model}
		&\multicolumn{4}{c?}{\textbf{Multi-graph}}
		&\multicolumn{4}{c}{\textbf{Single-graph}} 
		
		\\
		\cmidrule{2-5} \cmidrule{6-9}
		
		&\multicolumn{2}{c?}{\textbf{AMiner}}
		&\multicolumn{2}{c?}{\textbf{MAS}}  
		&\multicolumn{2}{c?}{\textbf{Alpha}} 
		&\multicolumn{2}{c}{\textbf{Yelp}} 
		
		\\
		\cmidrule{2-3} \cmidrule{4-5} \cmidrule{6-7} \cmidrule{8-9}
		& {AUC} & {MAP} & {AUC} & {MAP} & {AUC} & {MAP} & {AUC} & {MAP}   \\
		\midrule
		\multicolumn{9}{c}{\textbf{Graph Neural Networks} }  \\
		\midrule
		GCN
		&  75.68 & 59.28  & 78.44  & 63.89 & 82.12 & 78.09 & 70.85 & 29.18  \\
		GAT 
		&  74.97 & 59.87 &  78.25 & 62.75 & 78.45 & 76.32 & 75.28 & 32.11 \\
		GSAGE 
		&  73.01 & 60.51 & 76.55 & 62.12 & 79.23 & 75.14 & 76.34 & 34.12 \\
		GIN 
		&  74.28 & 57.75 & 75.62 & 59.96 & 87.39 & 86.79 & 76.67 & 34.60 \\
		\midrule
		\multicolumn{9}{c}{\textbf{Graph-based Anomaly Detection Models}} \\
		\midrule
		LogisticReg
		& 62.49 & 33.68 & 73.67  & 56.64 & 71.88 & 72.43 & 67.74 & 26.27  \\
		GeniePath
		& 76.18 & 50.67 & 81.93  & 65.66 & 75.02 & 68.89 & 76.78 & 33.67     \\
		GraphConsis
		& 71.37 & 52.21 & 76.67  & 60.25 & 86.99 & 86.63 & 70.27 & 26.64     \\
		CARE-GNN
		& 77.74 & 55.38 & 80.01  & 66.98 & 84.26 & 85.20 & 77.14  & 36.84     \\		
		\midrule
		\textbf{\model}
		&\textbf{89.84}    & \textbf{78.73} &  \textbf{87.62}  & \textbf{75.18} & \textbf{90.53}  & \textbf{91.20}  & \textbf{79.64} &  \textbf{37.15}     \\	
		\bottomrule
	\end{tabular}
	
\end{table}

\vpara{Evaluation Metrics.} We adopt Area Under ROC Curve (AUC), broadly adopted in anomaly detection~\cite{dou2020enhancing, liu2020alleviating}, and Mean Average Precision (MAP), which pays more attention to the rankings of the anomalies, as the evaluation metrics.

\hide{
\begin{figure*}[htbp]
	\centering
	\hspace{-0.2in}
	\subfigure[AMiner]{\label{subfig:aminer}
		\includegraphics[width=0.2\textwidth]{figures/pertrain_aminer_final}
	}
	\hspace{-0.15in}
	\subfigure[MAS]{\label{subfig:mas}
		\includegraphics[width=0.2\textwidth]{figures/pertrain_mag_final}
	}		
	\hspace{-0.15in}
	\subfigure[Alpha]{\label{subfig:alpha}
		\includegraphics[width=0.2\textwidth]{figures/pertrain_alpha_final}
	}
	\hspace{-0.15in}
	\subfigure[Yelp]{\label{subfig:yelp}
		\includegraphics[width=0.2\textwidth]{figures/pertrain_yelp_final}
	}
	\hspace{-0.15in}
	\subfigure[Performace on hard cases.]{\label{subfig:hardcases}
		\includegraphics[width=0.2\textwidth]{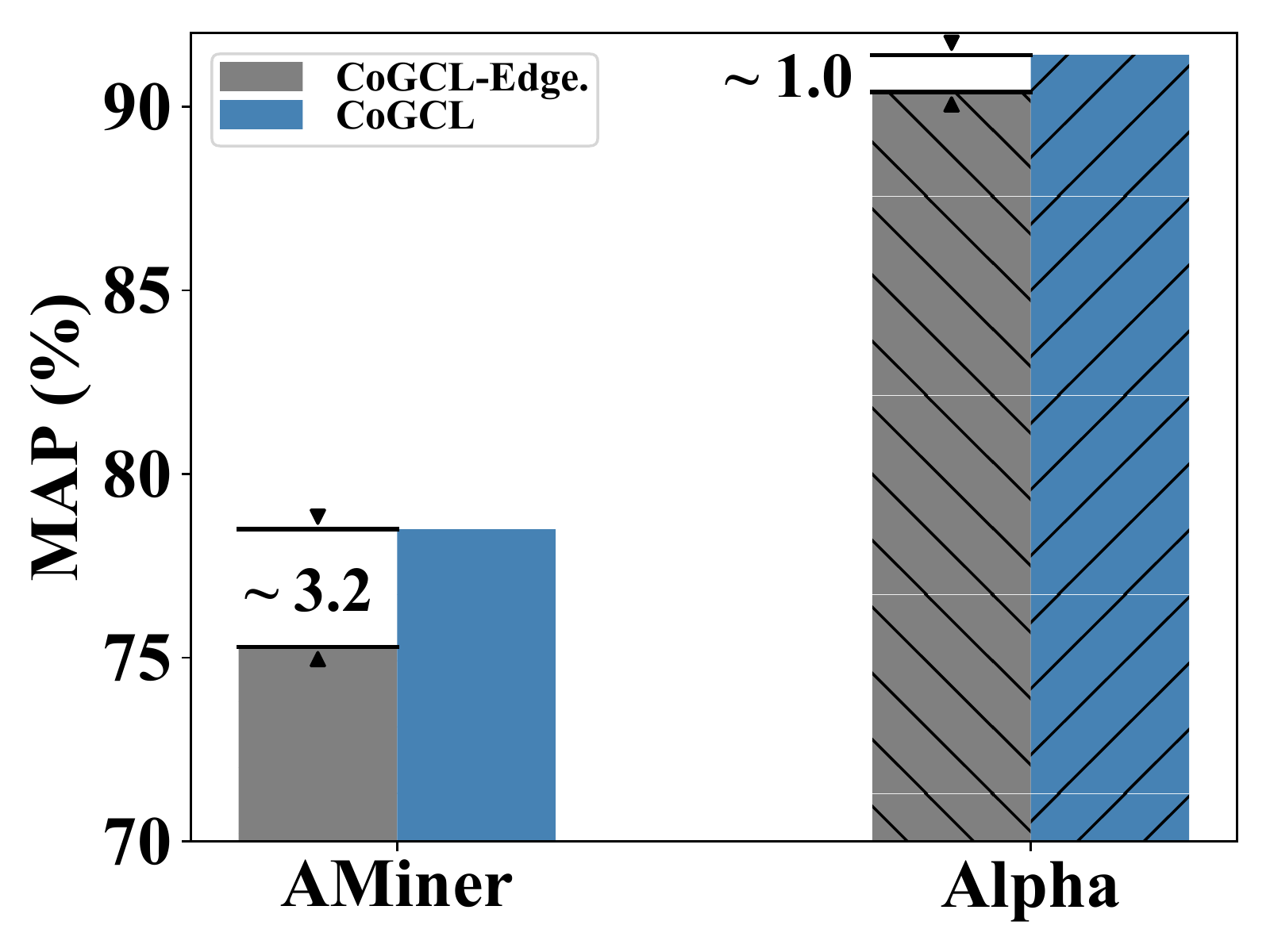}
	}
	\caption{\label{fig:para} The fine-tuning performances under different proportion of the labeled data in four datasets.}
\end{figure*}
}

\subsection{Evaluation of \smodel}

\vpara{Baselines.} We compare \smodel with four general GNNs models, including GCN~\cite{kipf2016semi}, GAT~\cite{velivckovic2017graph}, GraphSAGE~\cite{Graphsage}, and GIN~\cite{xu2018powerful}, and three specific GNN models for anomaly detection, i.e., GeniePath~\cite{liu2019geniepath}, GraphConsis~\cite{liu2020alleviating}, and CARE-GNN~\cite{dou2020enhancing}, which are further introduced in Section~\ref{sec:related}.
%To reduce the negative influence of the suspicious links, GeniePath adopts the LSTM-based attention mechanism, GraphConsis designs neighbor sampling and relation-aware attention strategies, and CARE-GNN applies reinforcement learning to filter noisy neighbors. 
Other GNN-based anomaly detection models such as GAS~\cite{li2019spam}, FdGars~\cite{FdGars}, Semi-GNN~\cite{SemiGNN}, and Player2Vec~\cite{zhang2019key}, are empirically proven to be less powerful than the adopted baselines, thus are ignored in the experiments.
We additionally compare with logistic regression by injecting top eigenvectors as features to perform node classification.  

\hide{
	\begin{itemize}[leftmargin=1em]
	\item \textit{Logistic Regression} Is a non-GCN classifier. Specifically, we feed the eigenvector of a node into logistic regression to predict whether it is an anomaly or not.  	
	\item \textit{GeniePath}~\cite{liu2019geniepath} Is GNN-based method which learns the neighborhood aggregation functions using LSTM and attention mechanism. It assigns low weights to the suspicious links for mitigating their negative influence.
	\item \textit{GraphConsis}~\cite{liu2020alleviating} Proposes three mechanisms of context embedding, neighbor sampling, and relation attention to address the inconsistency problems cased by anomalies.
	\item \textit{CARE-GNN}~\cite{dou2020enhancing} Modifies neighborhood aggregation process of GNNs via deep reinforcement learning. It gets rid of pre-defined thresholds and can be easily incorporated into other GNN-based frameworks. 
\end{itemize}
}

\begin{table}
	\newcolumntype{?}{!{\vrule width 1pt}}
	\newcolumntype{C}{>{\centering\arraybackslash}p{2.5em}}
	\renewcommand\tabcolsep{3.5pt} 
	\caption{
		\label{tb:ablation_study} Ablation Studies of \model (\%). We adopt the best node update strategy, i.e., GCN for Multi-graph datasets and GIN for Single-graph datasets.
	}
	\footnotesize
	\scriptsize
	%\centering 
	\renewcommand\arraystretch{1.0}
	\begin{tabular}{@{~}l?@{~}*{1}{CC?}*{1}{CC?}*{1}{CC?}*{1}{CC}@{~}}
		\toprule
		\multirow{3}{*}{\vspace{-0.3cm} Model}
		&\multicolumn{4}{c?}{\textbf{Multi-graph (GCN)}}
		&\multicolumn{4}{c}{\textbf{Single-graph (GIN)}} 
		
		\\
		\cmidrule{2-5} \cmidrule{6-9}
		&\multicolumn{2}{c?}{\textbf{AMiner}}
		&\multicolumn{2}{c?}{\textbf{MAS}} 
		&\multicolumn{2}{c?}{\textbf{Alpha}} 
		&\multicolumn{2}{c}{\textbf{Yelp}} 
		
		\\
		\cmidrule{2-3} \cmidrule{4-5} \cmidrule{6-7} \cmidrule{8-9}
		& {AUC} & {MAP} & {AUC} & {MAP} & {AUC} & {MAP} & {AUC} & {MAP}   \\
		\midrule
		\textbf{\model}
		&\textbf{89.84}    & \textbf{78.73} &  \textbf{87.62}  & \textbf{75.18} & \textbf{90.53}  & \textbf{91.20}  & \textbf{79.64} &  \textbf{37.15}     \\	
		\midrule
		\multicolumn{9}{c}{\textbf{Variant Loss Functions}} \\
		\midrule
		+ CE\_loss
		&80.30   & 61.02 & 81.53  & 67.42 & 88.64  & 89.10  & 79.18 & 37.05   \\	
		\midrule
		\multicolumn{9}{c}{\textbf{Variant Edge Update Strategies}} \\
		\midrule
		- LP\_constraint
		& 89.36  &78.57 & 87.25  & 74.56  & 89.70  & 90.55  & 78.63 & 35.62   \\
		- Global Info.
		& 88.67 & 77.13 & 86.78  & 73.48  & 89.93  & 90.63  & 78.61 & 35.20   \\	
		- Edge Update
		& 88.21 & 76.97 & 86.34  & 71.54 & 89.69  & 89.86  & 78.51 & 35.19   \\	
		\midrule
		\multicolumn{9}{c}{\textbf{Variant Node Update Strategies}} \\
		\midrule
		$\modelgcn$
		&\textbf{89.84}    & \textbf{78.73} & \textbf{87.62}  & \textbf{75.18} & 85.07  & 87.09  & 73.33 & 30.64   \\	
		$\modelgat$
		& 89.00 & 77.98 & 86.30  & 72.75 & 83.26  & 86.37  & 72.87 & 29.91   \\	
		$\modelsage$
		& 87.84  & 75.26 & 82.86  & 68.52  & 86.54  & 87.07  & 76.52  & 33.28   \\	
		$\modelgin$
		& 89.04   & 77.02 & 83.39  & 67.46 & \textbf{90.53}  & \textbf{91.20}  & \textbf{79.64} &  \textbf{37.15}  \\	
		\midrule
		\multicolumn{9}{c}{\textbf{Variant Graph Update Strategies}} \\
		\midrule	
		+ Avg Pooling
		& 89.01 & 77.66 & 86.39  & 72.73 & 89.02  & 89.57  & 78.70 & 34.55   \\
		+ Sum Pooling
		& 86.95 & 72.09 & 86.45  & 73.98 & 88.01  & 86.69   & 76.79 & 34.11   \\
		+ Max Pooling
		& 80.91 & 61.97 & 84.06  & 70.52 & 89.25  & 88.56  & 78.54 & 33.96   \\
		\bottomrule
	\end{tabular}
	
\end{table}

\vpara{Setup.}
%For AMiner and MAS, we select 70\% researchers' graphs as the training data and use the remaining ones as the test set. During training, we perform the graph convolution on each graph and use all the nodes in it to compute the loss function in Eq.\eqref{eq:loss}. For testing, we first rank nodes based on their cosine similarities calculated between node embeddings with their corresponding graph context embeddings in ascending order,  then we evaluate AUC and MAP for the ranked nodes in each test graph and average the metrics of all the test graphs as the final metrics.For Alpha and Yelp, we select 75\% labeled nodes from the single graph as the training data and use the remaining ones as the test set. We perform the graph convolution on the single graph, use all the nodes in the training set to compute the loss function, and evaluate AUC and MAP for the ranked nodes in the test data.
For the multi-graph datasets, i.e., AMiner and MAS, we extract 70\% graphs as the training set, 10\% as the validation set, and 20\% as the test set. During training, we perform \smodel on each graph following Algorithm~\ref{algo:training}. For testing, we rank nodes based on the cosine similarities between their node embeddings and the corresponding graph context in an ascending order, then we evaluate AUC and MAP of each graph and average the metrics of all test graphs as the final metrics. 
For the single-graph datasets, i.e. Alpha and Yelp, we also extract 70\% of labeled nodes from the single graph as the training data, 10\% as the validation set, and 20\% as the test set. We perform \smodel on the single graph, and evaluate AUC and MAP for the ranked nodes in the test data. Notably, for GraphConsis and CARE-GNN whose inputs are the graph with multi-relation links, we use its multi-relation information, while \smodel merges multi-relation links between two nodes into single-relation links.
% For all experiments, we run 5 trials and report the mean results.
We run 5 trials and report the mean results.

\vpara{Implementation.} 
For \smodel, the number of layers $L$ is empirically set as 2, the temperature hyperparameter $\tau$ is set as 0.1 in Eq.~\eqref{eq:loss}, and $\lambda$ is set as 0.6 in Eq.~\eqref{eq:GUMBEL}. We use Adam with learning rate 1$\times 10^{-3}$ for training. The learning rate is set as the initial value over the first 10\% steps, and then linearly decay to 1/10 of the initial value. 

For the GCN, GAT, GraphSAGE, and GIN, We leverage the implementations provided by Pytorch Geometric\footnote{https://github.com/rusty1s/pytorch\_geometric}, and set the number of convolutional layers as 2 for all general GNNs. For GeniePath, GraphConsis, and CARE-GNN, we use the authors' official codes with the same training settings. The data format is transformed appropriately to fit their settings. All models are running on Python 3.7.3, 1 NVIDIA Tesla V100 GPU, 512GB RAM, Intel(R) Xeon(R) Gold 5120 CPU @ 2.20GHz.

\vpara{Overall Results.} 
Table \ref{tb:overall_res} presents the performance of \smodel and all the comparison methods on four datasets. We can see that,  \smodel substantially improves over all other baselines, +2.5-27.35\% in AUC and +0.31-28.06\% in MAP, on all the datasets. Among all the baselines, logistic regression performs the worst as it only leverages the structural information and does not apply graph convolution to integrate neighbor messages. None of the specific graph-based anomaly detection methods can keep the advantage over the general GNNs on all the datasets. For example, GraphConsis~\cite{liu2020alleviating} performs worse than GIN~\cite{xu2018powerful} on AMiner. 
%which reveals the dedicated structure for anomaly detection can not maintain its strength in different scenarios. 
The highlighted results in the table are from \model, which adopts the context-aware contrastive objective and the  three-stage GNN encoder, performing stably the best over all the datasets.

\begin{figure}[t]
	\centering
	\subfigure[\smodel vs. w/o EdgeU]{\label{subfig:hard_case}
		\includegraphics[width=0.23\textwidth]{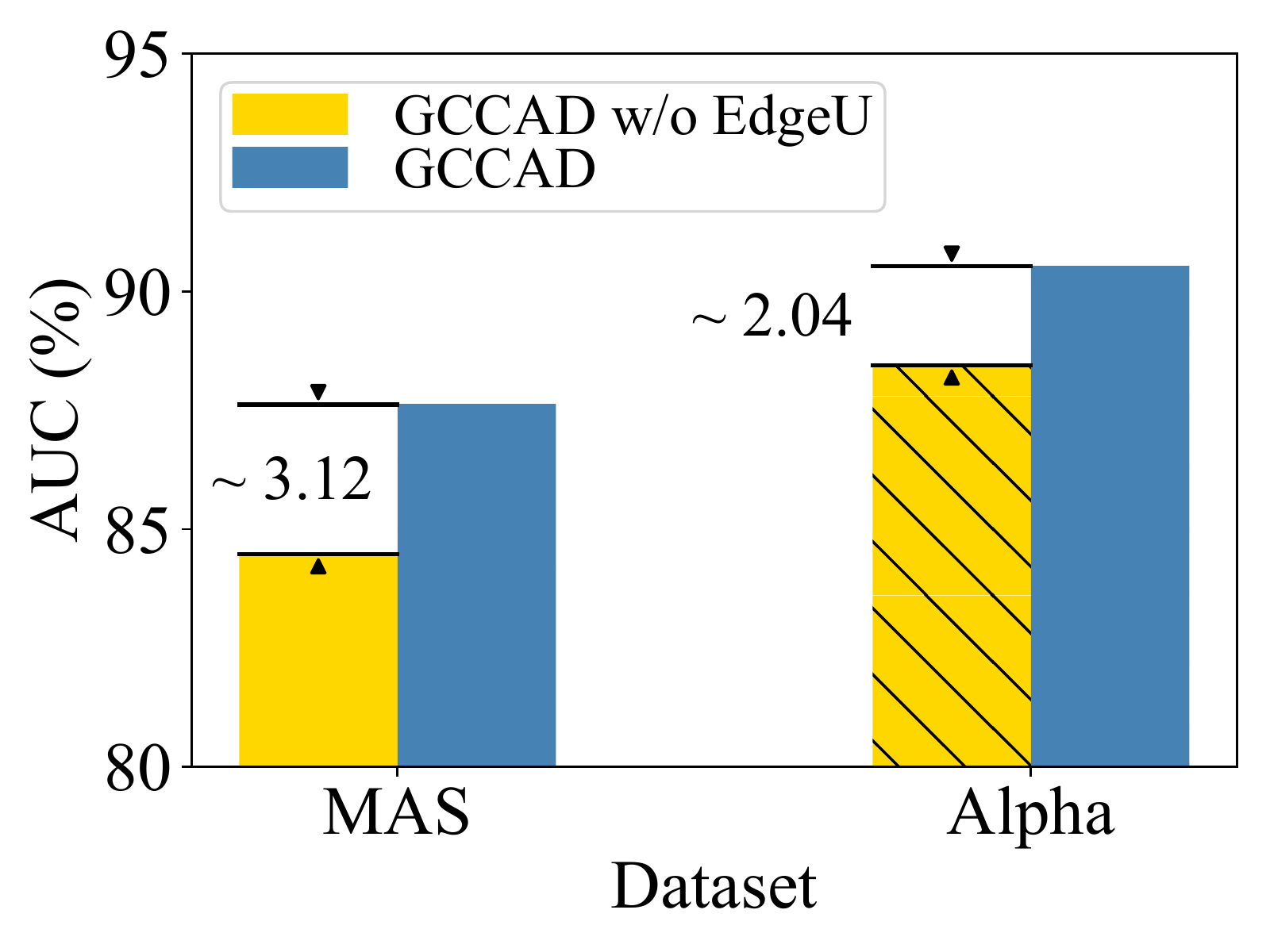}
	}
	\hspace{-0.1in}
	\subfigure[MAP vs. Corrupt Ratio]{\label{subfig:corrupt_ratio}
		\includegraphics[width=0.23\textwidth]{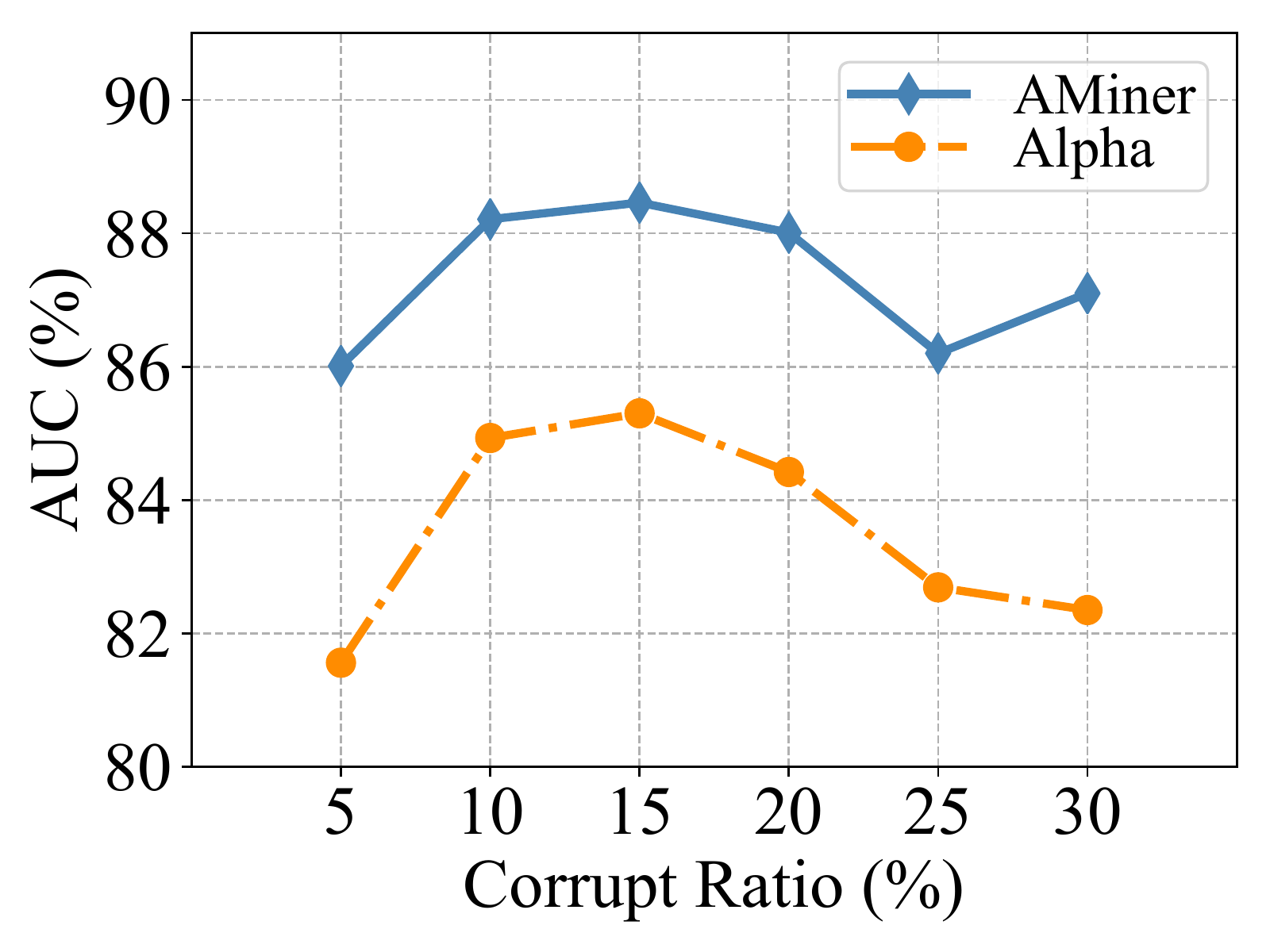}
	}
	\caption{\label{fig:hyperparameters} We study (a) the performance gain of the supervised \smodel obtained by edge update step on the filtered Alpha and MAS datasets;  (b) the best AUC of \spmodel without fine-tuning when varying the corrupt ratio, i.e., the ratio of the injected abnormal nodes, on AMiner and Alpha.}
\end{figure}

\vpara{Ablation Studies.} To this end, there have two major differences: the context-aware contrastive objective and the three-stage GNN encoder between \smodel and other baselines. Thus we conduct the ablation studies to verify the efficacy of different components in \model.  The four main model variants are presented as follows, and the results are illustrated in Table~\ref{tb:ablation_study}.

\begin{itemize}[leftmargin=1em]
	\item \textbf{w/ CE\_loss:} Change the contrastive loss in Eq.~\eqref{eq:loss} with cross-entropy objective.
    \item \textbf{Edge Update Strategies:} Change the edge update component in \secref{subsec:edge} to \textit{w/o LP\_constraint} that removes the constraint loss $\mathcal{L}_{\text{link}}$ in Eq~\eqref{eq:constraint},  
	\textit{w/o Global Info.} that removes $(\boldsymbol{h}^{(l-1)}_{(i/j)} - \boldsymbol{q}^{(l-1)})$ in Eq.\eqref{eq:linkage_likelihood}, or \textit{w/o Edge Update} that removes the entire edge update step.
	\item \textbf{Node Update Strategies:} Change the node update strategies in Eq.~(\ref{eq:aggregation}) and Eq.~(\ref{eq:combination}) to the node update strategies in GCN, GAT, GraphSAGE, or GIN. 
	\item \textbf{Graph Update Strategies:} Change the memory-based readout function in Eq.\eqref{eq:readout} to average pooling (\textit{Ave Pooling}), sum pooling (\textit{Sum Pooling}), and maximal pooling (\textit{Max Pooling}) of all the nodes.
\end{itemize}

%\ipara{Results. }The results are shown in Table~\ref{tb:ablation_study}. 

\ipara{Node Update Strategies.}  From Table~\ref{tb:ablation_study}, we can see that 
\smodel with various node update strategies outperforms other baselines in most cases, which suggests the superiority of context-aware contrastive objective and the three-stage GNN encoder.

We also observe that different strategies have the performance gap, with less than 6\% in AUC and 7\% in MAP.  On the multi-graph datasets, \smodel performs the best with the mean AGGREGATION and the concatenation COMBINE strategies in GCN~\cite{kipf2016semi}, while on the single-graph datasets, it performs the best with the sum AGGREGATION and the concatenation COMBINE strategies in GIN~\cite{xu2018powerful}. Obviously, the trend of performance changes over different node update strategies is akin to that of general GNNs, as shown at the top of Table~\ref{tb:overall_res}. Thus we conjecture the instability is mainly due to the intrinsic traits of the node update strategy in GNNs. 

Thus, unless stated otherwise, we only conduct experiments of \smodel with the best node update strategy, that is, GCN on the multi-graph datasets and GIN on the single-graph datasets, in the remaining parts.

\ipara{The Contrastive Loss Function.}
From Table~\ref{tb:ablation_study}, we see that the performance gain of \smodel over +CE\_loss on the multi-graph datasets (+6.09-9.54\% in AUC and +7.76-17.71\% in MAP) is much higher than that on the single-graph datasets (+0.46-1.89\% in AUC and +0.60-2.10\% in MAP). 
% Since the multi-graph datasets correspond to the scenario that the data distributions is dynamically changed in different graphs, while the single-graph datasets correspond to the scenario that the data distributions are relatively static,  
Such results empirically validate Theorem 1 that the context-aware contrastive objective is more resilient than cross-entropy in terms of the generation capacity. 
Since the abnormal node distributions in different graphs are usually quite different, the multi-graph setting presents much more diverse abnormal distributions than the single-graph setting. Thus optimizing the entropy of abnormal nodes take a critical role to learn a more generalized model across various graphs. 
What's more, our proposed context-aware contrastive loss performs slightly better than CE\_loss in the single-graph datasets, which shows that our method can tackle the class imbalance issue to some degree.

%, thus the performance gap between \smodel and other baselines on the multi-graph datasets is larger than that on the single-graph datasets. 

%To this end, there have two major differences: the context-aware contrastive objective and the three-stage GNN encoder between \smodel and other baselines. Thus to alleviate the effects caused by GNN encoder of \model, we keep the GNN encoder the same as the proposed three-stage encoder, but change the loss function to the cross-entropy loss, and show the results of +CE\_loss on Table~\ref{tb:ablation_study}. Compared with +CE\_loss, \smodel still achieves 6.09-9.54\% AUC improvements on the multi-graph datasets, while 0.46-1.89\% AUC improvements on the single-graph datasets,  which indicates that the proposed context-aware contrastive loss function makes an significant contribution. 

% when comparing \smodel with baselines, the ablation studies (\smodel vs. \smodel w/ CE\_loss) have already confirmed the contrastive objective mainly contributes to the improvements. 

% \ipara{Objective Functions.} Overall, \smodel outperforms \smodel w/ CE\_loss by up to 9.54\% in AUC and 17.71\% in MAP. However, the performance gain on the multi-graph datasets (6.09-9.54\% in AUC) are larger than that on the single-graph datasets (0.46-0.89\% in AUC), which agrees with the statements from Theorem 1 that the context-aware contrastive loss function is more robust than the cross-entropy loss function to address various scenarios. 

\ipara{Context-aware GNN Encoder.} 
\smodel w/ CE\_loss outperforms other state-of-the-art baselines by up to 17.81\% AUC, which elucidates the powerful representation capability of the proposed context-aware GNN encoder. The improvements can give credit to two perspectives: the context-aware edge update and the memory-based global update. Thus, We further verify the effectiveness of each component.   

\vpara{\ipara{Edge Update Strategies.}} \smodel w/o LP\_constraint performs slightly worse than \model, denoting link prediction loss can not only accelerate the optimization of link predictor but also contribute to the overall improvement. 

\smodel w/o Global Info. underperforms \model, revealing the context-aware link predictor that incorporates the distance to the global context as the implicit supervision, can benefit the suspicious link prediction.

Compared with above two variants, \smodel w/o Edge Update performs the worst (-0.84-1.63\% in AUC), verifying the efficacy of the proposed edge update mechanism. However, the performance enhancement brought by the edge update is not significant. We speculate the norm links which connect the nodes of the same labels occupy the majority of all the links (e.g., 90.5\% on AMiner and 84.4\% on MAS in Table.~\ref{tb:dataset}). Thus, removing the small amount of suspicious links by the link predictor can only exert limited effect. To verify this, we select the hard instances, i.e., the abnormal nodes that connect at least $k$\footnote{Empirically, $k$ is set as 1 in MAS and 3 in Alpha respectively.} normal nodes, from both MAS and Alpha, and re-evaluate the variants on filtered datasets. From Fig.~\ref{subfig:hard_case}, we can see that \smodel significantly outperforms w/o Edge Update by 3.12\% and 2.04\% AUC on the two datasets, respectively. The results indicate that the negative influence of the suspicious links can be effectively reduced by the edge update step.   

\vpara{\ipara{Global Update Strategies.}} The proposed memory-based global context update mechanism performs the best, which improves 0.83-8.93\% AUC and 1.07-16.76\% MAP over other variant mechanisms. Because the memory that records the global context of the last layer can help estimate the contribution of each node in the current layer, thus it will benefit the generation of precise global context in the current layer.

\begin{table}
	\newcolumntype{?}{!{\vrule width 1pt}}
	\newcolumntype{C}{>{\centering\arraybackslash}p{6em}}
	\renewcommand\tabcolsep{2.5pt} 
	\caption{
		\label{tb:tansferLearning} Transfer Learning about Cross-Entropy (CE) and Context Contrastive Loss (CL) on AMiner and MAS (AUC \%). The model trained on one dataset (top) and evaluate on another one (left).
	}
	\footnotesize
	\scriptsize
	\centering 
	\renewcommand\arraystretch{1.0}
	\begin{tabular}{@{~}l@{~}*{1}{?C?CC?CC}@{~}}
		\toprule
		\multicolumn{2}{c?}{Model} &\textbf{AMiner} & \textbf{MAS} & \textbf{Performance Drop}\\
		\midrule
		\multirow{2}{*}{\textbf{AMiner}}& w/ CL & 89.84 & 87.70 & \textbf{2.14}\\ 
		%\cmidrule{2-5}
		&w/ CE & 80.30 &77.05  & 3.25 \\
		\midrule
		\multirow{2}{*}{\textbf{MAS}}& w/ CL & 84.41  &  87.62 & \textbf{3.21}\\ 
		%\cmidrule{2-5}
		&w/ CE & 74.15 & 81.53 & 7.38 \\
		\bottomrule
	\end{tabular}
	
\end{table}

\hide{
\begin{table}
	\newcolumntype{?}{!{\vrule width 1pt}}
	\newcolumntype{C}{>{\centering\arraybackslash}p{6em}}
	\renewcommand\tabcolsep{2.5pt} 
	\caption{
	\label{tb:tansferLearning} Transfer Learning about Cross-Entropy (CE) and Context Contrastive Loss (CL) on AMiner and MAS (AUC \%). The model trained on one dataset and evaluate on another one.
}
	\footnotesize
	\scriptsize
	\centering 
	\renewcommand\arraystretch{1.0}
	\begin{tabular}{@{~}l@{~}*{1}{?C?CC?CC}@{~}}
		\toprule
		\multicolumn{2}{c?}{Model} &\textbf{AMiner} & \textbf{MAS} & \textbf{Performance Drop}\\
		\midrule
    	\multirow{2}{*}{\textbf{AMiner}}& w/ CL &72.10 &68.16 & 3.94\\ 
    	%\cmidrule{2-5}
		&w/ CE &70.23 &57.58 & \textbf{12.65} \\
		\midrule
    	\multirow{2}{*}{\textbf{MAS}}& w/ CL & 74.63 & 72.53 & 2.10 \\ 
    	%\cmidrule{2-5}
		&w/ CE & 72.49 & 66.16 & \textbf{6.33} \\
		\bottomrule
	\end{tabular}
	
\end{table}
}

\vpara{Transfer Learning.} To further investigate the robustness of context-aware contrastive loss compared with cross-entropy objective, we train \smodel and its variant w/ CE\_loss on AMiner/MAS and evaluate them on MAS/AMiner\footnote{Since the feature initialization process and the embedding dimension of Alpha and Yelp are obviously different from each other, we only conduct the transfer learning experiments between AMiner and MAS.}. As shown in Table~\ref{tb:tansferLearning}, the performance drop of \smodel (-2.14-3.21\% in AUC) is substantially less than that of  w/ CE\_loss (3.25-7.38\% in AUC) under the setting of transfer learning, which shows the superior robustness ability of the context-aware contrastive objective compared with the cross-entropy objective.

\subsection{Evaluation of \pmodel}
\label{subsec:pretrain}

\vpara{Baselines.}
We compare \spmodel with the state-of-the-art graph pre-training models, including GAE~\cite{kipf2016variational}, GPT-GNN~\cite{hu2020gpt}, GraphCL~\cite{you2020graph}, and the graph pre-training models tailored for anomaly detection, DCI~\cite{wang2021decoupling}, ANEMONE~\cite{jin2021anemone}, CoLA~\cite{liu2021anomaly}, in the unsupervised setting. 
GAE is to preserve the structural correlations via reconstructing the vertex adjacency matrix. In addition to the structural correlations, GPT-GNN preserves the attribute interplays via predicting the masked attributes. 
GraphCL maximizes the mutual information of two instances of the same graph obtained by graph data augmentation.  
DCI~\cite{wang2021decoupling} is a cluster-based version of DGI, 
which maximizes the fine-grained mutual information between the embeddings of a cluster and the nodes within it and is proved to be effective for graph anomaly detection. 
ANEMONE and CoLA both underperform the proposed \pmodel. Although ANEMONE and CoLA propose a similar node to sub-graph contrast learning, they sample the sub-graphs by the local random walking algorithm instead of the global clustering algorithm in \pmodel. The latter can result in more diverse normal samples that are more difficult to be distinguished. On the contrary, the former ignores such difficult normal samples during training, which will reduce the model generalization.
For a fair comparison, all the pre-training models adopt the same GNN encoder as \model.

% \begin{figure}[t]
% 	\centering
% 	%\hspace{-0.2in}
% 	%\subfigure[AMiner]{\label{subfig:paper_network}
% 	%	\includegraphics[width=0.33\textwidth]{figures/motivation}
% 	%}
% 	\subfigure[Before]{\label{subfig:t-sne}
% 		\includegraphics[width=0.23\textwidth]{figures/caseBefore}
% 	}
% 	%\hspace{-0.1in}
% 	\subfigure[After]{\label{subfig:gcn_sim}
% 		\includegraphics[width=0.23\textwidth]{figures/caseAfter}
% 	}
% 	\caption{\label{fig:case} A real example of detecting the papers that don't belong to ``Jun Lu". %, a professor from University of Munich. 
% 		The similarities between right papers and the global context (blue) vs. that between wrongly-assigned papers and the global context (red) by using input features (a) or embeddings yielded by \smodel(b).} 
% \end{figure}

\vpara{Datasets.} 
In the online AMiner system, we hide the labeled 1,104 graphs in Table~\ref{tb:dataset} and extract additional 4,800 corrupt graphs from it. According to the multi-graph corrupting strategy in~\secref{subsec:self}, we first build a single large paper graph for the whole extracted AMiner dataset and then divide it into multiple graphs according to the ownership of papers to different researchers. For each graph, we corrupt it by injecting the papers from other researchers' graphs as the anomalies with a certain corrupt ratio of 15\%, i.e., the ratio between the number of injected nodes with the number of existing nodes in the graph. The links in the original graph are kept between the nodes in corrupt graphs. Then we build the MAS corrupting graph dataset in the same way.

For the single-graph datasets, Alpha and Yelp, we first cluster it into $K$ sub-graphs by classic K-means algorithm based on their input features. 
We perform the clustering algorithm based on the feature information instead of the structure information, because structural information is more preferred to be tampered than the feature information by the abnormal users. Although the abnormal users might perform both the attribute abnormal behaviors which result in the corrupted features and the abnormal structural behaviors which result in the suspicious links between users, the latter can produce a more negative impact on the whole graph. For example, in social network, a rumor monger usually disguises themselves as regular users by building connections with regular users and dispersing the disinformation via the connections. Thus, identifying suspicious linkages between normal and abnormal ones is more crucial in graph anomaly detection. In light of this, it is inappropriate to split the graphs by the structure information that is highly probably polluted by the suspicious links. We didn’t try to split the graph of Alpha by the feature information, as it is unavailable in the dataset.

\begin{figure}[t]
	\centering
	%\hspace{-0.2in}
	%\subfigure[AMiner]{\label{subfig:paper_network}
	%	\includegraphics[width=0.33\textwidth]{figures/motivation}
	%}
	\subfigure[Alpha]{\label{subfig:alphaK}
		\includegraphics[width=0.23\textwidth]{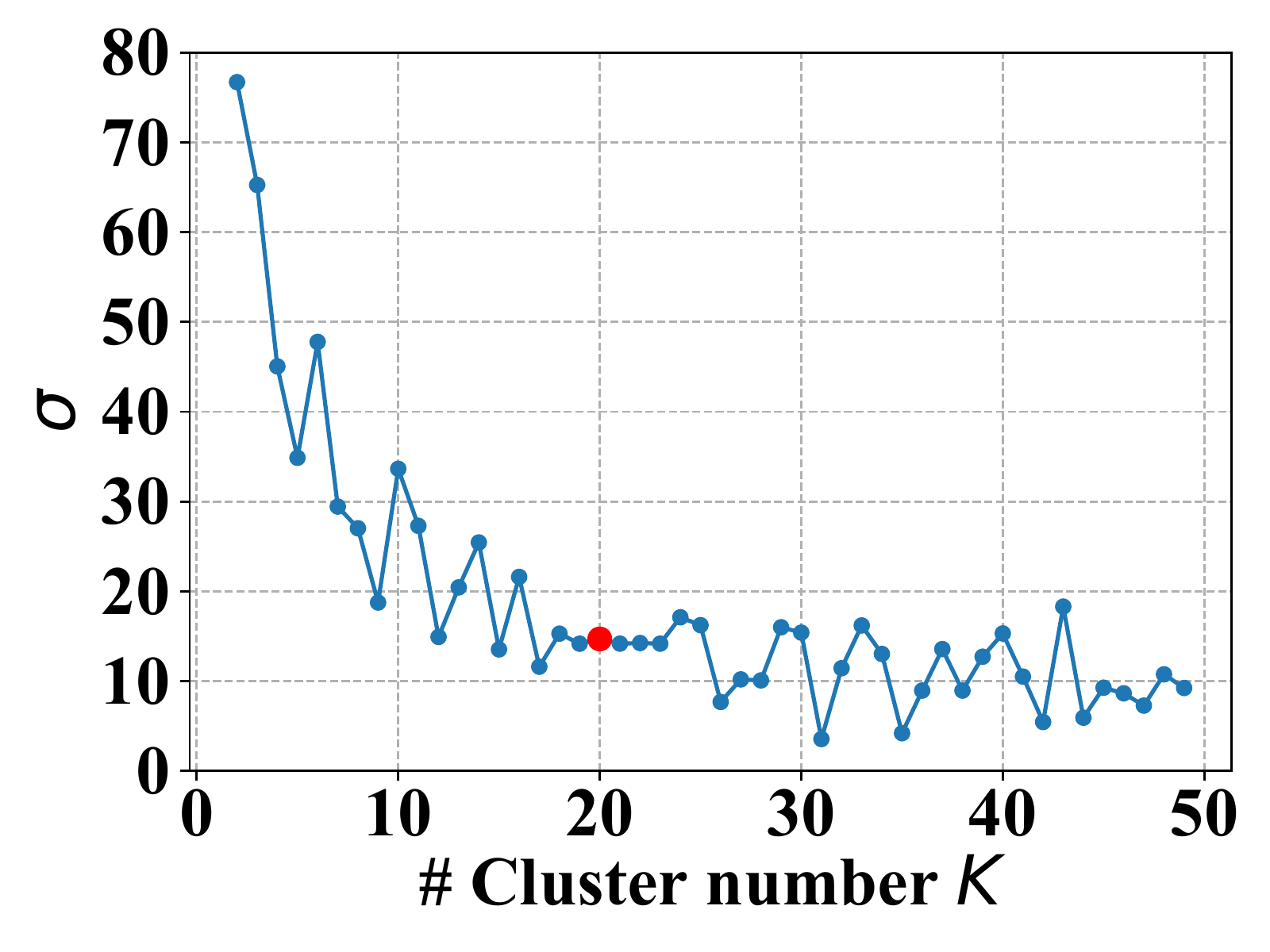}
	}
	%\hspace{-0.1in}
	\subfigure[Yelp]{\label{subfig:yelpK}
		\includegraphics[width=0.23\textwidth]{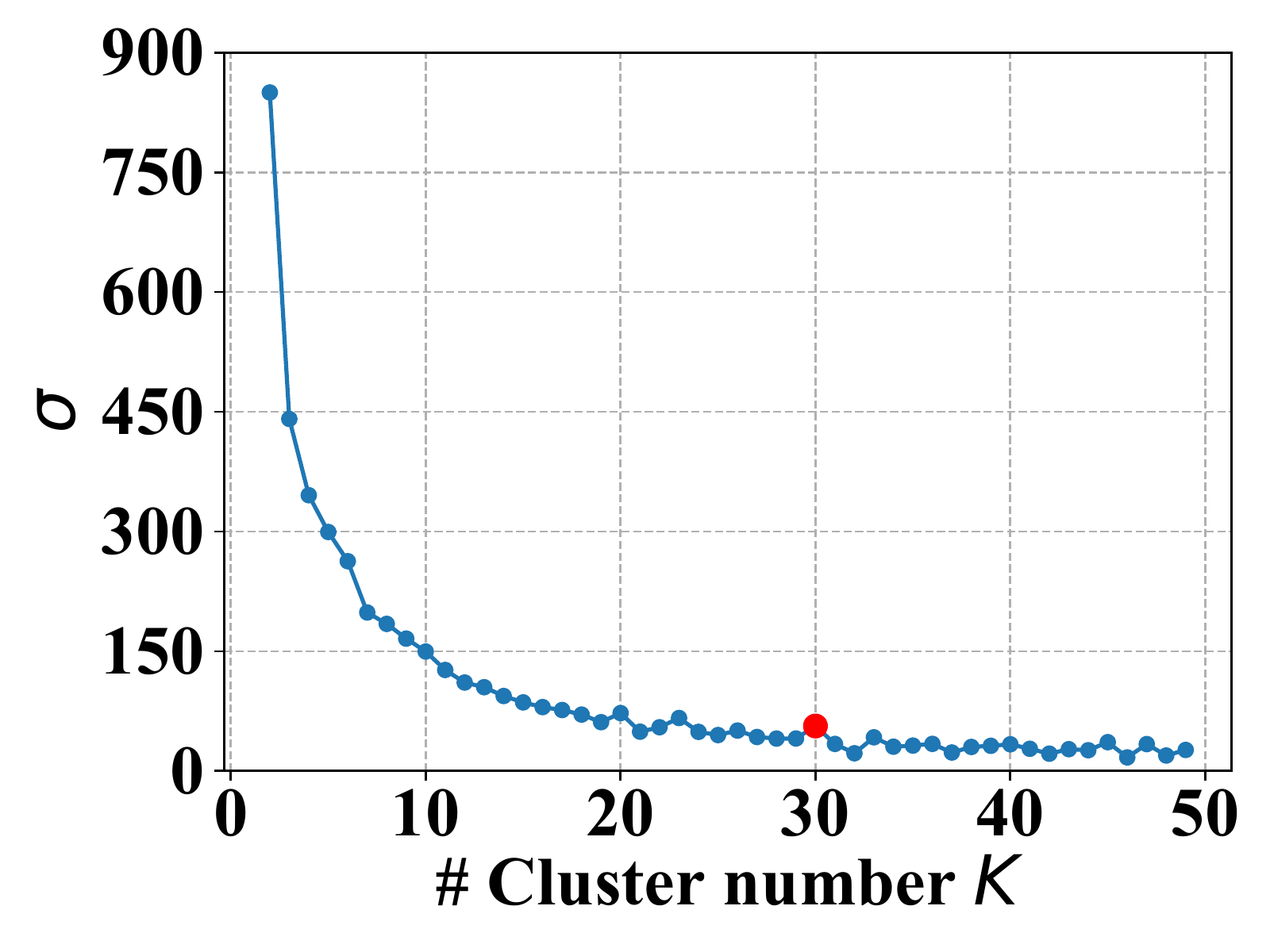}
	}
	\caption{\label{fig:cluster} The correlation between the graph (cluster) number $K$ and the inertia gap $\sigma\left(K\right)$ between $K$ and $K+1$ on Alpha (a) and Yelp (b). Red point denotes the $K^*$ that achieves the best performance of \spmodel without fine-tuning.} 
\end{figure}

The within-cluster correlation is measured by sum-of-squares criterion, namely inertia\footnote{https://scikit-learn.org/stable/modules/clustering.html\#k-means}, estimated by $\sum_{i=0}^{|V|}\min\limits_{\boldsymbol{u}_k\in C}\left(||\boldsymbol{x}_i - \boldsymbol{u}_k||^2\right)$, where $\boldsymbol{u}_k$ is the embeddings of $k$-th cluster center and $C$ is $K$ disjoint clusters. To automatically set a proper $K$, we leverage the Elbow Methods to choose the elbow point of inertia, i.e, to locate the optimal $K^*$ at which the trend of inertia changes from steep to stable (Cf. Fig.~\ref{fig:cluster}). Finally, the graph size $K$ is set as 20 on Alpha and 30 on Yelp.  

%In the online AMiner system, we hide the labeled 1,104 graphs in Table~\ref{tb:dataset} and extract additional 4,800 corrupt graphs from it.  According to Definition~\ref{eq:corrupt} in~\secref{subsec:self}, we first build a single paper graph for the whole AMiner dataset and then divide it  into multiple graphs according to the ownership of papers to different researchers. For each graph, we corrupt it by injecting the papers from other researchers' graphs as the anomalies with a certain corrupt ratio , i.e., the ratio between the number of injected nodes with the number of existing nodes in the graph. The links in the original graph are kept between the nodes in the corrupt graph. We corrupt the MAS graphs in the same way.

%Alpha or Yelp is a single graph itself. We partition it into $K$ graphs by HAC based on the features of the top eigenvectors and corrupt each graph in the same way as corrupting the academic graphs. The graph size $K$, determined by the optimal Silhouette Coefficient, is set as 5 on Alpha and 38 on Yelp,  

\vpara{Setup.} 
%For efficient training on the 4,800 corrupt academic graphs, we include 8 graphs in each batch and sample 48 abnormal nodes and XX normal nodes in each graph. To meet negative instances as much as possible, similar to simCLR~\cite{chen2020simple}, a popular pre-training scheme for visual representation learning, we also include all the nodes from other graphs in the same batch as the negative instances in addition to the pre-injected abnormal nodes in the concerned graph. But different from the pre-injected abnormal nodes, the sampled negative instances from other graphs during training only impact the loss function but are regardless of the graph convolution of the concerned graph. Since the number of the partitioned graphs in Alpha and Yelp is much smaller than that in AMiner and MAS, we set batch size as 1 on them. 
Without any labeled data (0\% labeled data), we train \spmodel and the baselines on the corrupt graphs and evaluate them on the same test set as the supervised \model. Then we also explore the few-shot learning settings, i.e., we further fine-tune the pre-trained models when given a proportion of labeled data.

\begin{figure*}[t]
	\centering
	\subfigure[AMiner]{\label{subfig:aminer}
		\includegraphics[width=0.24\textwidth]{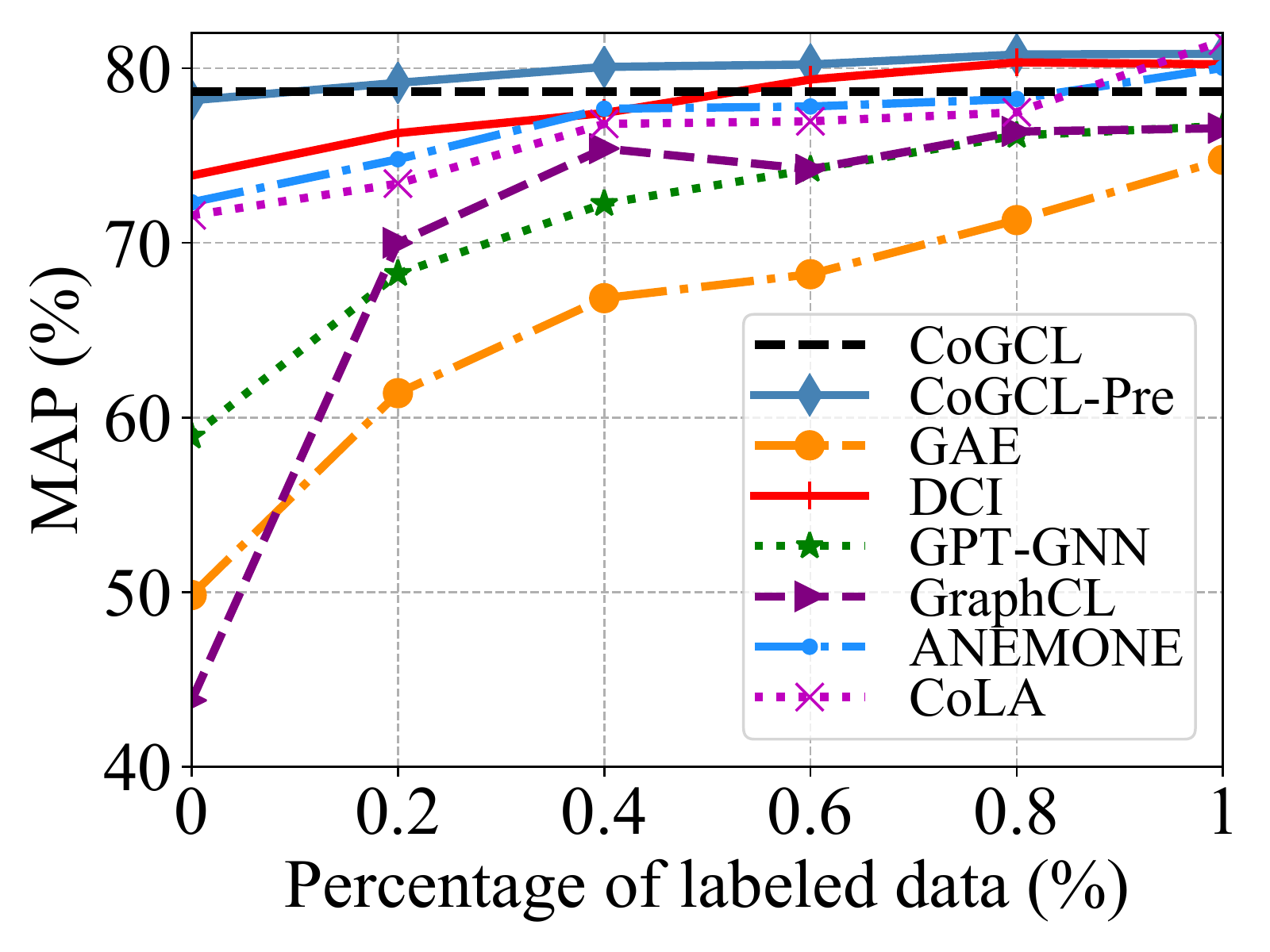}
	}
	\hspace{-0.1in}
	\subfigure[MAS]{\label{subfig:mas}
		\includegraphics[width=0.24\textwidth]{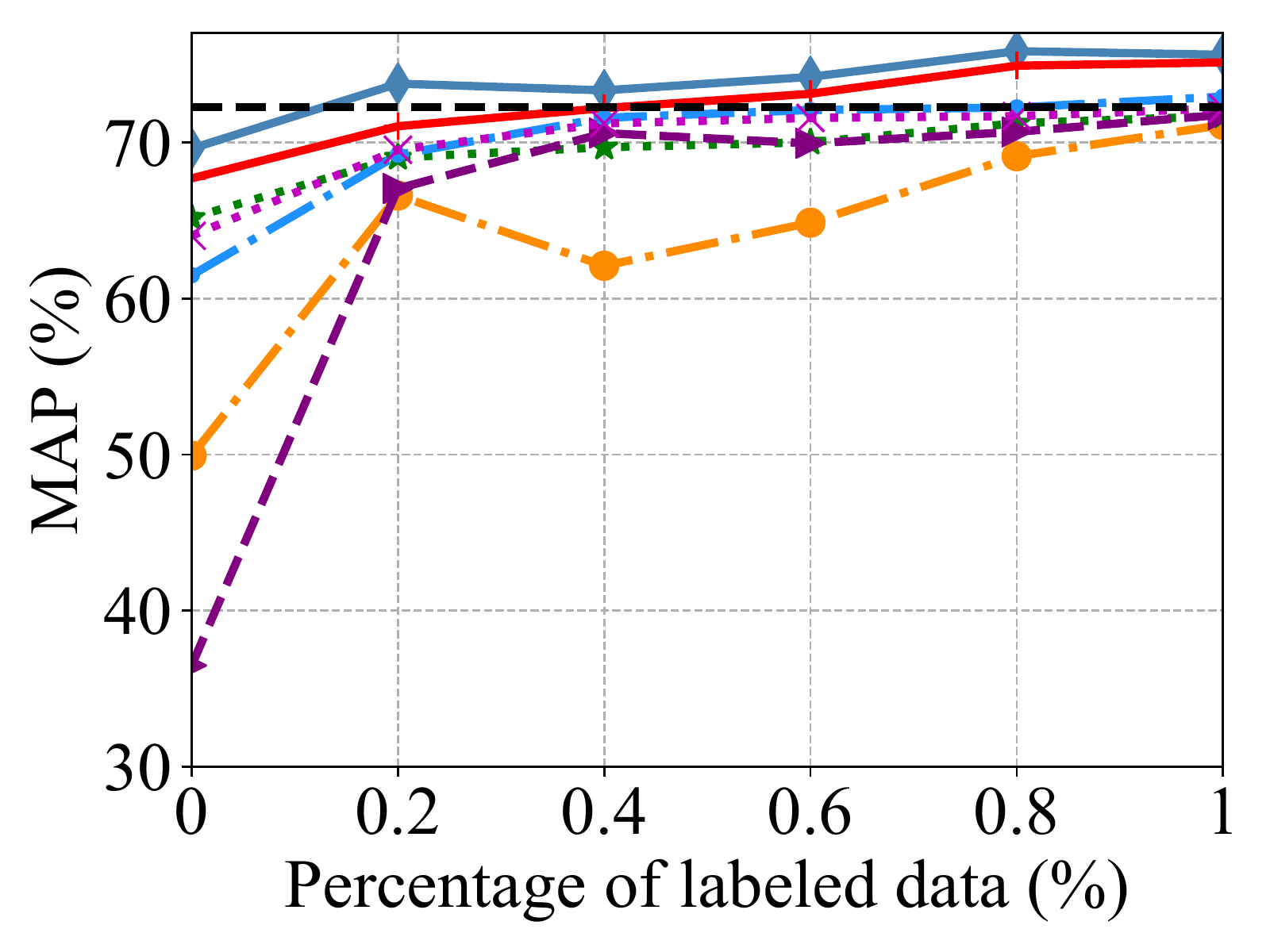}
	}	
	\hspace{-0.1in}
	\subfigure[Alpha]{\label{subfig:alpha}
		\includegraphics[width=0.24\textwidth]{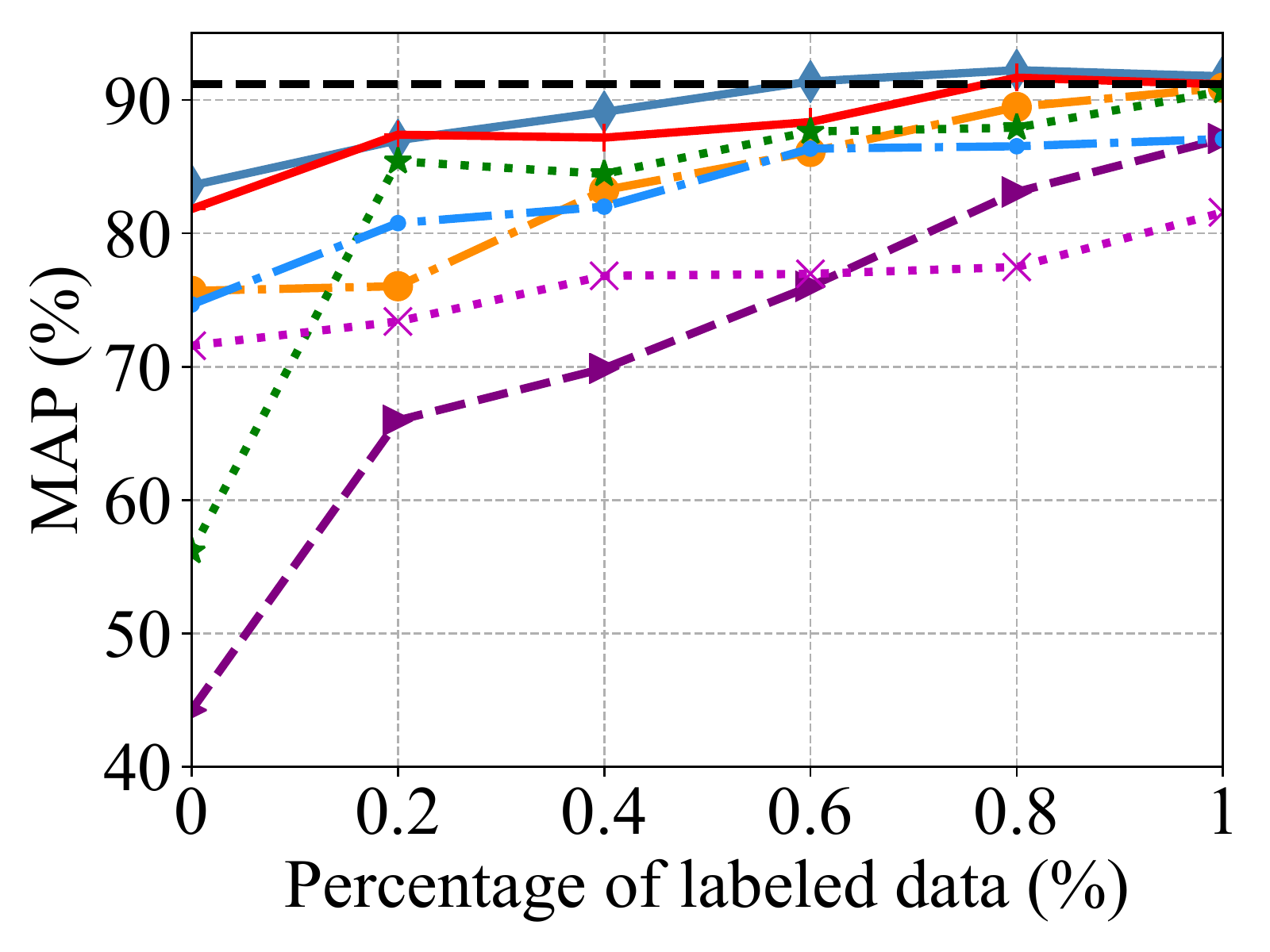}
	}
	\hspace{-0.1in}
	\subfigure[Yelp]{\label{subfig:yelp}
		\includegraphics[width=0.24\textwidth]{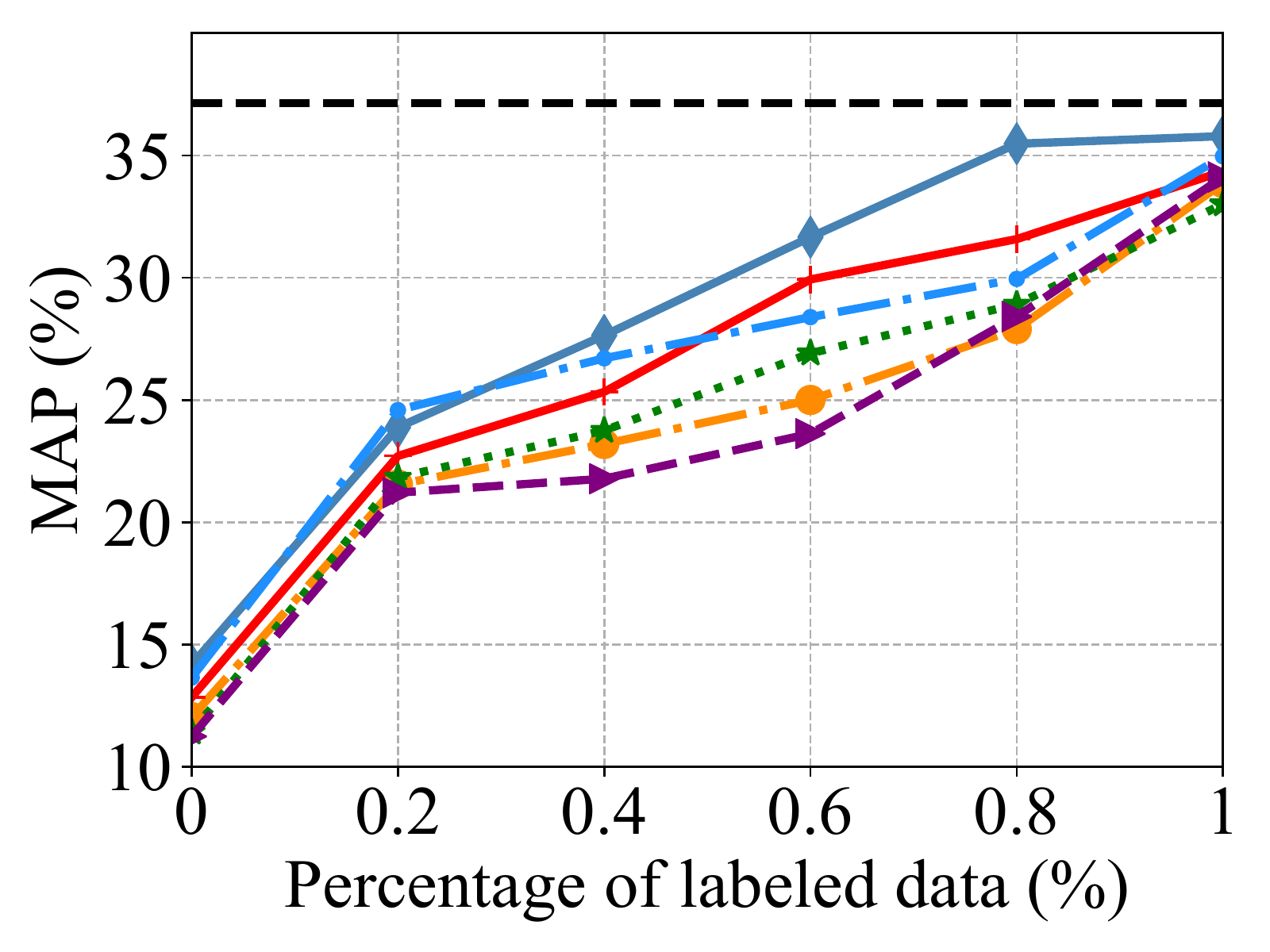}
	}
	\caption{\label{fig:pretrain} The pre-training and fine-tuning performance given different percentage of the labeled data on four datasets. The horizontal black dashed line represents the performance of the supervised \model.}
\end{figure*}

\vpara{Implementation.} 
For \pmodel, we follow the same setting as \model. For GAE, DCI, GPT-GNN, and GraphCL, we use the authors' official codes with the same training settings. Note that, for GraphCL, we try all the graph augmentation methods defined in the paper and and select the sub-graph augmentation that achieves the best performance in the test set without fine-tuning.

\vpara{Overall Results.} Fig.~\ref{fig:pretrain} shows the pre-training and fine-tuning performance of all the comparison methods given different percentage of labeled data on four datasets. 
From the results, we see that \spmodel (blue line) fine-tuned on 0\%, 10\%, and 60\% labeled data is comparable with the fully-supervised \smodel (black dashed line) on AMiner, MAS, and Alpha respectively. 
What's more, when \spmodel is fined-tuned on all the labeled data, it even outperforms \smodel by 1.72\%, 2.19\%, and 1.04\% in AUC on AMiner, MAS, and Alpha respectively, which reveals the effectiveness of the proposed pre-training framework. 

Our model also outperforms all the baselines when whatever percentage of labeled data is provided. 
GAE and GPT-GNN focus on reconstructing the links in the graphs.
GraphCL aims to capture the normal distribution of the whole graphs.
The objective of them is a little far away from anomaly detection. 
DCI performs the best among all the baselines. Given the global context as the query, DCI aims to contrast between the nodes within the sub-graph (normal nodes) and those from other sub-graphs (abnormal nodes). The objective is similar to \pmodel, but different from the corrupted graphs by \pmodel, the anomalies in DCI are thoroughly disconnected from the normal nodes and are independently embedded in other graphs, 
which may reduce the learning difficulty of the pseudo labels compared with the ground-truth labels. 
Both ANEMONE and CoLA under-perform the proposed GCCAD-pre. Although ANEMONE and CoLA propose a similar node to sub-graph contrast learning, they sample the sub-graphs by the local random walking algorithm instead of the global clustering algorithm in GCCAD-pre. The latter can result in more diverse normal samples which are more difficult to be distinguished. On the contrary, the former ignores such difficult normal samples during training, which will reduce the model generalization.

We also observe that on Yelp, none of pre-training models achieve prominent performance without any labeled data. 
On one hand, as shown in Table~\ref{tb:dataset}, because of the smallest concentration, Yelp is the most diverse dataset, which prevents \pmodel, DCI, and GraphCL from discovering the proper normal pattern.
On the other hand, since the ratio of the suspicious links on Yelp, 22.70\%, is the largest among the datasets, GAE and GPT-GNN that target at preserving the link homophily will wrongly reconstruct those noisy links. 
% \textcolor{blue}{The noisy structure information also prevents \spmodel to split sub-graphs rationally.} 
However, \spmodel still performs best compared with other pre-training models on the most of the percentage of labeled data is provided, which implies the ability of \spmodel to distinguish the normal nodes from the abnormal ones.     

\vpara{Corrupt Ratio.}
Fig.~\ref{subfig:corrupt_ratio} presents the correlation between the ratio of the injected nodes (corrupt ratio) and the performance of \spmodel on AMiner and Alpha without fine-tuning. The results show that the best performance is achieved when about 15\% of abnormal nodes are injected. The corrupt ratio is set in the same way as other datasets.

\vpara{Clustering Number.}
Fig.~\ref{fig:cluster} presents the correlation between the graph (cluster) number $K$ and the inertia gap $\sigma\left(K\right)$ between $K$ and $K+1$ on Alpha and Yelp. The optimal $K^*$ that achieves the best performance of \spmodel without fine-tuning are marked as red. We observe that \spmodel performs the best when $K^*$ (red) is approximately around the elbow point where the trend of inertia changes from steep to stable. Although we choose the elbow points on Yelp, Yelp is so diverse that none of  the graph pre-training methods can achieve desired performance.

\vpara{Features for Clustering.}
We compare the effect of the clustering algorithm by the structure information or feature information on the pre-training performance. We perform K-means on the graph of Yelp by the structure information, and the result is much poorer than clustering by the feature information (i.e., 12.56\% versus. 14.17\% in terms of MAP), which indicates the more noisy structural information hampers the clustering as well as the pre-training performance.

% Figure~\ref{fig:cluster} presents the correlation between the elbow point, where the trend of inertia changes from steep to stable, with respect to the graph (cluster) number $K$ and the performance of \spmodel on Alpha and Yelp, where the x-axis is the cluster number $K$, and the y-axis denotes $\sigma\left(K\right)$ representing the inertia gap between $K$ and $K+1$. We observe that \spmodel performs the best when $K^*$ (red) is around the elbow point. However, although we choose the elbow points on yelp, the best performance of \spmodel is also unsatisfied, because Yelp is so diverse that \spmodel cannot achieve desired performance.

%We present the correlation between the graph (cluster) number $K$ and the best performance of \spmodel on Alpha and Yelp without fine-tuning in Figure~\ref{subfig:k} and Figure~\ref{subfig:auc_k}. The results show that when $K$ equals 5, \spmodel performs the best and the Silhouette Coefficient is also optimal. A clear correlation between $K$ and the best performance is not easy to be found on Yelp, because Yelp is so diverse that \spmodel cannot achieve expected performance whatever $K$ is given.

%% file: related.tex
\section{Related Work}
\label{sec:related}

\subsection{Graph-based Anomaly Detection}
Graph neural networks have been studied to detect anomalies in various domains, such as detecting review spams in business websites~\cite{dou2020enhancing,li2019spam,liu2020alleviating, liu2021pick}, 
rumors in social media~\cite{bian2020rumor,wu2020graph,yang2020rumor}, fake news~\cite{AA-HGNN2020,FangModel}, financial fraud~\cite{liu2019geniepath,xu2021towards,rao2021xfraud}, insurance fraud~\cite{liang2019uncovering}, and bot fraud~\cite{yao2020botspot}.
Most of them target how to design a proper aggregator that can distinguish the effects of different neighbors and reduce the inconsistency issue~\cite{liu2020alleviating} during message passing. For example, GAS~\cite{li2019spam} adopts the vallina GCN~\cite{kipf2016semi}. SemiGNN~\cite{SemiGNN} and Player2Vec~\cite{zhang2019key} apply attention mechanisms to assign low weights to suspicious links. To thoroughly reduce the negative influence of the suspicious links, several attempts~\cite{franceschi2019learning, liu2020alleviating} have been made to remove the suspicious links before graph convolution. CARE-GNN~\cite{dou2020enhancing} further adopts reinforcement learning to sample links according to the suspicious likelihood.
None of them are aware of the limitations caused by the objective function. To our knowledge, we are the first to change the commonly-used binary classification into the graph contrastive learning paradigm. 

\subsection{Graph Pre-training Schemes}
With the advances of self-supervised (pre-training) techniques in visual representation learning~\cite{chen2020simple,he2020momentum, chen2021exploring}, graph pre-training schemes have also attracted increasing attention. A naive GNN pre-training scheme is to reconstruct the vertex adjacency matrix, and GAE~\cite{kipf2016variational} and GraphSAGE~\cite{Graphsage} are two representative models. In addition to preserve the structure homophily, GPT-GNN~\cite{hu2020gpt} preserves the attribute homophily by predicting the masked node attributes. Motivated by~\cite{hjelm2018learning}, DGI~\cite{velickovic2019deep} and Infograph~\cite{Infograph} have been proposed to maximize the mutual information between the embeddings of the entire graph and the node within it. 
GraphCL~\cite{you2020graph} maximizes the mutual information between the embeddings of two graph instances, which are obtained from the same graph via graph data augmentation. Later, GCC~\cite{qiu2020gcc} shrinks the contrast between graphs into that between ego-networks, where the ego-network instances are obtained via random walk with start from the concerned node. All of them are proven to be useful for the downstream node classification task, but are not specifically proposed to solve the anomaly detection problem. DCI~\cite{wang2021decoupling} is a cluster-based version of DGI, 
which maximizes the fine-grained mutual information between the embeddings of a cluster and the nodes within it. DCI is proposed for graph anomaly detection. However, it is thoroughly unsupervised model and different from the corrupted graphs by \pmodel, the anomalies in DCI are thoroughly disconnected from the normal nodes, 
which may reduce the learning difficulty of the pseudo labels.

%% file: conclusion.tex
\section{Conclusion and Future Direction}
\label{sec:con} 

This paper proposes \model, a %context-aware GNN model by contrastive learning, 
graph contrastive learning model
for anomaly detection. 
Instead of directly classifying node labels, \smodel contrasts abnormal nodes with normal ones in terms of their distance to the global context of the graph. We further extend \smodel to an unsupervised version by designing a graph corrupting strategy to generate the synthetic node labels. 
To achieve the contrastive goal, we design a three-stages GNN encoder to infer and remove suspicious links during message passing, as well as to learn the global context.
The experimental results on four real-world datasets demonstrate that \smodel significantly outperforms the baselines and \spmodel without any fine-tuning can achieve comparable performance with the fully-supervised version on the two academic datasets. 
As the performance of \smodel highly resorts to the consistent representation of the global context, \smodel performs relatively worse on Yelp which is a large single-graph and the most diverse dataset among all the evaluated ones. To ameliorate this, one can trivially sample a sub-graph among the node to be detected as the local context. We leave this in the future work.
% We will continue to study more promising pre-training strategies for detecting anomalies on Yelp, the most diverse dataset among all the evaluated ones.

%% file: ack.tex
\section{Acknowledgments}
This work is supported by Natural Science Foundation of China (62076245), Technology and Innovation Major Project of the Ministry of Science and Technology of China under Grant 2020AAA0108400 and 2020AAA0108402, National Science Foundation for Distinguished Young Scholars
(No.~61825602), and Natural Science Foundation of China under Grant (No.~61836013).